\documentclass[lettersize,journal]{IEEEtran}
\usepackage{amssymb}
\usepackage{algorithm}
\usepackage{algorithmic}
\usepackage{cite}
\usepackage{textcomp}
\usepackage{graphicx}
\usepackage{graphics}
\usepackage{epsfig}
\usepackage{epstopdf}
\usepackage{float}
\usepackage{color}
\usepackage[cmex10]{amsmath}
\usepackage{latexsym,amsfonts}
\usepackage{amsthm}
\usepackage{url}
\usepackage{longtable}
\usepackage[figuresright]{rotating}
\usepackage{listings}
\usepackage{etoolbox}
\usepackage{hhline}
\usepackage{subfig}
\usepackage{nomencl}
\makenomenclature

\usepackage{amsthm}
\usepackage{mathtools}
\usepackage{multirow}
\usepackage{tikz}
\usetikzlibrary{backgrounds,automata}
\usepackage{dblfloatfix} 
\usepackage{placeins}
\usepackage{multicol}
\usepackage{lipsum}
\usepackage{cuted}
\usepackage{algorithm}
\usepackage{algorithmic}
\usetikzlibrary{shapes,arrows,positioning}
\usetikzlibrary{arrows.meta}
\usepackage{circuitikzgit}
\usepackage{blindtext}
\usepackage{mathrsfs}
\usepackage{ nccmath}
\usetikzlibrary{dsp,chains}
\usepackage{fancyhdr}       
\fancyheadoffset[L]{0.25in}

\makeatletter
\patchcmd{\@makechapterhead}{50\p@}{20pt}{}{}
\patchcmd{\@makeschapterhead}{50\p@}{20pt}{}{}
\makeatother

\fancypagestyle{plain}{
	\fancyhf{} 
	\fancyhead[C]{\thepage} 
	
	}

\DeclareMathAlphabet{\mathpzc}{OT1}{pzc}{m}{it}

\usepackage{setspace}

\usepackage{ifthen,xkeyval,xfor,amsgen}
\usepackage[acronym,toc, nogroupskip]{glossaries}

\newglossary[slg]{symbols}{syi}{sbl}{List of Symbols}

\makeglossaries

\usepackage{hyperref}

\newcommand{\cref}[1]{Constraint~\ref{#1}}

\newcommand{\ignore}[1]{}

\def\BibTeX{{\rm B\kern-.05em{\sc i\kern-.025em b}\kern-.08em
    T\kern-.1667em\lower.7ex\hbox{E}\kern-.125emX}}
\usepackage{balance}
\begin{document}
\include{Lists}
\title{Kalman Filtering for Precise Indoor Position and Orientation Estimation Using IMU and Acoustics on Riemannian Manifolds}
\author{Mohammed H. AlSharif, Mohanad Ahmed, Mohamed Siala, Tareq Y. Al-Naffouri
\thanks{Mohammed H. AlSharif is with the research and development department, Aramco Research Center, Thuwal 23955, Saudi Arabia (e-mail: mohammed.alsharif@kaust.edu.sa). \\
Mohanad Ahmed, and Tareq Y. Al-Naffouri are with the Division of Computer, Electrical and Mathematical Sciences and Engineering, King Abdullah University of Science and Technology, Thuwal 23955, Saudi Arabia.\\ Mohamed Siala is with the Department of Applied Mathematics, Signals and Communication, Higher School of Communication of Tunis (Sup’Com), Ariana 2083, Tunisia.\\
Preprint. Under review.}}


\maketitle

\begin{abstract}

Indoor tracking and pose estimation, i.e., determining the position and orientation of a moving target, are increasingly important due to their numerous applications. While Inertial Navigation Systems (INS) provide high update rates, their positioning errors can accumulate rapidly over time. To mitigate this, it is common to integrate INS with complementary systems to correct drift and improve accuracy. This paper presents a novel approach that combines INS with an acoustic Riemannian-based localization system to enhance indoor positioning and orientation tracking. The proposed method employs both the Extended Kalman Filter (EKF) and the Unscented Kalman Filter (UKF) for fusing data from the two systems. The Riemannian-based localization system delivers high-accuracy estimates of the target's position and orientation, which are then used to correct the INS data. A new projection algorithm is introduced to map the EKF or UKF output onto the Riemannian manifold, further improving estimation accuracy. Our results show that the proposed methods significantly outperform benchmark algorithms in both position and orientation estimation. The effectiveness of the proposed methods was evaluated through extensive numerical simulations and testing using our in-house experimental setup. These evaluations confirm the superior performance of our approach in practical scenarios.
\end{abstract}

\section{Introduction}
Accurate position and orientation estimation is crucial for various applications such as inventory management \cite{sun2023indoor}, smart homes, smart manufacturing, and health monitoring \cite{feng2020kalman}, \cite{khelifi2019survey}. High accuracy is fundamental to achieving the desired performance in these applications. Advances in sensor technology and computational techniques have led to the increased use of inertial sensors for navigation \cite{titterton2004strapdown}. Historically, inertial sensors were mechanically isolated from the target's rotational motion, which ensured high accuracy \cite{titterton2004strapdown}. 
Modern inertial systems have simplified the mechanical complexity by rigidly attaching sensors to the target. These attached “strapped down” sensors reduced the cost and size but increased the computational complexity. In this paper, we focus on a strapdown inertial navigation system (INS) to estimate position, orientation, and velocity.\par

The INS operates based on kinematics and classical Newtonian mechanics, utilizing inertial measurement units (IMUs) comprised of accelerometers (motion sensors) and gyroscopes (rotation sensors) to measure acceleration and angular velocity. While the INS can use these parameters to estimate position, orientation and velocity at a high rate without external references, e.g. beacons, positioning and orientation errors accumulate over time \cite{feng2020kalman}. To mitigate these errors, navigation systems typically integrate additional positioning methods. In our previous work \cite{9528945}, we developed an acoustic-based location and orientation estimation system using Riemannian optimization, which performs well under line-of-sight (LOS) conditions but suffers in non-line-of-sight (NLOS) scenarios due to the nature of the acoustic waves. Moreover, the system in \cite{9528945} has a lower update rate compared to INS.\par

To address these limitations, we propose a Kalman filtering approach to integrate estimates from Riemannian localization methods \cite{9528945} with those from the INS. This integration aims to enhance accuracy of the estimated position and orientation in NLOS conditions and over extended periods.\par

The main contributions of this paper, detailed in Section \ref{fusion}, are as follows:
\begin{itemize}
\item A novel orientation correction method that models the orientation matrix as a state vector alongside position and velocity, resulting in new Extended Kalman Filter (EKF) formulations.
\item A new projection algorithm to improve the accuracy of both position and orientation estimates obtained using EKF and Unscented Kalman Filter (UKF).
\item Development and implementation of a new experimental setup to validate the proposed algorithms. 
\item An experimental evaluation of the proposed algorithms, complemented by numerical simulations.
\end{itemize}

This paper includes several symbols, which are defined in the nomenclature table for quick reference.

\nomenclature[01]{\(\mathbf{C}_b^n\)}{\small Orientation matrix transforming vectors from $b$-frame to $n$-frame}
\nomenclature[02]{\(\dot{\mathbf{C}}_b^n\)}{\small Rate of change of $\mathbf{C}_b^n$ with time}
\nomenclature[03]{\(\mathbf{\omega}_{nb}^b\)}{\small Angular velocity vector as measured by triaxial gyroscope}
\nomenclature[04]{\(\mathbf{\Omega}_{nb}^b\)}{\small Skew symmetric matrix formed from the elements of $\mathbf{\omega}_{nb}^b$ }
\nomenclature[05]{\(\mathbf{a}^b \)}{\small Acceleration of the center of the mobile device in $b$-frame }
\nomenclature[06]{\(\mathbf{a}^n \)}{\small Acceleration of the center of the mobile device in $n$-frame }
\nomenclature[07]{\(\mathbf{v}_c^n \)}{\small Velocity of the center of the mobile device in $n$-frame }
\nomenclature[08]{\(\mathbf{p}_c^n \)}{\small Position of the center of the mobile device in $n$-frame }
\nomenclature[09]{\(\mathbf{p}_i \)}{\small Position of the $i^{\text{th}}$ acoustic receiver in $n$-frame }
\nomenclature[10]{\(\mathbf{b}_i \)}{\small Position of the $i^{\text{th}}$ acoustic transmitter (i.e. beacon) in $n$-frame }
\nomenclature[11]{\(d \)}{\small Side length of the isosceles triangle }
\nomenclature[12]{\(\mathcal{M} \)}{\small Isosceles triangle manifold}
\nomenclature[13]{\(\mathcal{T}_{\mathbf{P}}\mathcal{M}\)}{\small Tangent space to $\mathcal{M}$ at $\mathbf{P}$}
\nomenclature[14]{\(r_{ij} \)}{\small Distance between the $i^{\text{th}}$ receiver and the $j^{\text{th}}$ transmitter}
\nomenclature[15]{\(\mathbf{P} \)}{\small Matrix of isosceles triangle vertices positions where $\mathbf{p}_i$ is the $i^{\text{th}}$ column}
\nomenclature[16]{\({\nabla}_{\mathbf{P}} f \)}{\small Euclidean gradient of the function $f$ with respect to $\mathbf{P}$}
\nomenclature[17]{\({\nabla}^2_{\mathbf{P}} f \)}{\small Euclidean Hessian of the function $f$ with respect to $\mathbf{P}$}
\nomenclature[18]{\({\overline{\nabla}}_{\mathbf{P}} f \)}{\small Riemannian gradient of the function $f$ with respect to $\mathbf{P}$}
\nomenclature[19]{\({\overline{\nabla}}^2_{\mathbf{P}} f \)}{\small Riemannian Hessian of the function $f$ with respect to $\mathbf{P}$}
\nomenclature[20]{\({\Pi_{\mathbf{P}}}(.)\)}{\small Orthogonal projection operator onto the tangent space $\mathcal{T}_{\mathbf{P}}\mathcal{M}$}
\nomenclature[21]{\(\text{R}_{\mathbf{P}}(.)\)}{\small Retraction operator from the tangent space $\mathcal{T}_{\mathbf{P}}\mathcal{M}$ to the manifold $\mathcal{M}$}
\nomenclature[22]{\(\mathbf{p}_c^R \)}{\small Position of the center of the mobile device in $n$-frame estimated using Riemannian localization }
\nomenclature[23]{\(\mathbf{x} \)}{\small State vector consists of centoid position and velocity, and the vectorized orienation matrix of the mobile device }
\nomenclature[24]{\(\bigotimes \)}{\small Kronecker product operation}
\nomenclature[25]{\({\mathbf{\Omega}_{nb}^b}^\times \)}{\small Kronecker product between $\mathbf{\Omega}_{nb}^b$ and a $3 \times 3$ identity matrix}
\nomenclature[26]{\({\mathbf{\Omega}_{nb}^2}^\times \)}{\small Square of the matrix ${\mathbf{\Omega}_{nb}^b}^\times$}

\printnomenclature

\par   
\section{Related Work}
\par 
With the increasing demand for precise indoor location estimation, indoor localization has gained significant attention. Therefore, various indoor positioning and orientation estimation systems have been developed to meet the need for highly accurate navigation solutions. Technologies used for indoor navigation include acoustic waves \cite{hightower2001location}, radio-frequency \cite{whitehouse2007practical}, infrared radiation \cite{yuzbacsiouglu2005improved}, and laser signals \cite{amann2001laser}. While Bluetooth and WiFi-based systems are cost-effective, they suffer from low accuracy and require pre-calibration. In contrast, infrared and laser-based systems offer high accuracy but come with high hardware costs \cite{zafari2019survey}. Acoustic and ultra-wideband (UWB) systems provide high accuracy at a lower cost \cite{zafari2019survey}.
\par 

While the choice of technology provides the foundation for indoor localization systems, the effectiveness of these systems ultimately hinges on the techniques employed to leverage these technologies. Positioning techniques have evolved significantly over the past decades, with the most popular being received signal strength (RSS)\cite{cappelli2023enhanced}, angle of arrival (AoA)\cite{gabbrielli2022rails}, time of flight (ToF)\cite{zafari2019survey}, and time difference of arrival (TDoA)\cite{santoro2023uwb}. RSS-based navigation is simple and low-cost but requires pre-calibration and suffers from low accuracy \cite{cypriani2009open}. AoA-based systems offer high accuracy at close range but degrade significantly with distance due to angular estimation errors \cite{kumar2014accurate}. In contrast, ToF and TDoA-based techniques provide higher accuracy, especially over longer distances \cite{alsharif2021range}. These techniques are categorized as infrastructure-based, requiring base stations and line-of-sight (LOS) for accurate measurements, which limits their effectiveness in non-LOS scenarios \cite{wang2022smartphone}.\par

To overcome the LOS requirement, infrastructure-free techniques based on inertial sensors have been developed \cite{wang2022smartphone}. INS uses IMUs containing accelerometers and gyroscopes to estimate position and orientation \cite{titterton2004strapdown}. While numerical integration of acceleration and angular velocity can provide accurate estimates over short periods, accuracy degrades over time due to error accumulation \cite{wang2022smartphone}. Pedestrian dead reckoning (PDR)\cite{shi2023pedestrian}, which estimates position based on step counts and step length, is another technique but is limited to specific movement patterns, making it less suitable for robots \cite{lee2021indoor}.\par 

Infrastructure-free systems like INS offer high accuracy in the absence of LOS but suffer from accuracy degradation over time \cite{wang2022smartphone}. Conversely, infrastructure-based systems maintain high accuracy over long periods but require LOS \cite{wang2022smartphone}. Consequently, integrating INS with infrastructure-based systems is common practice to achieve high accuracy over extended periods, even without LOS.\par

One notable indoor localization system is the Cricket system \cite{priyantha2005cricket}, which estimates both position and orientation using TDoA and AoA methods with five receivers. However, it requires LOS and does not leverage the fixed geometry of the receiver array, which could enhance accuracy. Fusion of INS with infrastructure-based systems can improve navigation accuracy. For instance, Fan \textit{et al.} \cite{fan2017data} developed an improved adaptive Kalman filter (IAKF) to fuse TDoA-based localization with INS, achieving accurate position estimates in 2D. Similarly, Zhong \textit{et al.} \cite{zhong2018integration} used EKF to integrate ToF-based positioning with INS for better accuracy. Li \textit{et al.} \cite{li2019indoor} employed EKF to fuse UWB-based and PDR-based positioning. The algorithm developed in \cite{li2019indoor}, focused on using PDR for 2D navigation, which would necessitate significant modifications to adapt the approach for 3D navigation. Another navigation system that combines acoustic-based positioning with PDR-based positioning using an EKF was proposed in \cite{gualda2021locate}. A limitation of the system described in \cite{gualda2021locate} is that it requires a smartphone to be attached to the pedestrian's leg. To enhance the accuracy of PDR-UWB-based positioning, Lee \textit{et al.} \cite{lee2021indoor} proposed a deep learning approach aimed at automatically adapting the PDR parameters for each individual. However, the algorithm in \cite{lee2021indoor} has a drawback in that it relies on the pedestrian walking in a straight line to accurately determine heading. Additionally, extending the algorithm in \cite{lee2021indoor} from 2D to 3D navigation poses significant challenges. In \cite{xu2020tightly}, the authors fuse a UWB-based positioning system with INS using a fixed-lag extended finite impulse response smoother (FEFIRS) algorithm. The algorithm described in \cite{xu2020tightly} has the main drawback of requiring a certain lag to improve positioning results. Furthermore, Feng \textit{et al.} \cite{feng2020kalman} proposed a navigation system that integrates INS with a UWB-based positioning system using both EKF and UKF alongside two novel motion models.\par 

While the fusion algorithms discussed in  \cite{fan2017data, zhong2018integration, li2019indoor, gualda2021locate, lee2021indoor, xu2020tightly, feng2020kalman} improve positioning accuracy, they do not necessarily enhance orientation accuracy, which can degrade performance over time. Orientation estimation typically relies on accelerometers, gyroscopes \cite{xie2017holding, nguyen2016user, bravo2017comparison, manos2018gravity, hong2010heading}, magnetometers \cite{zhou2016pedestrian, afzal2011magnetic, angermann2012characterization, cho2006mems}, or a combination of them \cite{kim2004step,shin2012indoor,ma2015hiheading,yang2016step}. Among these, magnetometer-based approaches are widely used \cite{poulose2019performance}, but their accuracy can be significantly impacted by magnetic disturbances in indoor environments \cite{poulose2019performance}. On the other hand, accelerometer-based methods suffer from error accumulation, while gyroscope-based methods experience drift errors in addition to error accumulation \cite{poulose2019performance}. Consequently, relying solely on these three sensors for orientation estimation can compromise both orientation and positioning accuracy. The performance analysis provided in \cite{poulose2019performance} compares various sensor fusion techniques for orientation estimation using smartphone sensors, including the Linear Kalman Filter (LKF), EKF, UKF, Particle Filter (PF), and Complementary Filter (CF). Although the work in \cite{poulose2019performance} addresses orientation fusion, it relies solely on inertial sensors, which leads to accuracy degradation over time.\par

While previous work improves positioning accuracy through various fusion algorithms, orientation accuracy often remains unaddressed. Our focus is on fusing acoustic-ToF-based positioning with INS to enhance both position and orientation estimation. We propose novel algorithms based on Riemannian optimization to correct both position and orientation estimates obtained from INS. The acoustic system employs an array of receivers arranged in the shape of an isosceles triangle. As demonstrated in our previous work \cite{9528945}, Riemannian optimization effectively addresses the localization problem with high accuracy. The proposed algorithm leverages the positions of the triangle vertices to refine both the position and orientation estimates provided by the INS via EKF-based and UKF-based fusion. The primary contribution of this paper is the development of a novel algorithm that integrates Riemannian optimization to enhance the correction of position and orientation estimates derived from the INS. Thanks to the innovative projection algorithm detailed in Section \ref{fusion}, the proposed method improves the accuracy of position and orientation in typical indoor environments, as validated by numerical simulations and experiments.
  
\section{Problem Formulation}
Accurate estimation of the position and orientation of a moving target is pivotal in INS \cite{titterton2004strapdown}. While inertial navigation based on IMUs has high accuracy over short periods, its accuracy severely degrades over time due to error accumulation from the integration operations used in INS \cite{titterton2004strapdown}. Therefore, it is common practice to fuse measurements from different sensors to improve the tracking accuracy of INS. We propose using three acoustic receivers arranged at the vertices of an isosceles triangle to enhance the accuracy of the estimated orientation and position obtained using INS.\par
\begin{figure}[h]
		\centering
		\subfloat[Illustration of MD]{\includegraphics[width=3cm, height = 3 cm]{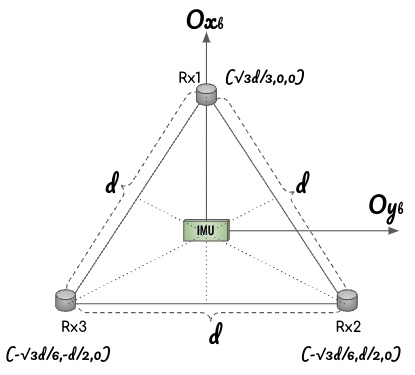}}
		\label{md_setup}
		\subfloat[Acoustic Rx array with IMU]{\includegraphics[width=5cm, height = 4 cm]{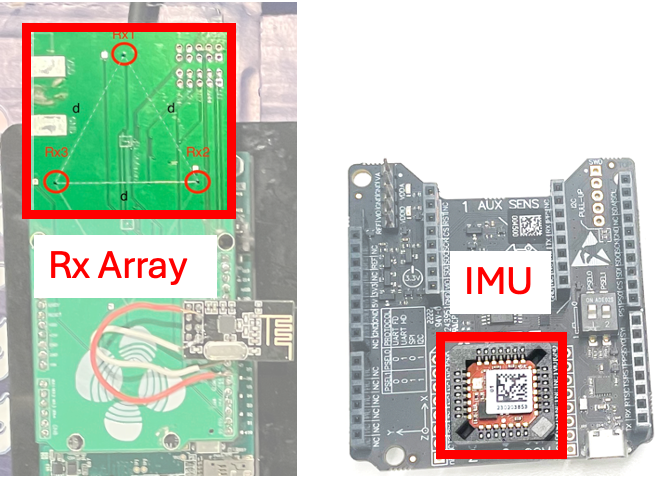}}
		\caption{Mobile device}
		\label{md}
\end{figure}

The proposed system consists of three acoustic receivers with unknown positions placed at the vertices of an isosceles triangle, four acoustic transmitters (beacons) with known positions, and an IMU placed at the centroid of the isosceles triangle. The three receivers, together with the IMU, form the mobile device (MD). While the proposed system is evaluated using an equilateral triangle, which is a special case of an isosceles triangle, the proposed algorithm can be applied to any isosceles triangle. Figure \ref{md} shows an illustration of the mobile device in the body frame.\par  

As mentioned previously, while INS provides high rates, it suffers from accuracy degradation due to error accumulation. In contrast to INS, the position and orientation of the mobile device can be estimated with high accuracy using the three acoustic receivers without degradation over time. However, the acoustic positioning system requires an LOS between the transmitters and receivers to achieve high accuracy. Additionally, acoustic-based measurements are carried out at a slower rate compared to INS. Therefore, the proposed algorithm fuses the two measurements to achieve high-accuracy position and orientation estimation at a high rate, even in NLOS scenarios and over long periods.\par 

The position and velocity of the centroid of the triangle in Figure \ref{md}, along with its orientation, are estimated using IMU measurements and classical INS equations \cite{titterton2004strapdown}. These estimates are corrected using the positions of the vertices of the triangle, estimated using Riemannian localization \cite{9528945}. The main contributions of this paper, presented in Section \ref{fusion}, include the correction algorithm with a new derivation of the EKF equations and a novel projection algorithm on the Riemannian manifold that improves tracking accuracy.\par

\section{Position and Orientation Estimation}
This section provides a brief summary of the inertial navigation \cite{titterton2004strapdown} and Riemannian localization algorithms \cite{9528945} used in this work. For more details on INS and Riemannian localization, we refer the reader to \cite{titterton2004strapdown} and \cite{9528945}.

\subsection{IMU Position and Orientation Estimation}
\begin{figure}[!h]
	\centering
	\includegraphics[width=5cm, height= 4cm]{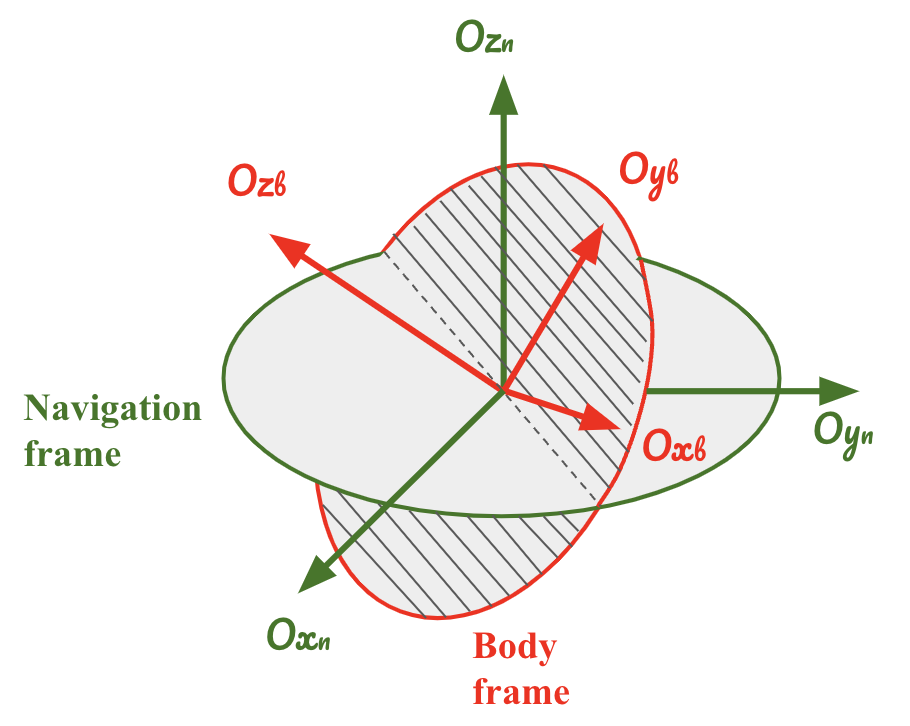}
	\caption{The navigation and body frames axes}
	\label{frames}
\end{figure}

Navigation systems should provide the position and orientation of an MD with respect to a fixed reference frame, known as the navigation frame. However, the IMU measures acceleration and angular velocity with respect to a moving frame, known as the body frame, which is attached to the MD. To explain the motion of the MD, we need to select a suitable coordinate system \cite{feng2020kalman}. Let's denote the navigation (reference) frame and body frame axes as $[O_{x_n}, O_{y_n}, O_{z_n}]$ and $[O_{x_b}, O_{y_b}, O_{z_b}]$, respectively. The centroid of the triangle, shown in Figure \ref{md}, represents the origin of the body frame. The axes of the body frame are aligned with the roll, pitch, and yaw axes of the MD. The origin of the navigation frame is fixed in the simulation and experimental setup. The axes of the navigation frame are aligned with the directions of north, east, and up. Each of the two frames form a right-hand system. Figure \ref{frames} shows the axes of the body and navigation frames.\par

Any vector $\mathbf{v}^b$ expressed in the body frame can be transformed to the navigation frame by pre-multiplying by the orientation matrix $\mathbf{C}_b^n$ \cite{titterton2004strapdown}:
\begin{equation}
	\mathbf{v}^n = \mathbf{C}_b^n \mathbf{v}^b,
\end{equation}
where $\mathbf{C}_b^n$ is a $3 \times 3$ matrix that defines the attitude of the body frame with respect to the navigation frame, and $\mathbf{v}^n$ is the transformed vector expressed in the navigation frame. The attitude (orientation) matrix can be represented using direction cosines, Euler angles, or quaternions \cite{titterton2004strapdown}.\par

The IMU used in this work contains a triaxial accelerometer and a triaxial gyroscope which measure the 3D acceleration and angular velocity, represented in the body frame of the MD, respectively. However, the final estimated position and velocity of the mobile device should be expressed in the navigation frame. Therefore, it is required to estimate the orientation matrix $\mathbf{C}_b^n$. Let the angular velocity measured by the IMU be $\mathbf{\omega}_{nb}^b=[\omega_x, \omega_y, \omega_z]^T$, which represents the turn rate of the MD with respect to the navigation frame as measured by the gyroscope. Under the assumption of known initial conditions, the angular velocity can be integrated over time to obtain the orientation of the MD using the following equation \cite{titterton2004strapdown}:

\begin{equation}
	\mathbf{C}_b^n(k+1)=\mathbf{C}_b^n(k) + \Delta t \Dot{\mathbf{C}_b^n} (k),
	\label{C_num}
\end{equation}
where $\Delta t$ is the IMU sampling period and $\dot{\mathbf{C}_b^n}$ is given by 

\begin{equation}
	\dot{\mathbf{C}_b^n} (k)=\mathbf{C}_b^n(k) \mathbf{\Omega}_{nb}^b  (k),
\end{equation}
where $\mathbf{\Omega}_{nb}^b$ is the skew-symmetric matrix given by
\begin{equation}
	\mathbf{\Omega}_{nb}^b (k) = \begin{bmatrix} 0 & -\omega_z  (k) & \omega_y  (k)\\ 
		\omega_z  (k) & 0 & -\omega_x  (k) \\
		-\omega_y  (k) & \omega_x  (k) & 0
	\end{bmatrix}.
\end{equation}
The orientation matrix is an orthogonal matrix, which means $\mathbf{C}_n^b(k)=(\mathbf{C}_b^n(k))^{-1}=(\mathbf{C}_b^n(k))^{T}$. The orthogonality of the orientation matrix is approximately achieved, assuming that the first-order approximation given in Equation \ref{C_num} is accurate under a high sampling rate. While the matrix $\mathbf{C}_b^n(k)$ transforms a vector from the body frame to the navigation frame, $\mathbf{C}_n^b(k)$ transforms a vector from the navigation frame to the body frame. The operation $[.]^T$ is the transpose operation. Consequently, the acceleration $\mathbf{a}^b = [a_x^b, a_y^b, a_z^b]^T$ measured by the IMU can be expressed in the navigation frame using the following equation:

\begin{equation}
	\mathbf{a}^{n'}(k) = [a_x^{n'}(k), a_y^{n'}(k), a_z^{n'}(k)]^T = \mathbf{C}_b^n(k)\mathbf{a}^{b}(k). 
\end{equation}

The acceleration vector $\mathbf{a}^{n'}$ contains a gravitational acceleration component $g$ which could be removed to obtain the acceleration at the center of the mobile device in the navigation frame as follows \cite{feng2020kalman}:

\begin{equation}
	\mathbf{a}^n(k) = \begin{bmatrix} a_x^{n'}(k) \\ a_y^{n'}(k) \\ a_z^{n'}(k) \end{bmatrix} - \begin{bmatrix} 0 \\ 0 \\ g \end{bmatrix}.
\end{equation}
Under a high IMU update rate, the acceleration during the sampling period can be assumed to be constant. Consequently, the velocity at the center of the MD $\mathbf{v}_c^n$ at time $k+1$ can be written as:
\begin{equation}
	\mathbf{v}_c^n(k+1) = \mathbf{v}_c^n(k) + \Delta t \mathbf{a}^n(k).
\end{equation}
Similarly, the position of the MD $\mathbf{p}_c^n$ can be written as:
\begin{equation}
	\mathbf{p}_c^n(k+1) = \mathbf{p}_c^n(k) + \Delta t \mathbf{v}_c^n(k) + \frac{\Delta t ^2}{2} \mathbf{a}^n(k).
\end{equation}

Therefore, assuming the initial conditions $\mathbf{p}_c^n(0)$, $\mathbf{v}_c^n(0)$, and $\mathbf{C}_b^n(0)$ are known, the angular velocity $\omega_{nb}^b$ and the acceleration $\mathbf{a}^b$, measured using the IMU, can be used to estimate the position, velocity, and orientation of the MD at each time step.

\subsection{Position and Orientation Estimation Using Riemannian Optimization}
\begin{figure}[h!]
\centering
\includegraphics[scale=0.4]{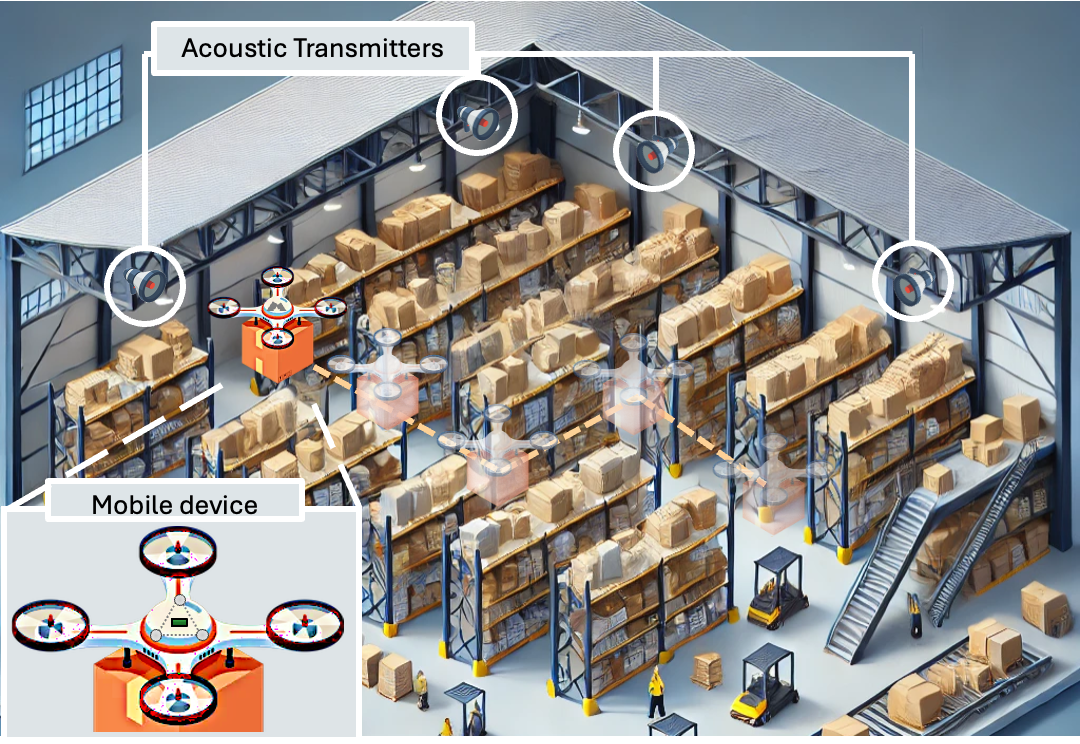}
\caption{Application for indoor localization} 
\label{usp}
\end{figure}

This section summarizes our previous work on Riemannian optimization to estimate the centroid position and orientation of the MD \cite{9528945}. The Riemannian optimization algorithm described in \cite{9528945} is used to determine the positions of three receivers arranged as vertices of an isosceles triangle. In this paper, we specifically consider the case where the triangle is equilateral, meaning all sides, including the base, are of equal length, denoted as $d$. The localization system includes three acoustic receivers arranged in an equilateral triangle rigidly attached to the MD, and four acoustic transmitters, or beacons, with known positions. Figure \ref{usp} illustrates this localization setup.

The MD's position is defined as the centroid of the equilateral triangle formed by the receivers. Thus, by estimating the 3D positions of the three vertices of this triangle, we can determine the 3D position and orientation of the MD. Let $\mathbf{p}_i \in \mathbb{R}^3$ denote the 3D location of the $i$-th receiver, and let $\mathbf{b}_j \in \mathbb{R}^3$ represent the position of the $j$-th beacon.These positions are organized in a matrix $\mathbf{B} \in \mathbb{R}^{4 \times 3}$, where the $j$-th row of the matrix corresponds to the location of the $j$-th beacon, i.e., $\mathbf{b}^T_j$. The time of flight (TOF), $\tau_{ij}$, between the $j$-th transmitter and the $i$-th receiver can be estimated using correlation-based methods \cite{alsharif2017zadoff} \cite{alsharif2021range}. The distances, $r_{ij}$ between the transmitters and receivers are then calculated by multiplying the TOFs, $\tau_{ij}$, by the speed of sound. Assuming that all the distances $r_{ij}$, are subject to normally distributed noise and accounting for the fixed geometry of the receiver array, the position estimation problem can be formulated as the following constrained optimization problem \cite{9528945}:

{\begin{subequations}
		\label{eq:1_2}
		\begin{align}
			\min_{\mathbf{p}_1,\mathbf{p}_2,\mathbf{p}_3 \in \mathbb{R}^{3}} \ & \sum_{i=1}^3 \Big|\Big|\mathbf{B}\mathbf{p}_i - \frac{1}{2}||\mathbf{p}_i||_2^2 \mathbf{1}_{4} - \mathbf{y}_i\Big|\Big|_2^2 \label{eq:2_2} \\
			\label{eq:3_2} {\rm s.t.\ } & (\mathbf{p}_1-\mathbf{p}_2)^T(\mathbf{p}_2-\mathbf{p}_3) = - d^2 \cos(\frac{\pi}{3})\\
			\label{eq:4_2} & (\mathbf{p}_1-\mathbf{p}_3)^T(\mathbf{p}_2-\mathbf{p}_3) = d^2 \cos(\frac{\pi}{3}),
		\end{align}
\end{subequations}}
where $\mathbf{1}_{4}$ is the all-ones vector of dimension $4$, and $\mathbf{y}_i = \cfrac{1}{2}(\mathbf{b}^2-\mathbf{r}^2_i)$ with $\mathbf{b}^2$ and $\mathbf{r}^2_i$ given by
{\begin{align*}
		\mathbf{b}^2 = \begin{pmatrix}
			||\mathbf{b}_1||_2^2 \\
			||\mathbf{b}_2||_2^2 \\
			||\mathbf{b}_3||_2^2 \\
			||\mathbf{b}_4||_2^2 
		\end{pmatrix}, \mathbf{r}_{i}^2 = \begin{pmatrix}
			r_{i1}^2 \\
			r_{i2}^2 \\
			r_{i3}^2 \\
			r_{i4}^2
		\end{pmatrix}.
\end{align*}}

As demonstrated in our previous work \cite{9528945}, the constraints (\ref{eq:3_2}) and (\ref{eq:4_2}) fix the length of the base $||\mathbf{p}_2 - \mathbf{p}_3||_2 = d$ and ensure that the lengths of the two side legs of the triangle are equal, i.e., $||\mathbf{p}_1 - \mathbf{p}_2||_2 = ||\mathbf{p}_1 - \mathbf{p}_3||_2$. We showed that the set of points satisfying constraints (\ref{eq:3_2}) and (\ref{eq:4_2}) forms a Riemannian manifold \cite{9528945}. Therefore, each feasible solution $\mathbf{p}_1,\mathbf{p}_2,\mathbf{p}_3 \in \mathbb{R}^{3}$ to (\ref{eq:1_2}) belongs to a set, named the \emph{{isosceles} triangle manifold}, which is defined as follows \cite{9528945}:
{\begin{align*}
		\mathcal{M} = \Big\{\{\mathbf{p}_i\}_{i=1}^3 \in \mathbb{R}^{3} \ \Big| (\mathbf{p}_1-\mathbf{p}_2)^T(\mathbf{p}_2-\mathbf{p}_3) &= -d^2 \cos(\frac{\pi}{3}) \\
		(\mathbf{p}_1-\mathbf{p}_3)^T(\mathbf{p}_2-\mathbf{p}_3) &= d^2 \cos(\frac{\pi}{3})\Big\}. 
\end{align*}}

Our algorithm in \cite{9528945} efficiently addresses the constrained optimization problem using Riemannian optimization techniques. This section provides a summary of the first- and second-order Riemannian positioning algorithms, along with a straightforward method for determining the orientation of the MD based on the estimated positions. For further details on the algorithm, please refer to our previous work \cite{9528945}.\par 

At each point $\mathbf{P} =[\mathbf{p}_1,\mathbf{p}_2,\mathbf{p}_3] \in \mathcal{M}$, the manifold can be locally approximated by a linear space called the \emph{tangent space}, denoted by $\mathcal{T}_{\mathbf{P}} \mathcal{M}$. To perform optimization on a manifold $\mathcal{M}$, it is necessary to define the notion of length that applies to tangent vectors by equipping each tangent space with an inner product known as the Riemannian metric. In our Riemannian positioning algorithm \cite{9528945}, we use the canonical inner product for matrix space, $\langle \mathbf{X},\mathbf{Y}\rangle = \text{Tr}(\mathbf{X}^{\text{T}}\mathbf{Y})$.\par

Riemannian optimization begins with a feasible point  $\mathbf{P} \in \mathcal{M}$. At each iteration, it determines a descent direction $\xi_{\mathbf{P}}$ in the tangent space $\mathcal{T}_{\mathbf{P}} \mathcal{M}$ and a step size $\eta$ to update the point $\mathbf{P}$. The tangent space $\mathcal{T}_{\mathbf{P}} \mathcal{M}$ provides a local linear approximation of the manifold’s curvature around $\mathbf{P}$. Descent directions in this tangent space can be found using methods similar to those in unconstrained optimization due to its linear nature. However, since the Euclidean gradient ${\nabla}_{\mathbf{P}} f$ and Hessian ${\nabla}^2_{\mathbf{P}} f$ are defined on the original high-dimensional space $\mathcal{E}$ and not necessarily on $\mathcal{T}_{\mathbf{P}} \mathcal{M}$, we need to define the Riemannian gradient $\overline{\nabla}_{\mathbf{P}} f$ and Hessian $\overline{\nabla}^2_{\mathbf{P}} f$  \cite{9528945}. The Riemannian gradient $\overline{\nabla}_{\mathbf{P}} f$ is the projection of the Euclidean gradient ${\nabla}_{\mathbf{P}} f$ onto the tangent space $\mathcal{T}_{\mathbf{P}} \mathcal{M}$. Similarly, the Riemannian Hessian is the orthogonal projection of the directional derivative of the Riemannian gradient onto the tangent space. After determining a descent direction $\xi_{\mathbf{P}}$, which is a tangent vector in the tangent space $\mathcal{T}_{\mathbf{P}} \mathcal{M}$, and a step size $\eta$, the current point is updated using $\mathbf{Z} = \mathbf{P} + \eta \xi_{\mathbf{P}}$. This update may yield a point outside the manifold, i.e, $\mathbf{Z} \in \mathbb{R}^{3 \times 3}$. Therefore, it is necessary to "project" this update back onto the manifold while preserving its descent property, a process achieved using a \emph{retraction} \cite{absil2009optimization}.\par

 First-order Riemannian optimization algorithms require only gradient information to determine the descent direction, while second-order algorithms also require the Riemannian Hessian. For instance, in the steepest descent method on Riemannian manifolds, the descent direction is given by $\xi_{\mathbf{P}} = - {\overline{\nabla}_{\mathbf{P}} f}/{||\overline{\nabla}_{\mathbf{P}} f||_{\mathbf{P}}}$ {\cite{absil2009optimization}}. In contrast, Newton’s method on Riemannian manifolds, a second-order algorithm, determines the descent direction $\xi_{\mathbf{P}} \in \mathcal{T}_{\mathbf{P}} \mathcal{M}$ by solving the Newton's equation $\overline{\nabla}^2_{\mathbf{P}} f[\xi_{\mathbf{P}}] = -\overline{\nabla}_{\mathbf{P}} f$.

\subsubsection{Riemannian-Steepest-Descent-based Positioning Algorithm}
The algorithm begins with an initial point $\mathbf{P}= \mathbf{P}_0 \in \mathcal{M}$. It then iterates between determining a search direction and updating the current position. For two real numbers $\alpha$ and $\beta$, we define a matrix $\mathbf{U}_{\alpha}^{\beta}$ and a matrix $\mathbf{S}$ as follows:
\begin{align}
	\mathbf{U}_{\alpha}^{\beta} &= \begin{pmatrix}
		0 & \alpha+\beta & -\alpha-\beta \\
		\alpha+\beta & -2\alpha & \alpha-\beta \\ 
		-\alpha-\beta & \alpha-\beta & 2 \beta
	\end{pmatrix} \nonumber,\\
 \nonumber \\
	\mathbf{S} &= \begin{pmatrix}
		|| \mathbf{P} \mathbf{U}_{1}^{0} ||^2_{\mathbf{P}} & \langle \mathbf{P} \mathbf{U}_{0}^{1}, \mathbf{P} \mathbf{U}_{1}^{0} \rangle_{\mathbf{P}} \\
		\langle \mathbf{P} \mathbf{U}_{1}^{0}, \mathbf{P} \mathbf{U}_{0}^{1} \rangle_{\mathbf{P}} & || \mathbf{P} \mathbf{U}_{0}^{1} ||^2_{\mathbf{P}}
	\end{pmatrix},
	\label{S_mat2}
\end{align}
where $\langle.,.\rangle_{\mathbf{P}}$ denotes the inner product for matrices, and $||\mathbf{P}||^2_{\mathbf{P}}=\langle\mathbf{P},\mathbf{P}\rangle_{\mathbf{P}}$.  The search direction is given by $\xi_{\mathbf{P}} = - {\overline{\nabla}_{\mathbf{P}} f}/{||\overline{\nabla}_{\mathbf{P}} f||_{\mathbf{P}}}$, where the Riemannian gradient $\overline{\nabla}_{\mathbf{P}}f$ is computed using the following equation:
\begin{align} \label{eq:13_2}
	\overline{\nabla}_{\mathbf{P}}f = \nabla_{\mathbf{P}}f - \mathbf{P} \mathbf{U}_{\alpha}^{\beta}
\end{align}
with the reals $\alpha$ and $\beta$ are determined as the solution to the linear system:
\begin{align} 
	&\mathbf{S} \begin{pmatrix}
		\alpha \\ \beta
	\end{pmatrix} =  \begin{pmatrix}
		\langle \nabla_{\mathbf{P}}f, \mathbf{P} \mathbf{U}_{1}^{0} \rangle_{\mathbf{P}} \\
		\langle \nabla_{\mathbf{P}}f, \mathbf{P} \mathbf{U}_{0}^{1}\rangle_{\mathbf{P}}
	\end{pmatrix}. 
	\label{alpha_beta_grad_2}
\end{align}
The step size $\eta$ is then chosen via backtracking to satisfy the Wolfe conditions \cite{wolfe1969convergence}. Then, the point is updated using the step size and descent direction $\mathbf{Z} =[\mathbf{z}_1,\mathbf{z}_2,\mathbf{z}_3] = \mathbf{P} + \eta \mathbf{\xi}_{\mathbf{P}}$. The tangent vector $\xi_{\mathbf{P}}$, scaled by the step size $\eta$, is retracted onto the manifold using the following retraction formula:
{\begin{align} \label{eq:16_2}
		\text{R}_{\mathbf{P}}(\xi_{\mathbf{P}}) = \lambda \begin{pmatrix}
			\gamma \mathbf{z}_1  \\
			\mathbf{z}_2  \\
			\mathbf{z}_3
		\end{pmatrix},
	\end{align} 
	where $\lambda$ and $\gamma$ are defined as: 
	{\begin{align}
			\lambda &= \sqrt{\cfrac{d^2 \cos(\frac{\pi}{3})}{\left(\gamma \mathbf{z}_1-\mathbf{z}_3\right)^T(\mathbf{z}_2-\mathbf{z}_3)}} \nonumber \\
			\nonumber \\
			\gamma &= \cfrac{(\mathbf{z}_2+\mathbf{z}_3)^T(\mathbf{z}_2-\mathbf{z}_3)}{2 \mathbf{z}_1^T (\mathbf{z}_2-\mathbf{z}_3)}.
			\label{gamma_eq_2}
	\end{align}}
	Define the isosceles triangle $\mathbf{U}$ as $[\mathbf{u}_1,\mathbf{u}_2,\mathbf{u}_3]=[\gamma \mathbf{z}_1,\mathbf{z}_2,\mathbf{z}_3]$. To obtain the point on the manifold, scale the sides of $\mathbf{U}$ using:
 \begin{align*}
     \mathbf{P} = \sqrt{\cfrac{d^2 \cos(\frac{\pi}{3})}{(\mathbf{u}_1-\mathbf{u}_3)^T(\mathbf{u}_2-\mathbf{u}_3)}} \ \mathbf{U}.
 \end{align*}
	The issues with poles in (\ref{gamma_eq_2}) are addressed using a simple algorithm detailed in \cite{9528945}.
	\vspace{0.0cm}
	\subsubsection{Riemmanian Trust Region Localization Algorithm} The Riemannian trust region method on the manifold $\mathcal{M}$ for a cost function $f$ involves adding to the current iterate $\mathbf{P} \in \mathcal{M}$ the update vector $\mathbf{\xi}_\mathbf{P} \in \mathcal{T}_\mathbf{P} \mathcal{M}$, obtained by solving the following trust-region subproblem on $\mathcal{T}_{\mathbf{P}_k} \mathcal{M}$ \cite{absil2007trust} :
	\begin{align}
		\min_{\mathbf{\xi} \in \mathcal{T}_{\mathbf{P}_k} \mathcal{M}} m_{\mathbf{P}_k} (\mathbf{\xi}) &= f(\mathbf{P}_k) + \langle \overline{\nabla}_{\mathbf{P}_k}f(\mathbf{P}_k),\mathbf{\xi}\rangle + \frac{1}{2} \langle\overline{\nabla}_{\mathbf{P}_k}^2f(\mathbf{\xi}),\mathbf{\xi}\rangle \;\; \\ 
		&\text{s.t.} \;\; ||\xi||_{\mathbf{P}_k}\le \Delta_k^2, \nonumber
	\end{align}
	where $\Delta_k$ is the trust region radius, and ${\nabla}_{\mathbf{P}_k}^2f$ is the Riemannian hessian which can be obtained using the following equations:
	\begin{align} \label{eq:17_2}
		&\overline{\nabla}^2_{\mathbf{P}_k}f[\xi_{\mathbf{P}_k}] =  \Pi_{\mathbf{P}_k}(\nabla^2_{\mathbf{P}_k}f[\xi_{\mathbf{P}_k}] - \xi_{\mathbf{P}_k} \mathbf{U}_{\alpha}^{\beta} - \mathbf{P}_k \mathbf{U}_{\dot{\alpha}}^{\dot{\beta}}), 
	\end{align}
	where the orthogonal projection $\Pi_{\mathbf{P}_k}$ is given by:
 \begin{align*}    
 \Pi_{\mathbf{P}}(\mathbf{Z})= \mathbf{Z} - \mathbf{P} \mathbf{U}_{\alpha}^{\beta}.
 \end{align*}
 The parameters $\alpha$ and $\beta$ are the solution to \ref{alpha_beta_grad_2}, where $\mathbf{S}$ is defined in \ref{S_mat2}. Their directional derivatives $\dot{\alpha}$ and $\dot{\beta}$ are obtained as the solution to the system: 
	\begin{small}
	\begin{align*}
		\mathbf{S} \begin{pmatrix} \dot{\alpha} \\ \dot{\beta} \end{pmatrix} = &\begin{pmatrix}
			\langle \nabla^2_{\mathbf{P}_k}f[\xi_{\mathbf{P}_k}], \mathbf{P}_k \mathbf{U}_{1}^{0} \rangle_{\mathbf{P}_k} + \langle\nabla_{\mathbf{P}_k}f, \xi_{\mathbf{P}_k} \mathbf{U}_{1}^{0} \rangle_{\mathbf{P}_k} \\
			\langle \nabla^2_{\mathbf{P}_k}f[\xi_{\mathbf{P}_k}], \mathbf{P}_k \mathbf{U}_{0}^{1}\rangle_{\mathbf{P}_k} +  \langle \nabla_{\mathbf{P}_k}f, \xi_{\mathbf{P}_k} \mathbf{U}_{0}^{1}\rangle_{\mathbf{P}_k}
		\end{pmatrix} \\
  &- \dot{\mathbf{S}} \begin{pmatrix} \alpha \\ \beta \end{pmatrix}.
	\end{align*}
	\end{small}
	The directional derivative of $\mathbf{S}$ in the direction $\xi_{\mathbf{P}_k}$ is given by:
	\begin{align*}
		\dot{\mathbf{S}} = 
		\left(
		\begin{array}{c|c}
			\multirow{2}{*}{$2\langle \xi_{\mathbf{P}_k} \mathbf{U}_{1}^{0}, \mathbf{P}_k \mathbf{U}_{1}^{0} \rangle_{\mathbf{P}_k}$} & \langle \xi_{\mathbf{P}_k} \mathbf{U}_{0}^{1}, \mathbf{P}_k \mathbf{U}_{1}^{0} \rangle_{\mathbf{P}_k}  \\
			& \quad + \langle \mathbf{P}_k \mathbf{U}_{0}^{1}, \xi_{\mathbf{P}_k} \mathbf{U}_{1}^{0} \rangle_{\mathbf{P}_k} \\  
			\hline 
			\langle \xi_{\mathbf{P}_k} \mathbf{U}_{1}^{0}, \mathbf{P}_k \mathbf{U}_{0}^{1} \rangle_{\mathbf{P}_k}  & \multirow{2}{*}{$2\langle \xi_{\mathbf{P}_k} \mathbf{U}_{0}^{1},\mathbf{P}_k \mathbf{U}_{0}^{1}\rangle_{\mathbf{P}_k}$}\\
			\quad+ \langle \mathbf{P}_k \mathbf{U}_{1}^{0}, \xi_{\mathbf{P}_k} \mathbf{U}_{0}^{1} \rangle_{\mathbf{P}_k} & \\
		\end{array}
		\right).
	\end{align*}
	
	The trust region subproblem is solved using any of the available methods to obtain $\mathbf{\xi}_k$. The new iterate is then computed by retracting $\mathbf{\xi}_k$ onto the manifold, i.e., $\mathbf{P}_+ = \mathbf{R}_{\mathbf{P}_k}(\mathbf{\xi}_k)$. To decide whether to accept or reject the new iterate and to update the trust region radius, the following quotient is used:
	\begin{equation}
		\rho_k = \frac{f(\mathbf{P}_k)-f(\mathbf{R}_{\mathbf{P}_k}(\mathbf{\xi}_k))}{m_{\mathbf{P}_k} (\mathbf{0}_{\mathbf{P}_k})-m_{\mathbf{P}_k} (\mathbf{\xi}_k)}.
	\end{equation}
	
	Here, $\rho_k$ is compared against predefined thresholds to determine the next steps. If $\rho_k$ is very small, then the new iterate should be rejected, and the trust region radius must be reduced. If $\rho_k$ is small but not too small, the new iterate should be accepted and the trust radius should be kept the same. If $\rho_k$ is close to 1, then the new iterate should be accepted and the radius can be increased. 
	\subsubsection{Centroid Position and Orientation of the MD}
    The centroid position and orientation of the MD can be determined using the estimated positions of the vertices of the equilateral triangle. The centroid position is calculated by averaging the positions of the vertices:
	\begin{equation}
		\mathbf{p}_c^R = \sum_{i=1}^{3}\frac{\mathbf{p}_i}{3}.
	\end{equation}
Here, the superscript $R$ in $\mathbf{p}_c^R$ distinguishes the centroid position obtained using the Riemannian algorithm from the position obtained using an INS. The orientation matrix $\mathbf{C}_b^n$ is determined by computing the body frame axes $\mathbf{C}_b^n = [O_{{x}_{b}}^{n}, O_{{y}_{b}}^{n}, O_{{z}_{b}}^{n}]$, expressed in the navigation frame as follows: 
	\begin{align}
		O_{{x}_{b}}^n &= \frac{\mathbf{p}_1 -\mathbf{p}_c^R}{||\mathbf{p}_1 -\mathbf{p}_c^R||}\\
		O_{{y}_{b}}^{n'} &= \mathbf{p}_2 -\mathbf{p}_c^R - <\mathbf{p}_2 -\mathbf{p}_c^R,O_{{x}_{b}}^n> O_{{x}_{b}}^n\\ 
		O_{{y}_{b}}^n &= \frac{O_{{y}_{b}}^{n'}}{||O_{{y}_{b}}^{n'}||}\\
		O_{{z}}^{n} &= O_{{x}_{b}}^{n} \times O_{{y}_{b}}^{n}.
	\end{align}

	Finally, as shown in Figure \ref{md}, the centroid position can be transformed into the vertices' positions using the orientation matrix and the following equations: 
	\begin{align}
		\mathbf{P} &= [\mathbf{p}_c^R,\mathbf{p}_c^R,\mathbf{p}_c^R] + \mathbf{C}_b^n\begin{bmatrix}
			\frac{\sqrt{3}d}{3} & 	\frac{-\sqrt{3}d}{6} & \frac{-\sqrt{3}d}{6}\\ 
			0 &  \frac{d}{2}  & -\frac{d}{2}  \\ 
			0 & 0 &  0
		\end{bmatrix}.
		\label{pc2pi}
	\end{align}

	\section{Fusion of INS and Acoustic Position and Orientation Using Kalman Filter }
	\label{fusion}
 This section introduces a new formulation of the state equations that incorporates the orientation matrix as part of the state vector, leading to new EKF equations. Additionally, a novel retraction operator is presented, which projects the output of the EKF and UKF onto the isosceles triangle manifold, thereby enhancing the accuracy of tracking the target's position and orientation.\par 
	Let $\mathbf{x} = [\mathbf{p}_c^T, \mathbf{v}_c^T , \mathbf{c}^T]^T \in \mathbb{R}^{15}$ denote the state vector to be estimated. Here, $\mathbf{v}_c$ represents the velocity of the center of the MD, and $\mathbf{c}$ is the vectorized orientation matrix $\mathbf{C}_b^n$. It is established that the linear Kalman filter is optimal for state estimation when the IMU and acoustic sensor noises are independent and normally distributed with zero mean \cite{jazwinski2007stochastic}. However, since the state equations are nonlinear in this case, we compare the performance of the EKF and UKF for estimating the MD's position and orientation. The state equations are given by:
	
	\begin{footnotesize}
	\begin{align}
		\mathbf{x}(k+1) &=\underset{\mathbf{F(\mathbf{a}_b,\mathbf{\Omega}_{nb}^b)}}{\underbrace{\begin{bmatrix}
			\mathbf{I}_{3\times 3} & \Delta t \mathbf{I}_{3\times 3}& \frac{\Delta t^2}{2} \mathbf{a}_b^\times \\
			\mathbf{0}_{3\times 3} & \mathbf{I}_{3\times 3} & \Delta t \mathbf{a}_b^\times + \frac{\Delta t^2}{2} (\mathbf{a}_b^\times)({\mathbf{\Omega}_{nb}^b}^\times)\\
			\mathbf{0}_{9\times 3} &	\mathbf{0}_{9\times 3} & 	\mathbf{I}_{9\times 9}  + \Delta t ({\mathbf{\Omega}_{nb}^b}^\times) + \frac{\Delta t^2}{2} ({\mathbf{\Omega}_{nb}^2}^\times)
		\end{bmatrix}}} \mathbf{x}(k),
	\end{align} 
	\end{footnotesize}
	where $\mathbf{a}_b^\times = \mathbf{a}_b^T \bigotimes \mathbf{I}_{3\times 3}$ is the Kronecker product of the acceleration vector $\mathbf{a}_b^T$ and the identity matrix $\mathbf{I}_{3 \times 3}$, ${\mathbf{\Omega}_{nb}^b}^\times = {\mathbf{\Omega}_{nb}^b}^T\bigotimes \mathbf{I}_{3\times 3}$ is the Kronecker product of ${\mathbf{\Omega}_{nb}^b}^T$ and the identity matrix, and $({\mathbf{\Omega}_{nb}^2}^\times) = ({\mathbf{\Omega}_{nb}^b}\bigotimes \mathbf{I}_{3\times 3})({\mathbf{\Omega}_{nb}^b}\bigotimes \mathbf{I}_{3\times 3})$. 
Note that in the derivations throughout this section, the acceleration $\mathbf{a}_b$ is assumed to be free acceleration (gravity component subtracted).\par
 The nonlinearity in the state-transition matrix $\mathbf{F}(\mathbf{a}_b,\mathbf{\Omega}_{nb}^b)$ arises from the multiplication of the inputs, acceleration $\mathbf{a}_b$ and angular velocity $\mathbf{\omega}_b$, by the state vector. The input noise can be modeled as additive Gaussian noise: 
	\begin{align}
	    \mathbf{u}(k) &= \begin{bmatrix}
	    \mathbf{a}_b (k) \\
	   \mathbf{\omega}_b (k) 
	    \end{bmatrix} = \Tilde{\mathbf{u}}(k) + \mathbf{n}(k)
	\end{align}
	where $\Tilde{\mathbf{u}}(k)$ is the noiseless input, i.e. the actual acceleration and angular velocity of the MD, and $\mathbf{n}(k)$ is the input noise, produced by the IMU's electronic components, normally distributed with zero mean and covariance matrix $\mathbf{Q} = \text{diag}(\sigma_{\mathbf{a}_x}^2,\sigma_{\mathbf{a}_y}^2,\sigma_{\mathbf{a}_z}^2,\sigma_{\mathbf{\omega}_x}^2,\sigma_{\mathbf{\omega}_y}^2,\sigma_{\mathbf{\omega}_z}^2)$.\par 
 Let the measurement vector $\mathbf{q} = [\mathbf{p}_1^T, \mathbf{p}_2^T, \mathbf{p}_3^T]^T \in \mathbb{R}^{9}$ represent the positions of the vertices, estimated using the Riemannian localization method. The measurement equation is:  
	\begin{align}
		\mathbf{q}(k+1) &= \mathbf{H}\mathbf{x}(k+1) + \mathbf{\nu}(k+1),
	\end{align} 
	where the measurement (observation) matrix $\mathbf{H}$ is:
	\begin{align}
		\mathbf{H} &= \begin{bmatrix}
			\mathbf{I}_{3\times 3} & \mathbf{0}_{3\times 3} & \mathbf{d}_1^T \bigotimes \mathbf{I}_{3\times 3} \\
			\mathbf{I}_{3\times 3} & \mathbf{0}_{3\times 3} & \mathbf{d}_2^T \bigotimes \mathbf{I}_{3\times 3} \\
			\mathbf{I}_{3\times 3} & \mathbf{0}_{3\times 3} & \mathbf{d}_3^T \bigotimes \mathbf{I}_{3\times 3} \\ 
		\end{bmatrix},
	\end{align}
	where $\mathbf{d}_i$ represents the vector from the centroid of the triangle to the $i^{\text{th}}$ vertex, and $\mathbf{\nu}(k)$ is the measurement noise, normally distributed with zero mean and covariance matrix $\mathbf{R}_k = \text{diag}(\sigma_{\mathbf{p}_{1_x}}^2,\sigma_{\mathbf{p}_{1_y}}^2,...,\sigma_{\mathbf{p}_{3_z}}^2)$. While the measurement equation is linear, the process equation is nonlinear. Therefore, the linear Kalman filter is not optimal. This paper evaluates the performance of the EKF and UKF in estimating the position and orientation of the MD.
	
	\subsection{Extended Kalman Filter}
 The EKF operates in two main steps; prediction and correction. Given an initial estimate of the state $\mathbf{x}(0)$, the state equations are used to predict the state and its covariance matrix as follows:
 	\begin{align}
		\bar{\mathbf{x}}(k+1) &= \mathbf{F}(\mathbf{a}_b,\mathbf{\omega}_b)\hat{\mathbf{x}}(k) \\
		\bar{\mathbf{P}}(k+1) &= \mathbf{F}(\mathbf{a}_b,\mathbf{\omega}_b)\hat{\mathbf{P}}(k)\mathbf{F}^T(\mathbf{a}_b,\mathbf{\omega}_b) + \mathbf{F}_\mathbf{u}\mathbf{Q}_{k}\mathbf{F}^T_\mathbf{u},
	\end{align}
	where $\bar{\mathbf{x}}(k+1)$ and $\bar{\mathbf{P}}(k+1)$ represent the predicted state and its covariance matrix at time step $k+1$, while $\hat{\mathbf{x}}(k)$ and $\hat{\mathbf{P}}(k)$ denote the corrected state and covariance matrix at time step $k$. $\mathbf{F}_\mathbf{u}$ is the Jacobian matrix of $\mathbf{F}(\mathbf{a}_b,\mathbf{\omega}_b)$ with respect to $\mathbf{p}_c$, $\mathbf{v}_c$, and $\mathbf{c}$. The Jacobian matrix $\mathbf{F}_\mathbf{u}$ is given by:

	\begin{align}
	    \mathbf{F}_\mathbf{u} &= \begin{bmatrix}
	    \nabla_{\mathbf{a}_b}\mathbf{p}_c^T & \nabla_{\mathbf{\omega}_b} \mathbf{p}_c^T \\
	    \nabla_{\mathbf{a}_b} \mathbf{v}_c^T & \nabla_{\mathbf{\omega}_b} \mathbf{v}_c^T \\
	    \nabla_{\mathbf{a}_b} \mathbf{c}^T & \nabla_{\mathbf{\omega}_b} \mathbf{c}^T \\
	    \end{bmatrix}, 
	\end{align}

	where the elements of $  \mathbf{F}_\mathbf{u}$ are calculated as follows:
\begin{align}
	\nabla_{\mathbf{a}_b}\mathbf{p}_c^T &= \frac{\Delta t^2}{2} \mathbf{C}_b^n, \;\;\;\;\;\; \nabla_{\mathbf{\omega}_b}\mathbf{p}_c^T = \mathbf{0}_{3 \times 3}, \;\;\;\;\;\; \nabla_{\mathbf{a}_b}\mathbf{c}^T = \mathbf{0}_{9 \times 3},\\
	 \nabla_{\mathbf{\omega}_b}\mathbf{c}^T &= \Delta t [\mathbf{E}_1^{\times^T} \mathbf{c}, \; \mathbf{E}_2^{\times^T} \mathbf{c}, \; \mathbf{E}_3^{\times^T} \mathbf{c}] +\\
	   &\frac{\Delta t^2}{2} [({_x\mathbf{\Omega}_{nb}^2}^\times)\mathbf{c}, \; ({_y\mathbf{\Omega}_{nb}^2}^\times)\mathbf{c}, \; ({_z\mathbf{\Omega}_{nb}^2}^\times)\mathbf{c}], \\
	   \nabla_{\mathbf{a}_b} \mathbf{v}_c^T &= \Delta t \mathbf{C}_b^n +  \frac{\Delta t^2}{2} \mathbf{C}_b^n \mathbf{\Omega}_{nb}^b, \\ 
	   \nabla_{\mathbf{\omega}_b} \mathbf{v}_c^T &=\frac{\Delta t^2}{2} \mathbf{C}_b^n \mathbf{\Xi}_b^T, 
\end{align}
	where the $\mathbf{A}^\times$ denotes the Kronecker product of $\mathbf{A}$ with the identity matrix $\mathbf{I}_{3 \times 3}$, $\mathbf{e}_1=[1,0,0]$, $\mathbf{e}_2=[0,1,0]$, $\mathbf{e}_3=[0,0,1]$, and $\mathbf{E}_i$ is the skew-symmetric matrix of $\mathbf{e}_i$ where $i=1,2,3$. Moreover, the matrices $_x\mathbf{\Omega}_{nb}^2$, $_y\mathbf{\Omega}_{nb}^2$, $_z\mathbf{\Omega}_{nb}^2$, and $\mathbf{\Xi}_b^T$ are given by:
\begin{align}
	    _x\mathbf{\Omega}_{nb}^2 = \frac{\partial}{\partial \omega_x}(\mathbf{\Omega}_{nb}^2)&= \begin{bmatrix}
	    0 & \omega_y & \omega_z \\
	    \omega_y  & -2 \omega_x & 0 \\
	     \omega_z & 0 & -2 \omega_x \\
	    \end{bmatrix},\\
	    _y\mathbf{\Omega}_{nb}^2 = \frac{\partial}{\partial \omega_y}(\mathbf{\Omega}_{nb}^2)&= \begin{bmatrix}
	    -2 \omega_y & \omega_x & 0 \\
	    \omega_x  & 0 & \omega_z \\
	     0 & \omega_z & -2 \omega_y \\
	    \end{bmatrix}, \\
	    _z\mathbf{\Omega}_{nb}^2 = \frac{\partial}{\partial \omega_z}(\mathbf{\Omega}_{nb}^2)&= \begin{bmatrix}
	    -2 \omega_z & 0 & \omega_x \\
	    0  & -2 \omega_z & \omega_y \\
	     \omega_x & \omega_y & 0 \\
	    \end{bmatrix},\\
     	    \mathbf{\Xi}_b^T &= \begin{bmatrix}
	   0 & {\mathbf{a}_b}_z & -{\mathbf{a}_b}_y \\
	    -{\mathbf{a}_b}_z  & 0 & {\mathbf{a}_b}_x \\
	     {\mathbf{a}_b}_y & -{\mathbf{a}_b}_x & 0 \\
	    \end{bmatrix}.
	\end{align}
	
	To correct the predicted state and covariance matrix at time step $k+1$, compute the Kalman gain $\mathbf{K}_{k+1}$ as follows:
	\begin{equation}
		\mathbf{K}_{k+1} = \bar{\mathbf{P}}(k+1) \mathbf{H}^T(\mathbf{H} \bar{\mathbf{P}}(k+1)\mathbf{H}^T + \mathbf{R}_{k+1})^{-1}
	\end{equation}
	Finally, update the corrected state and covariance matrix using: 
	\begin{align}
		\hat{\mathbf{x}}(k+1) &= \bar{\mathbf{x}}(k+1) + \mathbf{K}_{k+1} (\mathbf{q}(k+1)-\mathbf{H}\bar{\mathbf{x}}(k+1))\\
		\hat{\mathbf{P}}(k+1) &= (\mathbf{I}-\mathbf{K}_{k+1}\mathbf{H})\bar{\mathbf{P}}(k+1).
	\end{align}
	
	\subsection{Unscented Kalman Filter}
 EKF uses a first-order approximation of the nonlinear state equations, which introduces linearization errors. In contrast, UKF avoids linearization. Instead, the UKF estimates the mean and covariance of the state variable through a process known as the unscented transform. The UKF employs a deterministic sampling method to represent the state vector's distribution using a set of carefully chosen samples called sigma points. These sigma points capture the mean and covariance of the state vector, and after being propagated through the nonlinear transformation, they provide estimates of {\textit{a posteriori}} mean and covariance \cite{wan2000unscented}.\par 
The unscented transform (UT) is an algorithm used to compute the mean and covariance of a random variable undergoing a nonlinear transformation \cite{wan2000unscented}. Given a state vector $\mathbf{x}$ of dimension $15$, $31$ sigma points $\mathbf{\chi}_i$ are computed \cite{wan2000unscented}. Let $\mathbf{\mu}_\mathbf{x}$ be the mean of the state vector and $\mathbf{P}_\mathbf{x}$ its covariance matrix. The sigma points are calculated as follows:
\begin{align}
    \mathbf{\chi}^{(0)} &= \mathbf{\mu}_\mathbf{x},\\
    \mathbf{\chi}^{(i)} &= \mathbf{\mu}_\mathbf{x} + \sqrt{15+\kappa} \; \text{Col}_i \mathbf{L}, \;\; i=1,2,..., 15,\\
    \mathbf{\chi}^{(i+15)} &= \mathbf{\mu}_\mathbf{x} - \sqrt{15+\kappa}\; \text{Col}_i \mathbf{L}, \;\; i=1,2,..., 15,
\end{align}
where $\mathbf{L}$ is a lower triangular matrix obtained from the Cholesky decomposition of $\mathbf{P}_\mathbf{x}$, i.e., $ \mathbf{L}\mathbf{L}^T = \mathbf{P}_\mathbf{x}$. The parameter $\kappa = (\alpha^2-1)\times 15$ is a scaling factor where $\alpha$ determines the spread of the sigma points around the mean, and $\text{Col}_i \mathbf{L}$ represents the $i^\text{th}$ column of the matrix $\mathbf{L}$.
After calculating the sigma points, they are propagated through the state equation:
\begin{align}
    \bar{\mathbf{\chi}}^{(i)} &= \mathbf{F}(\mathbf{a}_b,\mathbf{\omega}_b)\mathbf{\chi}^{(i)}, \;\; i=0,1,...,30.
\end{align}
The predicted mean and covariance of the state vector are then computed using the propagated sigma points:
\begin{align}
    W_m^{(0)} &= \frac{\kappa}{\kappa+15},\;\;\; W_c^{(0)} = \frac{\kappa}{\kappa+15}+3-\alpha^2,\\
    W_m^{(i)} &= W_c^{(i)} = \frac{1}{2(\kappa+15)},\\
    \bar{\mathbf{x}}(k+1)&= \sum_{i=0}^{30} W^{(i)}_m \bar{\mathbf{\chi}}^{(i)}, \\
    \bar{\mathbf{P}}_{\mathbf{x}}(k+1)&= \sum_{i=0}^{30} W^{(i)}_c \Big(\bar{\mathbf{\chi}}^{(i)}-\bar{\mathbf{x}}(k+1)\Big)\Big(\bar{\mathbf{\chi}}^{(i)}-\bar{\mathbf{x}}(k+1)\Big)^T.
\end{align}
Next, the measurements are predicted from the propagated sigma points:
\begin{align}
    \bar{\mathbf{\mathcal{Y}}}^{(i)} &= \mathbf{H} \bar{\mathbf{\chi}}^{(i)}, \;\; i=0,1,...,30. 
\end{align}
The mean and covariance of the measurements are:
\begin{align}
    \bar{\mathbf{q}}(k+1)&= \sum_{i=0}^{30} W^{(i)}_m \bar{\mathbf{\mathcal{Y}}}^{(i)}, \\
    \bar{\mathbf{P}}_{\mathbf{q}}(k+1)&= \sum_{i=0}^{30} W^{(i)}_c \Big(\bar{\mathbf{\mathcal{Y}}}^{(i)}-\bar{\mathbf{q}}(k+1)\Big)\Big(\bar{\mathbf{\mathcal{Y}}}^{(i)}-\bar{\mathbf{q}}(k+1)\Big)^T.
\end{align}
Finally, the cross-covariance and Kalman gain are calculated as follows:
\begin{align}
    \bar{\mathbf{P}}_{\mathbf{x}\mathbf{q}}(k+1)&= \sum_{i=0}^{30} W^{(i)}_c \Big(\bar{\mathbf{\chi}}^{(i)}-\bar{\mathbf{x}}(k+1)\Big)\Big(\bar{\mathbf{\mathcal{Y}}}^{(i)}-\bar{\mathbf{q}}(k+1)\Big)^T.\\
    \mathbf{K}(k+1)&=  \bar{\mathbf{P}}_{\mathbf{x}\mathbf{q}}(k+1)  \bar{\mathbf{P}}_{\mathbf{q}}^{-1}(k+1).
\end{align}
The corrected mean and covariance are then computed as:
\begin{align}
    \hat{\mathbf{x}}(k+1) &= \bar{\mathbf{x}}(k+1) +  \mathbf{K}(k+1)(\mathbf{q}(k+1)-  \bar{\mathbf{q}}(k+1)),\\
    \hat{\mathbf{P}}_\mathbf{x}(k+1) &= \bar{\mathbf{P}}_\mathbf{x}(k+1) - \mathbf{K}(k+1) \hat{\mathbf{P}}_\mathbf{q}(k+1) \mathbf{K}^T(k+1).
\end{align}
	
\subsection{Projection of the Estimated State to the Manifold }
The estimated position and orientation matrix are transformed into the positions of three vertices of a triangle. However, whether using the EKF or the UKF, these vertices positions are not guaranteed to lie on the isosceles triangle manifold. Therefore, the final step in the proposed algorithm is to "project" the estimated vertices positions onto the isosceles triangle manifold. This section introduces a novel and efficient retraction algorithm that projects the vertices obtained from the EKF or UKF onto the isosceles triangle manifold.\par 

The retraction operator described in \cite{9528945} achieves high localization accuracy, even when the initial estimate is far from the optimal solution. However, it assigns more weight to one of the vertices when the term $\mathbf{z}_1^T(\mathbf{z}_2-\mathbf{z}_3)$ in (\ref{gamma_eq_2}) approaches zero, which may not be efficient near the true solution. This section presents a more efficient retraction method for points close to the optimal solution. Given high sampling rates, the output of the EKF or UKF, while not necessarily on the manifold, is assumed to be close to the optimal solution. Hence, an efficient retraction is one that adjusts the EKF or UKF output to the manifold with minimal perturbation.\par 

Let $\mathbf{Z}^k=[\mathbf{z}_1^k,\mathbf{z}_2^k,\mathbf{z}_3^k]$ denote the positions of the vertices obtained using EKF or UKF. The new retraction algorithm is implemented in four steps. First, the centroid of $\mathbf{Z}^k$ is translated to the origin. This is achieved using the translation:    
\begin{align} \label{nr:s1}
    \begin{pmatrix}
	\mathbf{z}_1'  \\
	\mathbf{z}_2'   \\
	\mathbf{z}_3' 
    \end{pmatrix} = \begin{pmatrix}
	\mathbf{z}_1^k-\mathbf{z}_c\\
	\mathbf{z}_2^k-\mathbf{z}_c\\
	\mathbf{z}_3^k-\mathbf{z}_c
    \end{pmatrix},
\end{align}
where $\mathbf{z}_c=({\mathbf{z}_1^k+\mathbf{z}_2^k+\mathbf{z}_3^k})/{3}$. The second step involves scaling the triangle such that the length of the base side is $d$. This scaling is performed as follows:
\begin{align} \label{nr:s2}
    \begin{pmatrix}
	\mathbf{z}_1''  \\
	\mathbf{z}_2''   \\
	\mathbf{z}_3'' 
    \end{pmatrix} = \lambda^k \begin{pmatrix}
	\mathbf{z}_1'\\
	\mathbf{z}_2'\\
	\mathbf{z}_3'
    \end{pmatrix},
\end{align}
where $\lambda^k = {d}/{||\mathbf{z}_2'-\mathbf{z}_3'||}$. In the third step, $\mathbf{z}_1''$ is translated in the direction of the unit vector $\mathbf{u}_z=({\mathbf{z}_3''-\mathbf{z}_2''})/{d}$. The magnitude of this translation vector $\gamma^k$ is chosen such that the translated point $\mathbf{z}_1'''$ falls onto the bisector of the triangle’s base. The result of this operation is an isosceles triangle centered around the origin. The median $[\mathbf{z}_1''\mathbf{z}_{2,3}'']$, from the vertex $\mathbf{z}_1''$ that bisects the base is $\mathbf{z}_1''- ({\mathbf{z}_3''+\mathbf{z}_2''})/{2}$. The magnitude of the translation $\gamma^k$ is simply found as the projection of the median $[\mathbf{z}_1''\mathbf{z}_{2,3}'']$ onto the unit vector $\mathbf{u}_z$:
\begin{align}
    \gamma^k =\Big\langle\mathbf{z}_1''- \frac{\mathbf{z}_3''+\mathbf{z}_2''}{2},\frac{\mathbf{z}_3''-\mathbf{z}_2''}{d}\Big\rangle,
\end{align}
where $\langle\cdot,\cdot\rangle$ is the inner product of vectors. The updated positions of the vertices after this step are:
\begin{align}
    \begin{pmatrix}
        \mathbf{z}_1'''  \\
	\mathbf{z}_2'''   \\
	\mathbf{z}_3''' 
    \end{pmatrix} = \begin{pmatrix}
	\mathbf{z}_1''+\gamma^k \frac{\mathbf{z}_3''-          \mathbf{z}_2''}{d}\\
	\mathbf{z}_2''\\
	\mathbf{z}_3''
    \end{pmatrix},   
\end{align}
An illustration of the third step is shown in Figure \ref{fig:step3}.
\begin{figure}[H]
    \centering
    \includegraphics[width=4cm,height=3cm]{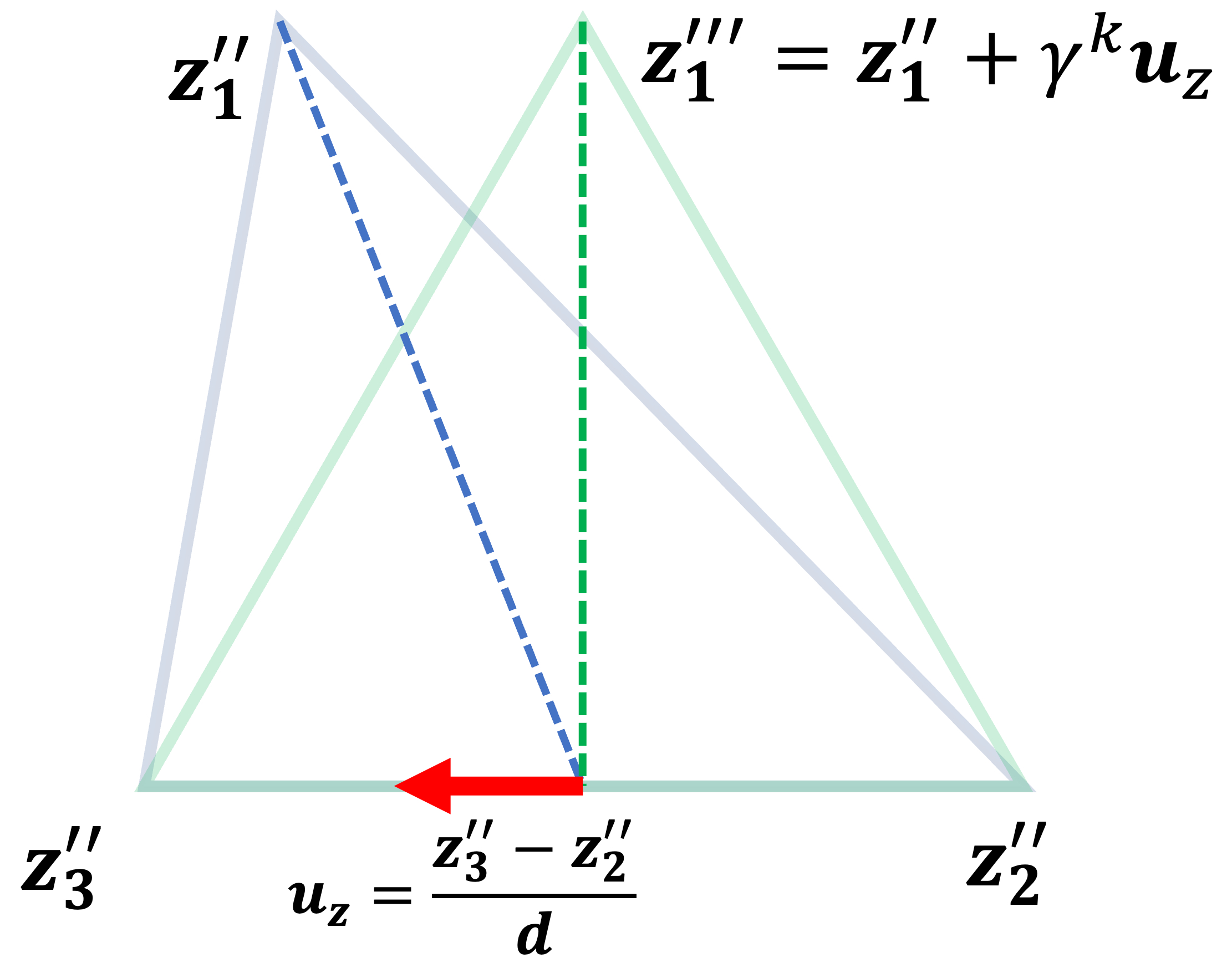}
    \caption{Illustration of step 3 in the new retraction algorithm}
    \label{fig:step3}
\end{figure}
The fourth and final step of the new retraction algorithm involves translating all the points back by the centroid $\mathbf{z}_c$. This ensures that the final positions of the vertices are properly adjusted relative to their original placement. An illustration of all four steps of the new retraction process is provided in Figure \ref{fig:nr}. After completing these steps, the output of the retraction algorithm consists of the three vertices of an isosceles triangle, with the base length $d$. 
\begin{figure}[H]
    \centering
    \includegraphics[width=8.75cm,height=5.0cm]{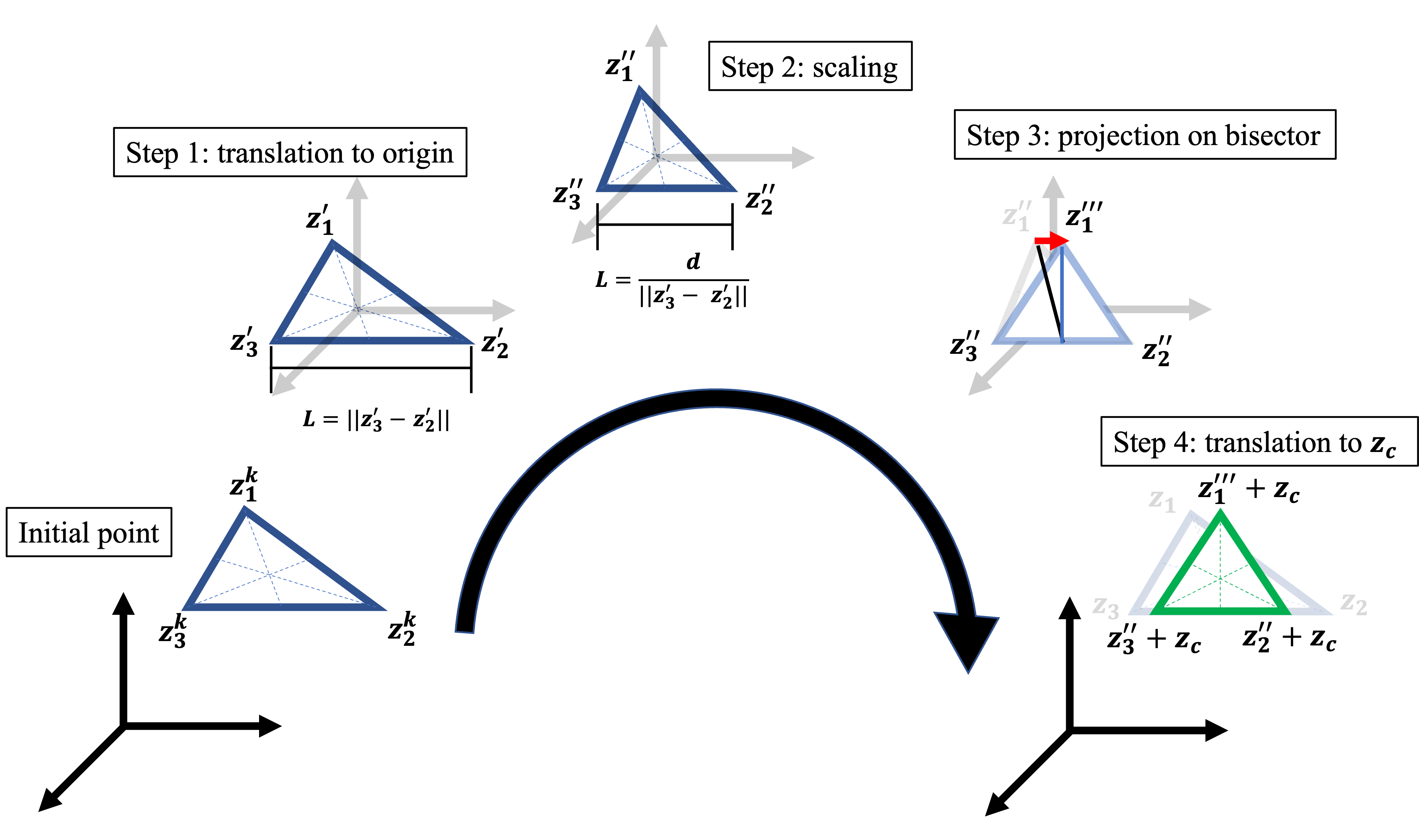}
    \caption{Illustration of the new retraction algorithm}
    \label{fig:nr}
\end{figure}
With straightforward mathematical manipulations, the new retraction process can be simplified as detailed in Algorithm \ref{alg3}.	
\begin{algorithm}[H]
\begin{algorithmic}[1]
\REQUIRE Length $d >0$, initialization $\mathbf{z}_1^k,\mathbf{z}_2^k,\mathbf{z}_3^k$
\vspace{0.2cm}
\STATE Compute the centroid $\mathbf{z}_c = \frac{\mathbf{z}_1^k+\mathbf{z}_2^k+\mathbf{z}_3^k}{3}$ and scaling factor $\lambda^k = \frac{d}{||\mathbf{z}_3^k-\mathbf{z}_2^k||}$
\STATE Compute the parameter $\gamma^k = (\lambda^k)^2 \Big\langle\mathbf{z}_1- \frac{\mathbf{z}_3+\mathbf{z}_2}{2},\frac{\mathbf{z}_3-\mathbf{z}_2}{d}\Big\rangle$
\vspace{0.2cm}
\STATE Compute the projected positions $[\mathbf{p}_1^R,\mathbf{p}_2^R,\mathbf{p}_3^R]$ 
	\begin{align*}
	 	\begin{pmatrix}
		    \mathbf{p}_1^R  \\
			\mathbf{p}_2^R   \\
			\mathbf{p}_3^R 
	\end{pmatrix} = \begin{pmatrix}
		\lambda^k(\mathbf{z}_1-\mathbf{z}_c)+\frac{\gamma^k}{d} \lambda^k(\mathbf{z}_3-\mathbf{z}_2) + \mathbf{z}_c \\
		\lambda^k (\mathbf{z}_2- \mathbf{z}_c)+ \mathbf{z}_c\\
		\lambda^k (\mathbf{z}_3- \mathbf{z}_c)+ \mathbf{z}_c
		\end{pmatrix}  
	\end{align*}
\end{algorithmic}
\caption{Retraction onto isosceles triangle manifold}
\label{alg3}
\end{algorithm}

 \section{Experimental Setup}
 The acoustic localization system, depicted schematically in Figure \ref{acoustic_schem}, comprises three main components: the Master Station, Acoustic Receiver, and Acoustic Transmitters. Each component is equipped with an NRF24L01 wireless module. The Master Station's module operates in transmit-only mode, while those on the Acoustic Transmitters and Receiver operate in receive-only mode.
\begin{figure}[H]
    \centering
    \includegraphics[width=\linewidth]{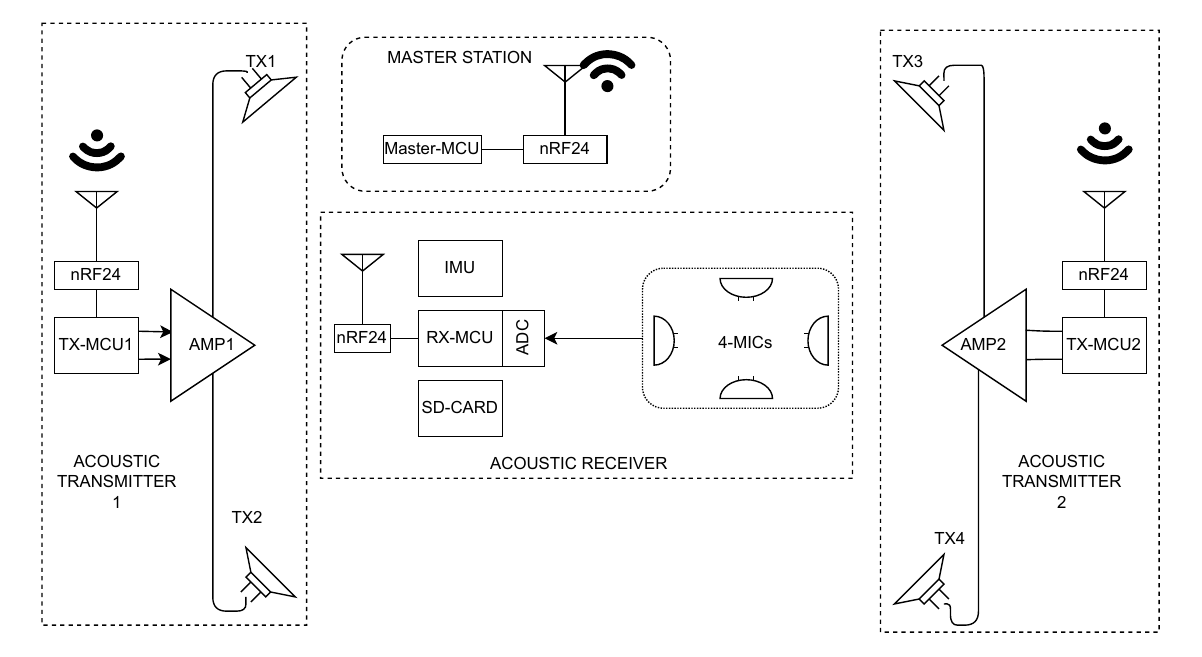}
    \caption{The system is composed of three conceptual components: (1) Master Station (2) Acoustic Receiver (3) Acoustic Transmitters}
    \label{acoustic_schem}
\end{figure}
The master station, depicted in Figure \ref{master_exp}, utilizes an STM32F469 Discovery kit \cite{STM32} with an NRF24L01 module \cite{NRF}. Each of the two acoustic transmitter stations also employs STM32F469 Discovery kits with NRF24L01 modules, XM-N1004 Sony stereo amplifiers \cite{sony}, and two Pioneer TS-T110 tweeters mounted on fixtures with retro-reflective spheres on tripods. The Pioneer TS-T110 tweeter has a bandwidth (BW) of 7 kHz and a central frequency of 20 kHz. Figure \ref{tx_exp} illustrates the TS-T110 tweeter setup with its fixture.
\begin{figure}[ht!]
    \centering
    \includegraphics[width=0.6\linewidth]{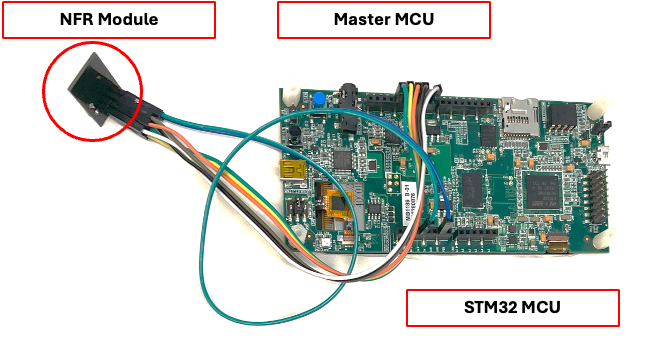}
    \caption{Master station}
    \label{master_exp}
\end{figure}

\begin{figure}[ht!]
    \centering
    \includegraphics[width=0.85\linewidth]{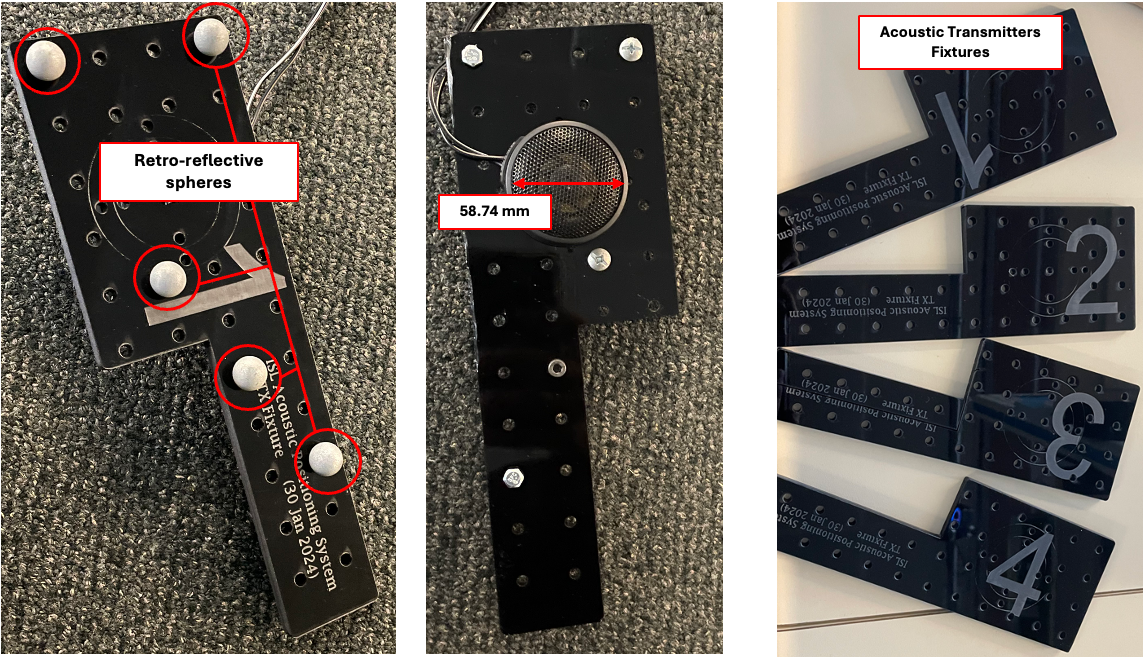}
    \caption{Acoustic transmitter}
    \label{tx_exp}
\end{figure}
The acoustic receiver station, forming the MD, utilizes an STM32F469 Discovery kit with an NRF24L01 module, an MTi-1 IMU kit \cite{MTi}, and a custom-made PCB board with four microphones arranged at the vertices of two side-to-side equilateral triangles, as shown in Figure \ref{rx_exp}. The side length of this equilateral triangle is $36.80$ mm. For this study, three out of the four microphones are utilized.
\begin{figure}[t!]
    \centering
    \includegraphics[width=0.75\linewidth]{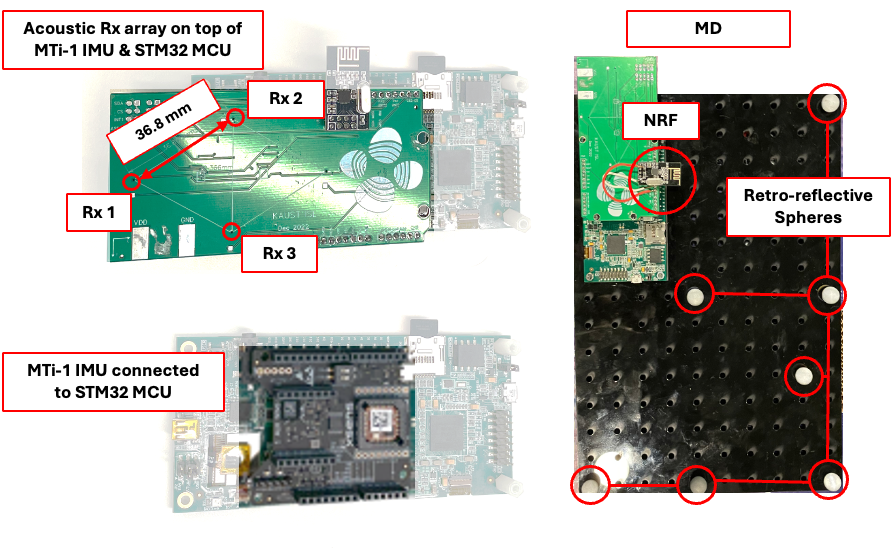}
    \caption{Acoustic receiver}
    \label{rx_exp}
\end{figure}
The experimental setup includes the master station, two acoustic transmitter stations, the MD (acoustic receiver station), and an OptiTrack motion capture system \cite{OPTR} providing ground truth for position and orientation, as depicted in Figure \ref{exp_set}.
\begin{figure}[t!]
    \centering
    \includegraphics[width=\linewidth]{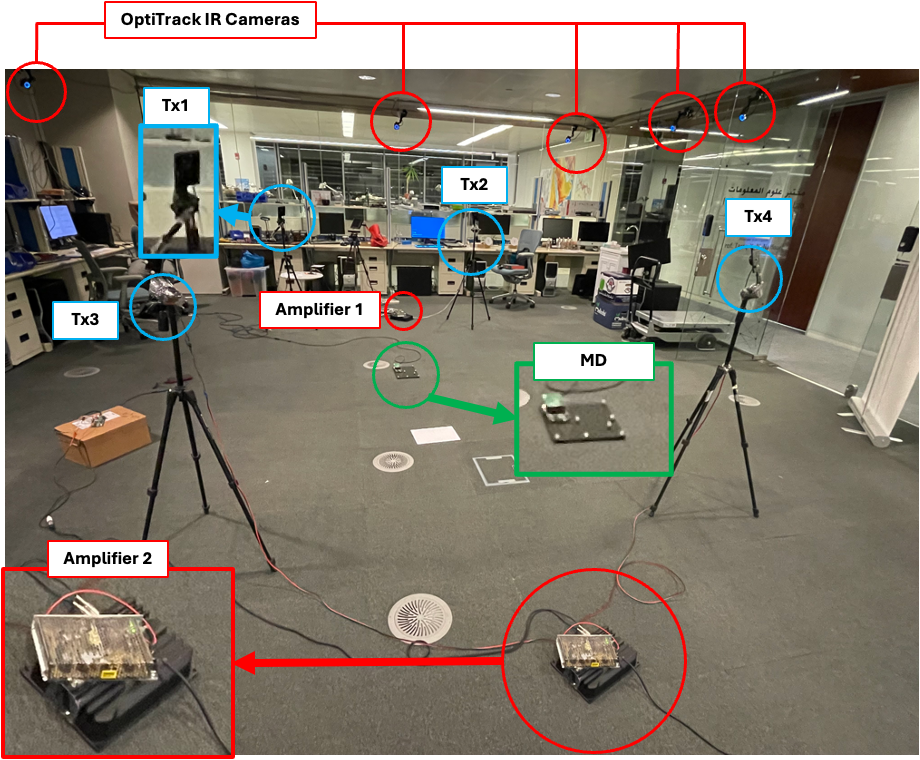}
    \caption{Experimental setup}
    \label{exp_set}
\end{figure}
The master station emits a periodic signal via a wireless link, establishing a common time reference upon reception by the acoustic devices. Acoustic transmitters emit signals detected by MEMS microphones, processed to determine time-of-flight. The acoustic receiver includes an IMU and an SD card for recording IMU data and acoustic signals at regular intervals. 

During lab experiments, the master station connects to a PC. A user initiates experiments of specified duration via commands relayed through the master’s RF link to prepare the acoustic devices. The master initiates acoustic transmission with a predetermined RF command, triggering the receiver to record signals from all four microphones, though only three are used in this work.

Ground truth positions of the acoustic transmitters and receiver are captured using an OptiTrack motion capture system \cite{OPTR}. retro-reflective markers on each acoustic transmitter and the receiver ensure accurate tracking. Post-experiment, OptiTrack records are synchronized with acoustic receiver data to facilitate error calculation and further analysis.

\section{Results}
\subsection{IMU and Acoustic Emulation Setup}
The evaluation setup consists of a $10 \text{ m} \times 5 \text{ m} \times 3 \text{ m}$ room equipped with an OptiTrack infrared tracking system. This system offers ground truth with $0.1$ mm 3D position accuracy, a maximum update rate of $250$ Hz \cite{OPTR}, and a rotational error of less than 0.05 degrees. OptiTrack, utilizing up to sixteen infrared cameras, tracks small spheres coated with retro-reflective film to provide accurate 3D location and orientation information. Figure \ref{KF_exp_set}  illustrates the target tracked by the IR cameras and the motion trajectory in the experimental setup. In the first part of the results section, the system measures the 3D position and orientation of the moving target to derive acceleration and angular velocity in the body frame, effectively creating synthetic IMU measurements.

To simulate IMU data, we add additive Gaussian noise with zero mean and variances $\sigma_a^2$ and $\sigma_{\omega}^2$ to the acceleration and angular velocity, respectively. Additionally, the vertices of an equilateral triangle are computed from the measured centroid position using Equation \ref{pc2pi}. In this mixed experimental-simulation setup, no actual IMU or acoustic beacons are used. Instead, we numerically simulate what would have been IMU or acoustic range measurements. Distances from each of the three vertices to each beacon are calculated using Euclidean distance, with additive Gaussian noise of zero mean and variance $\sigma_d^2$ added to simulate an acoustic ranging system. In the numerical simulation setup, the side length of the equilateral triangle is set to $20$ cm.
\begin{figure}[h!]
    \centering
    \subfloat[Localization's Target]{\includegraphics[width=2cm, height = 1.5 cm]{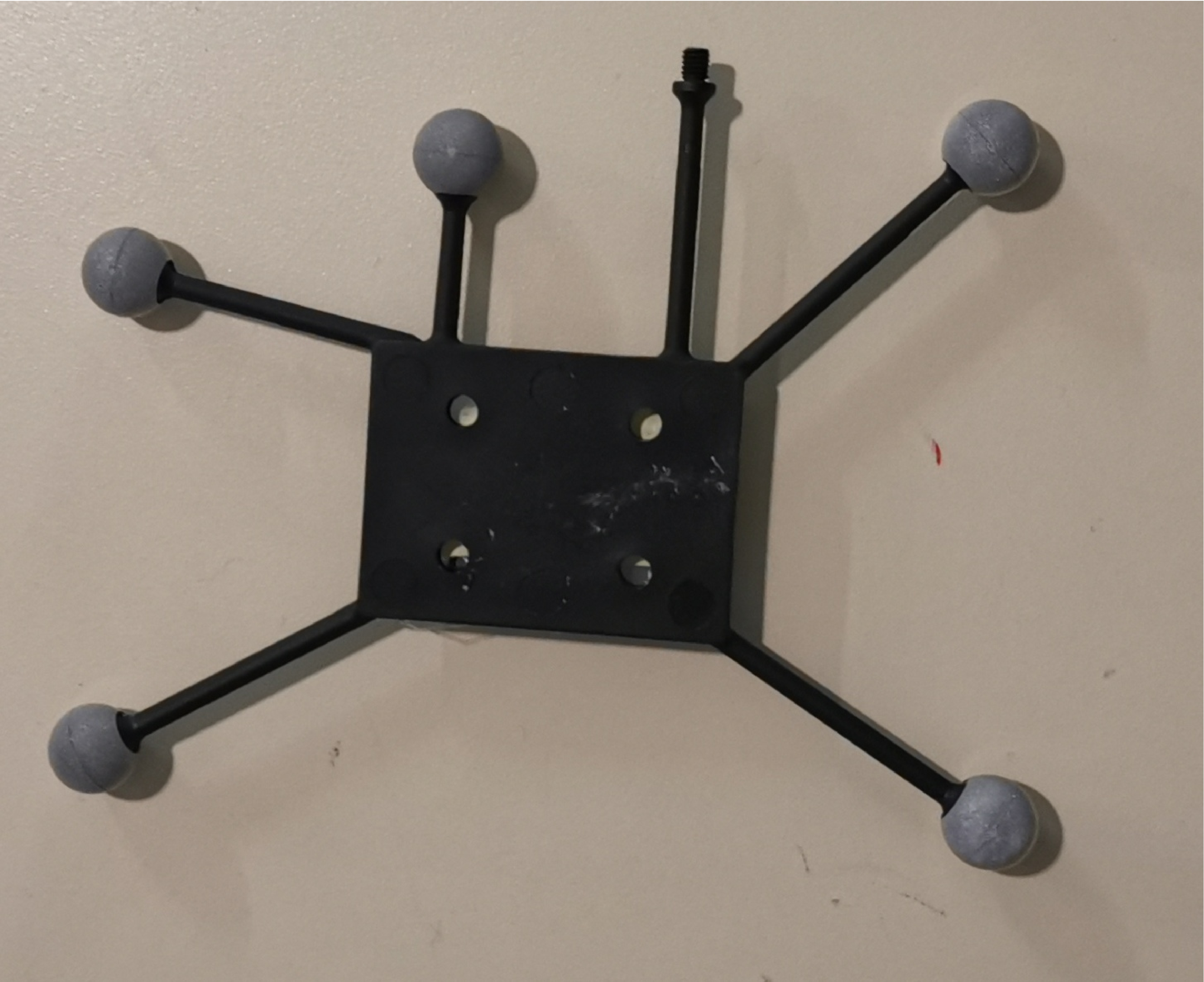}}
    \label{KF_target}
    \subfloat[Room equipped with high-precision infrared tracking system]{\includegraphics[width=9cm, height = 5 cm]{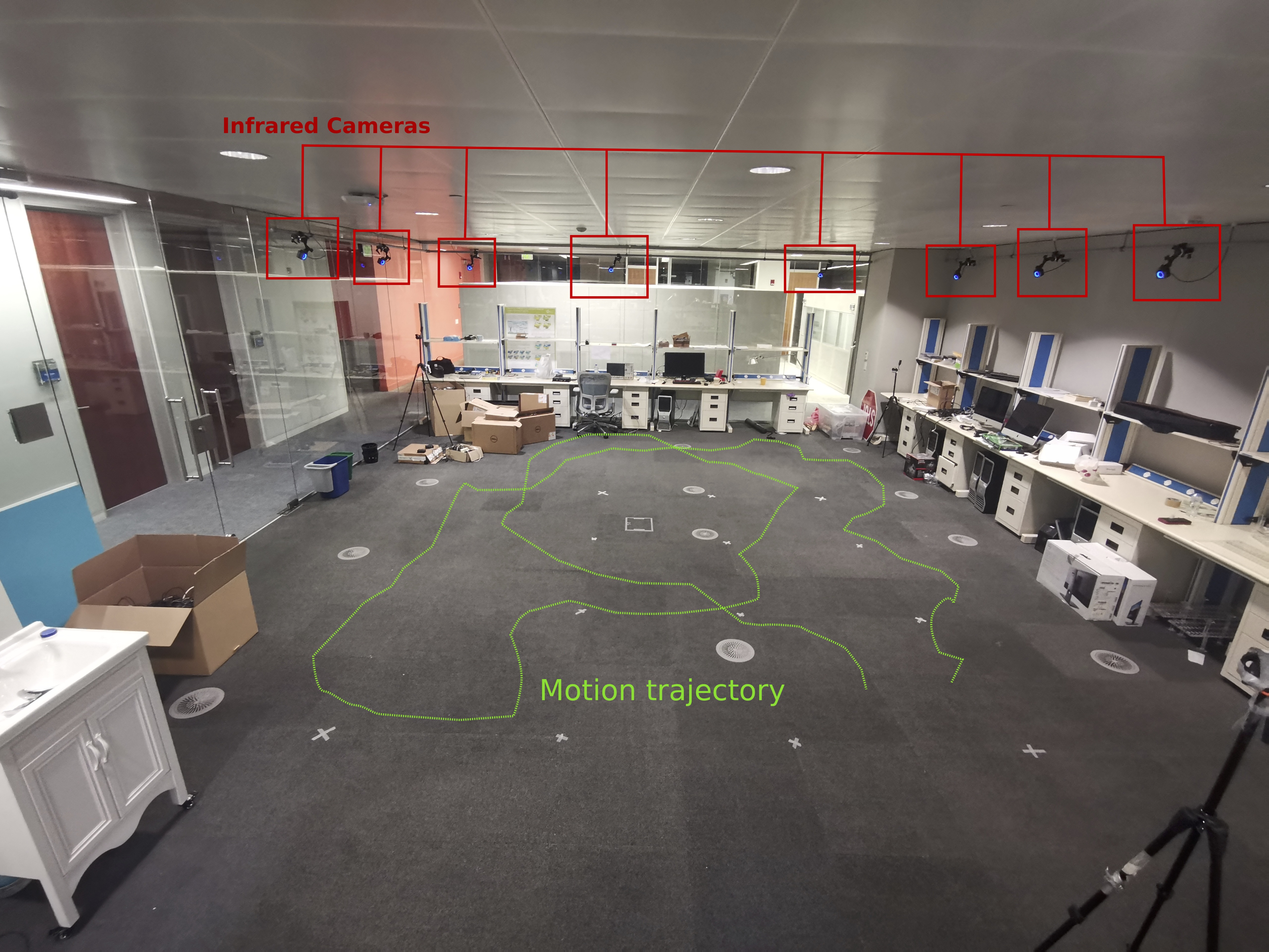}}
    \caption{Mixed Experimental-Simulation setup}
    \label{KF_exp_set}
\end{figure}

\subsection{Numerical Simulations}
We evaluate the performance of the proposed algorithm by computing the Root Mean Square Error (RMSE) of the three vertices' positions. The algorithms assessed include the EKF and UKF combined with Riemannian Steepest Descent (RSD) and Riemannian Trust Region (RTR) methods. The setup employs an acoustic ranging system to determine the positions of a triangle's vertices, which then provides the centroid position and orientation of the MD. The effectiveness of fusing the estimated position and orientation is assessed as follows:
\begin{itemize}
    \item Centroid Position Correction: We compare two systems: (1) A single-receiver acoustic positioning system, labeled “1RX” in the RMSE plots, and (2) A three-receiver acoustic positioning system, labeled “P$_c$-GN” in the RMSE plots, which corrects the INS using linear Kalman filter. The localization method used in these two systems is based on Gauss-Newton (GN) method.
    \item Centroid Position Correction: We also compare two three-receiver acoustic systems: (1) those employing Riemannian-based methods, labeled “P$_c$-RSD” and “P$_c$-RTR”, and (2) those using GN-based methods, labeled “P$_c$-GN”. These systems use EKF and UKF. 
    \item Centroid Position and Orientation Correction: We compare: (1) three-receiver acoustic systems using Riemannian-based methods, labeled “EKF-RSD”, “EKF-RTR”, “UKF-RSD”, and “UKF-RTR”, and (2) GN-based methods, labeled “EKF-GN” and “UKF-GN”.
\end{itemize}}

The proposed and benchmark algorithms utilize Kalman Filters (KF), EKF, and UKF to fuse position and orientation estimates from the INS with those obtained via acoustic-based positioning. Note that the input data for INS and acoustic positioning algorithms, such as acceleration, angular velocity, and acoustic ranges, are numerically generated. The acoustic-based positioning uses either; (1) Gauss-Newton (GN) method as an unconstrained-localization solver or (2) RSD and RTR as a constrained-localization solvers. 

\begin{figure}[h!]
    \centering
    \subfloat[3D motion path]{\includegraphics[scale=0.2,trim={70 90 0 100},clip]{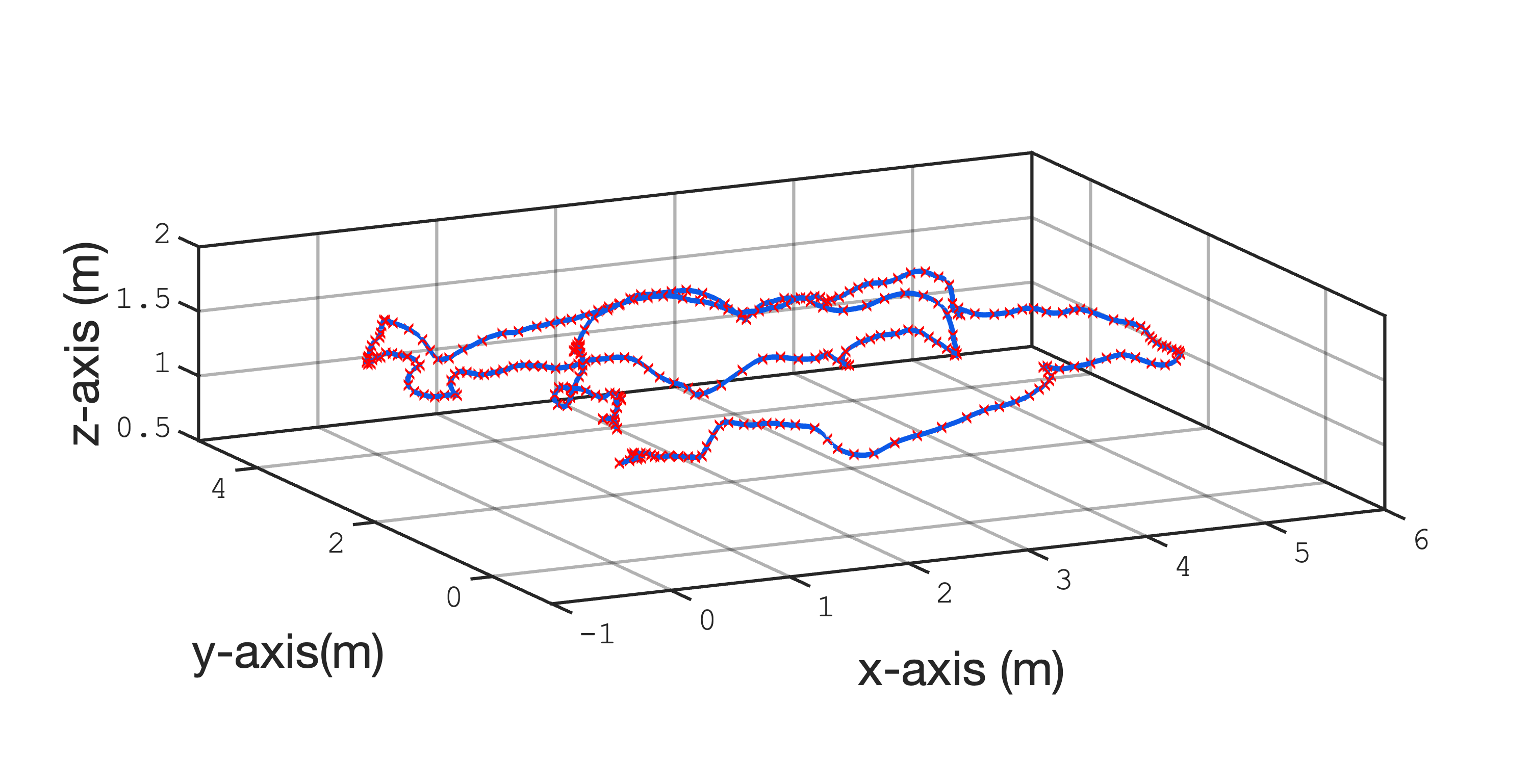}}
    \vfill
    \centering
    \subfloat[Top view]{\includegraphics[scale=0.2,trim={250 10 230 40},clip]{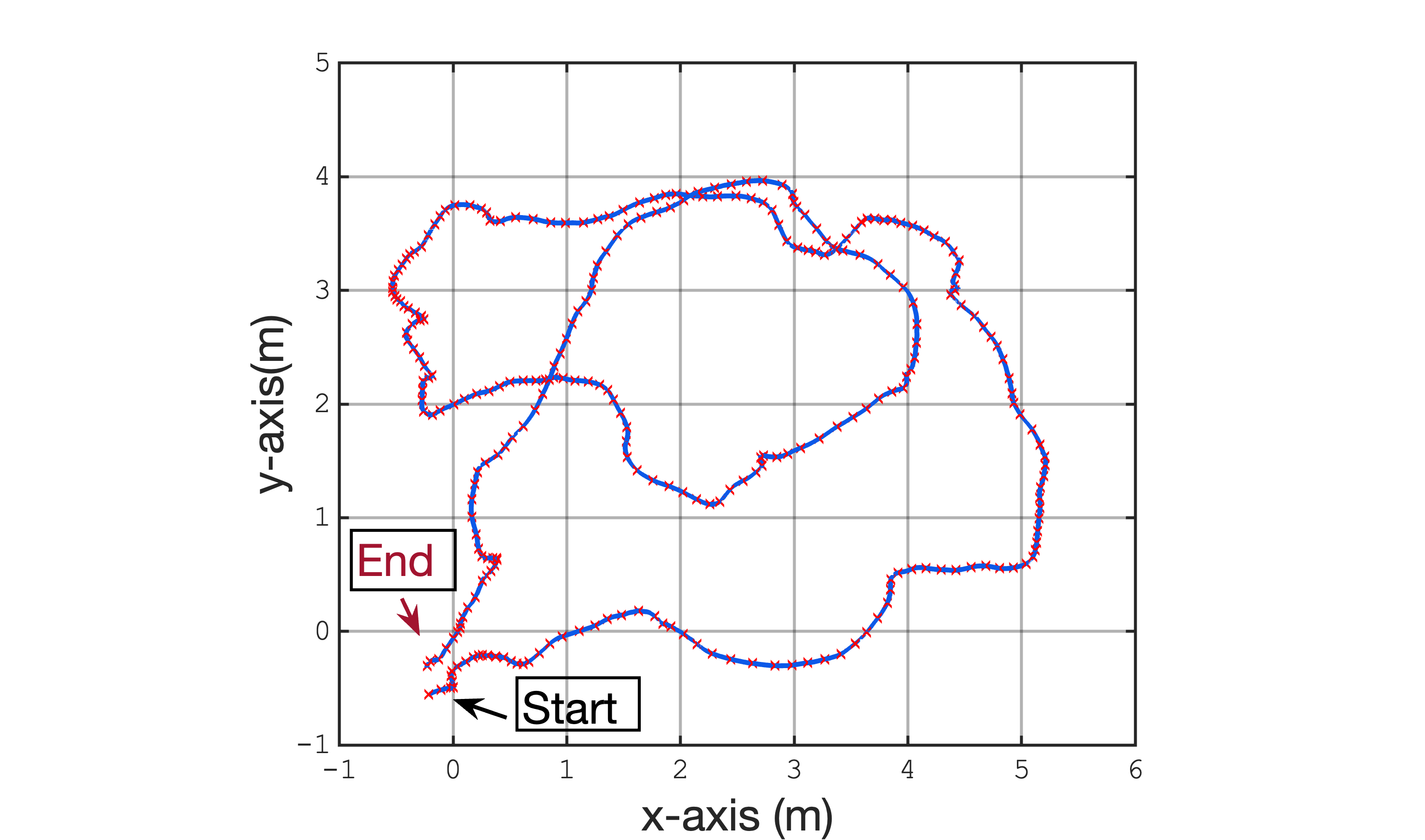}}
    \caption{Motion trajectory in emulation}
    \label{motion_traj}
\end{figure}

\begin{figure*}[h!]
\begin{minipage}[t]{0.3\textwidth}
    \centering
    \subfloat[Yaw angle with emulated IMU]
    {\includegraphics[scale=0.115]{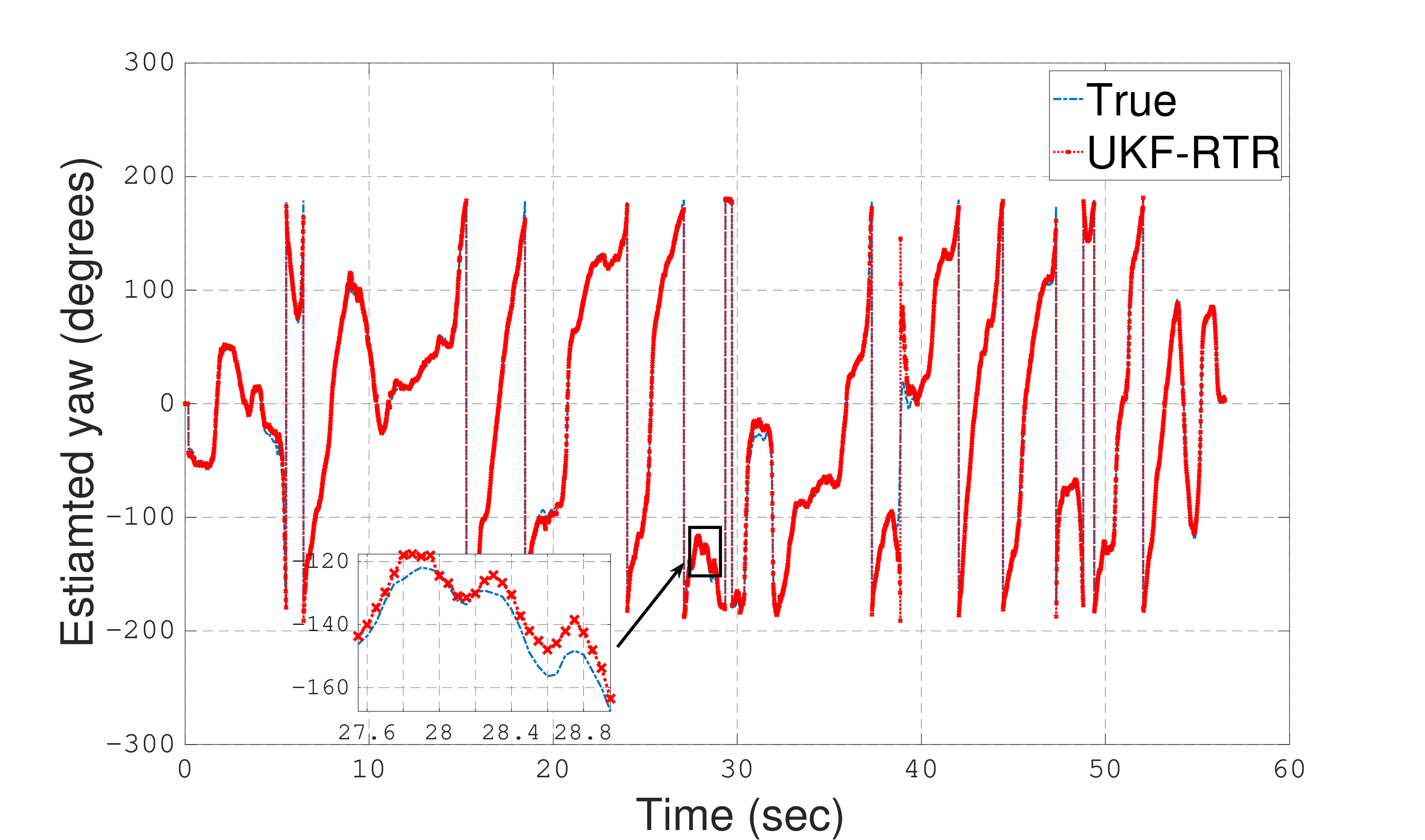}}
\end{minipage}
\begin{minipage}[t]{0.3\textwidth}
    \subfloat[Pitch angle with emulated IMU]
    {\includegraphics[scale=0.115]{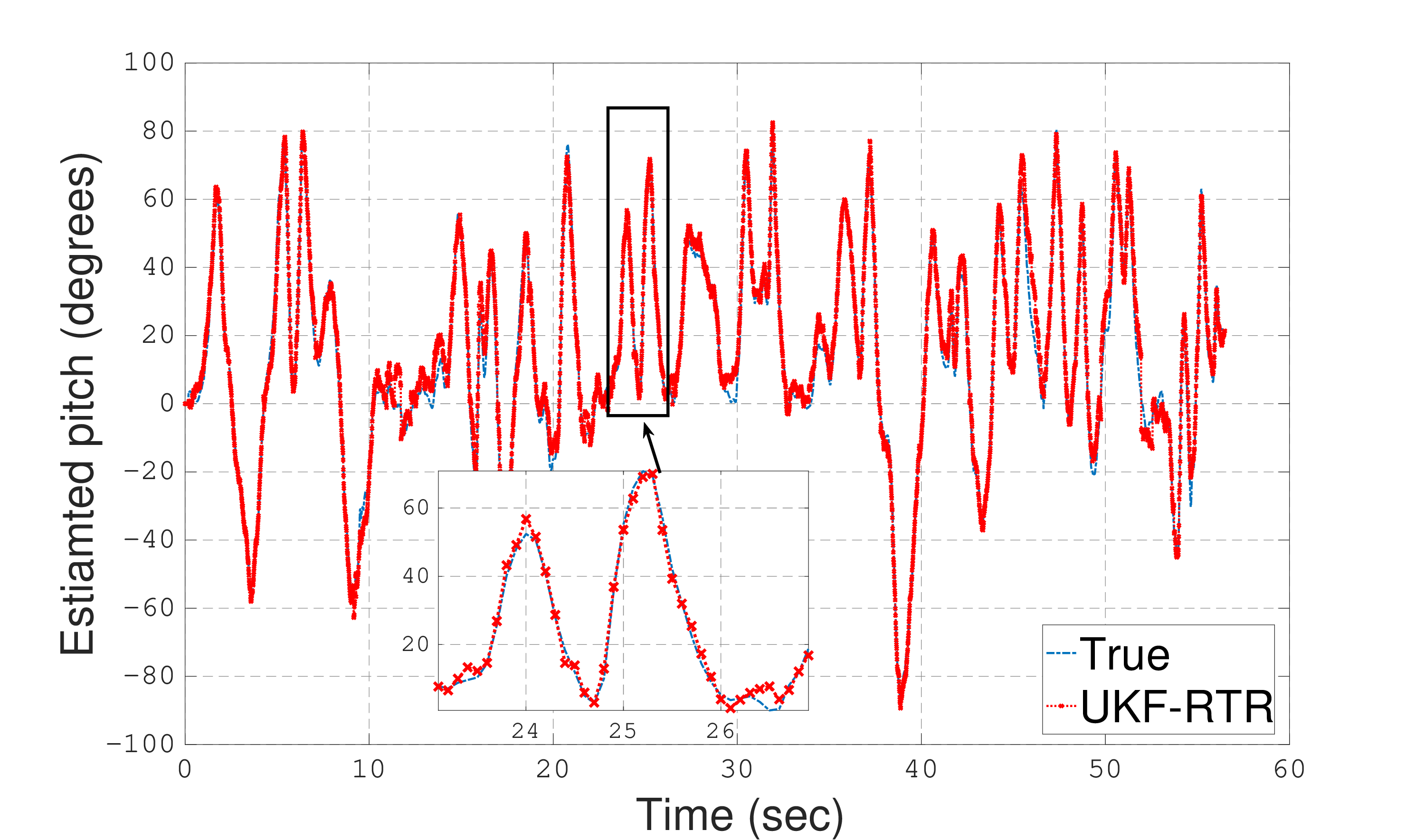}}
\end{minipage}
\begin{minipage}[t]{0.3\textwidth}
    \subfloat[Roll angle with emulated IMU]
    {\includegraphics[scale=0.115]{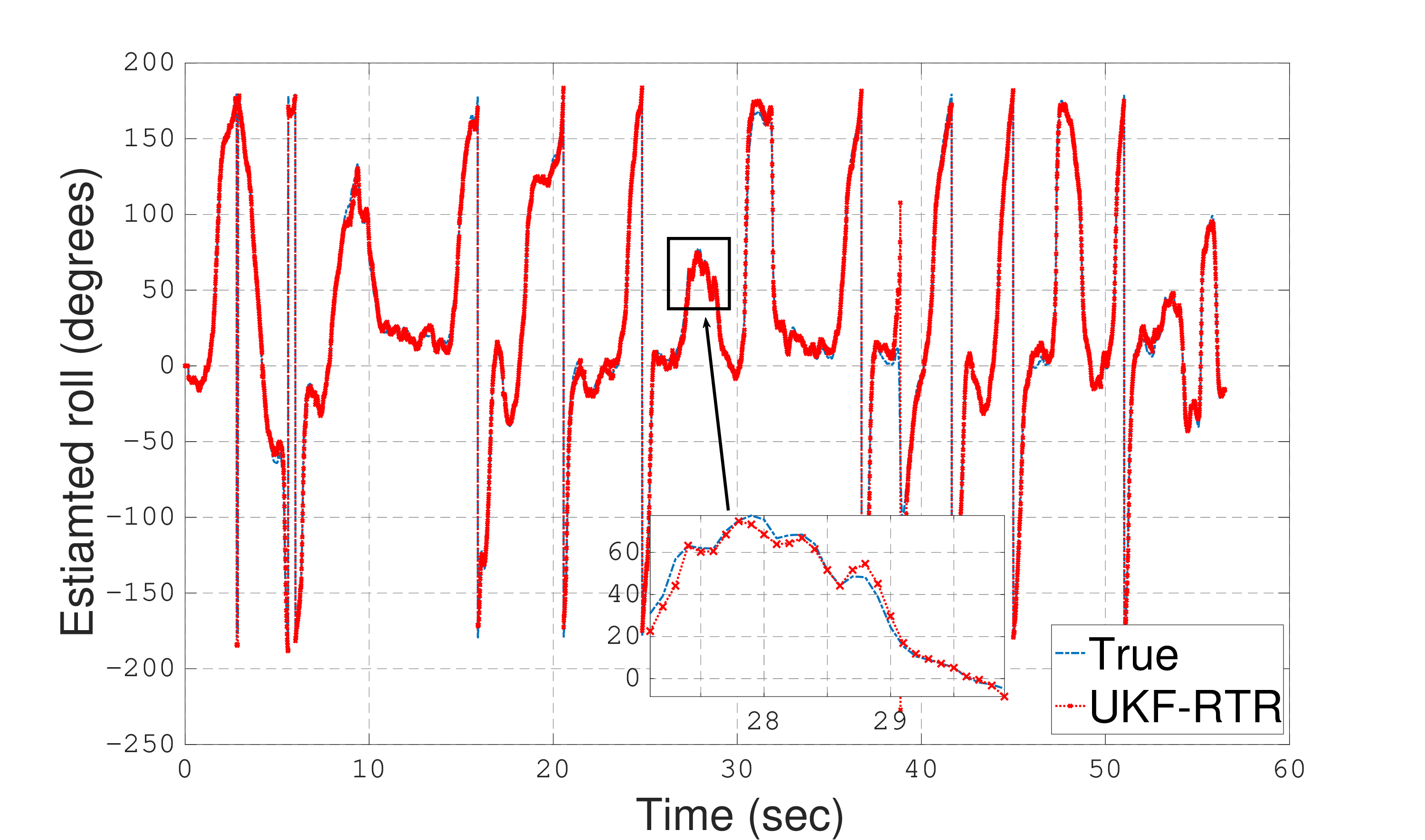}}
\end{minipage}
    \caption{Moving target's orientation }
    \label{sim_orient}
\end{figure*}

To generate test data, a person holding the target shown in Figure \ref{KF_exp_set} walks around the test area at a normal walking speed ($1$ m/s to $1.8$ m/s), with the motion trajectory illustrated in Figure \ref{motion_traj}. The “emulated” IMU update rate is fixed at $100$ estimates per second for all simulations. 
True and estimated Euler angles of the moving target are shown in Figure \ref{sim_orient}.

\begin{figure}[h!]
    \centering
    \includegraphics[scale=0.17]{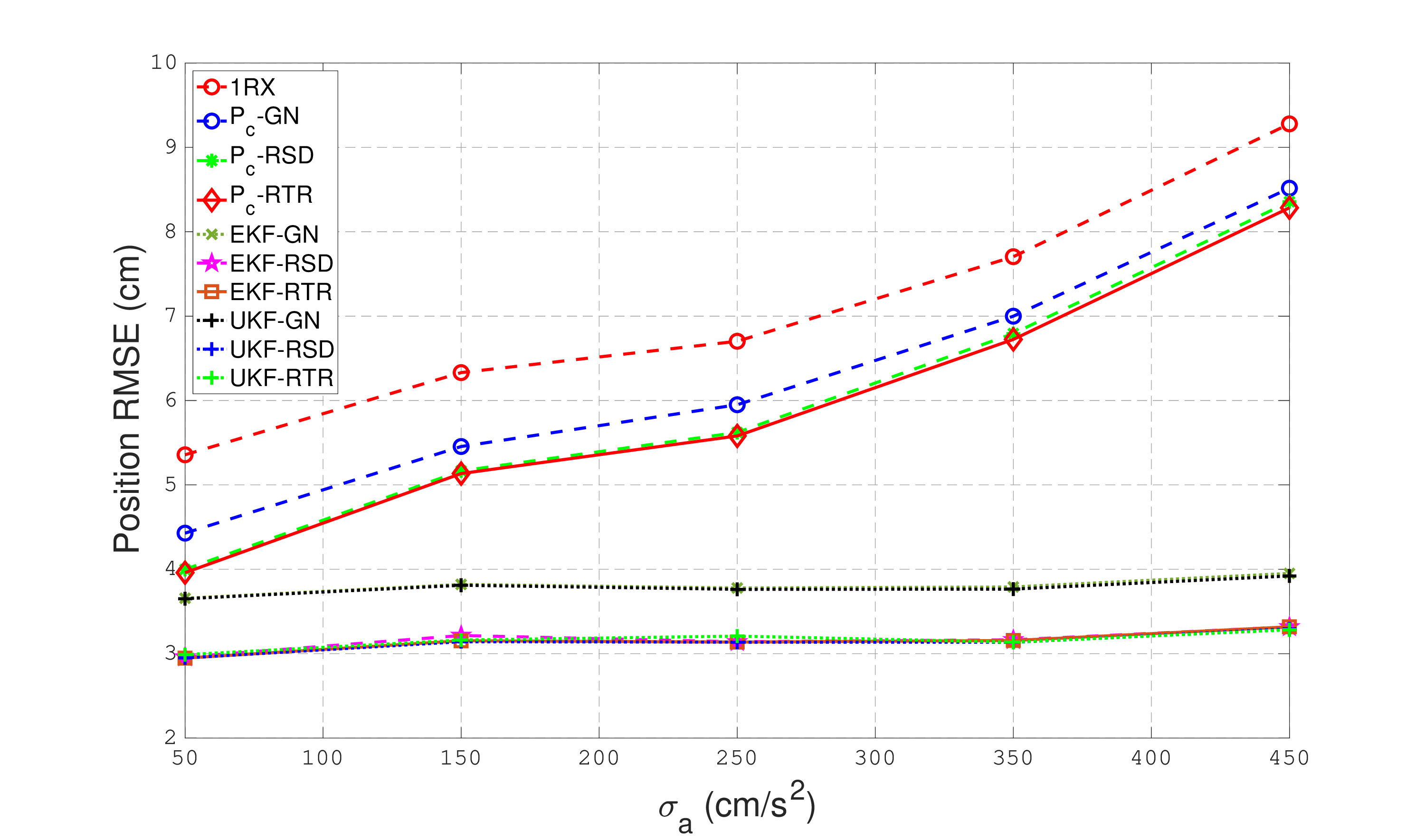}
    \caption{RMSE vs $\sigma_a$}
    \label{pc_v_snr_a}
\end{figure}

\begin{figure}[h!]
    \centering
    {\includegraphics[scale=0.17]{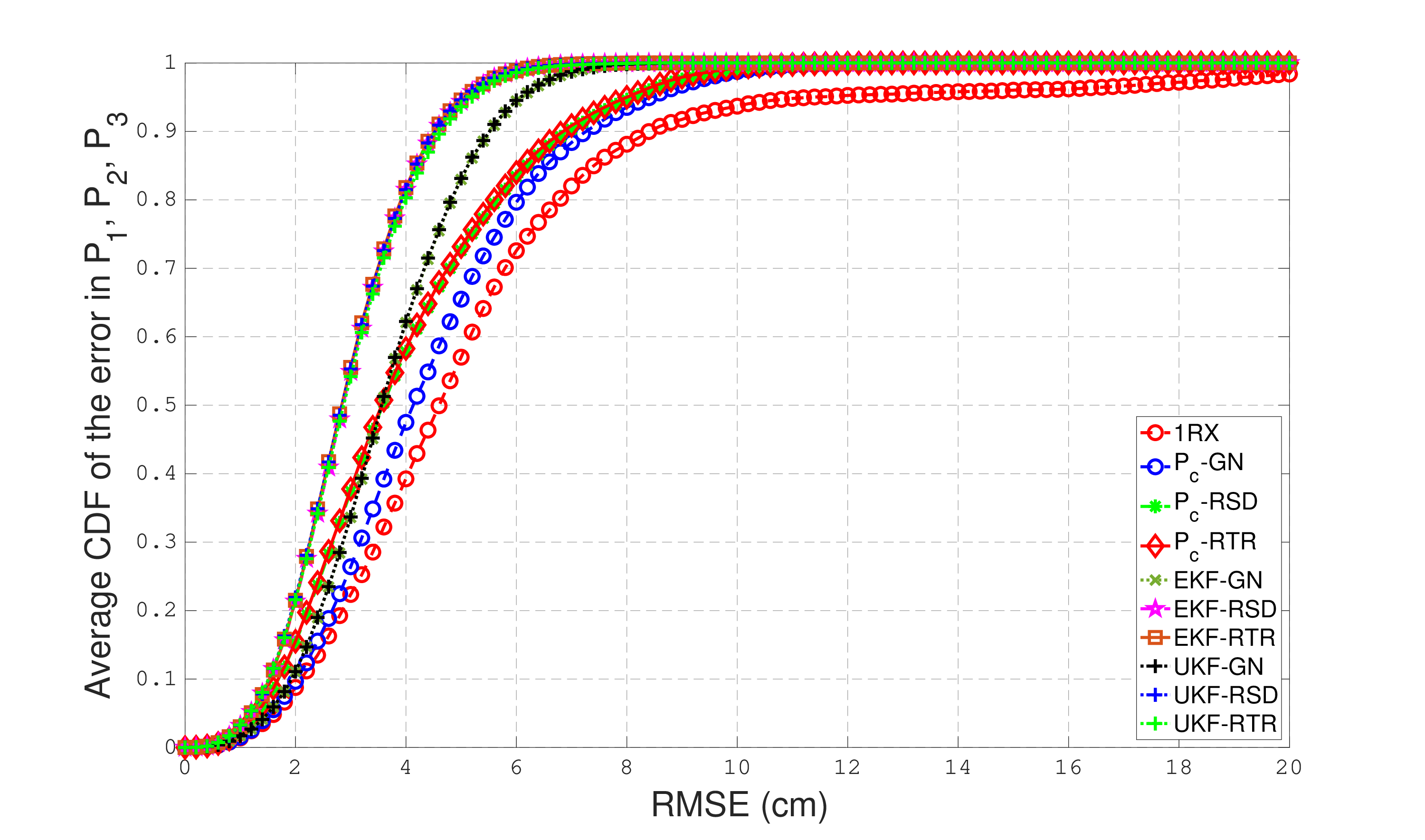}}
    \caption{Average CDF at $\sigma_a = 50 \text{ cm/s}^2$, $\sigma_\omega = 0.5 \text{ rad/s}$, $\sigma_d = 2.5 \text{ cm}$}
    \label{cum_a}
\end{figure}

First, we evaluate the impact of varying Gaussian noise added to the measured acceleration. In this simulation, Gaussian noise added to measured angular velocity and distances have fixed standard deviations of $\sigma_{\omega} = 0.5 \text{ rad/s}$ and $\sigma_d = 2.5 \text{ cm}$, respectively, while the standard deviation of the Gaussian noise added to measured acceleration $\sigma_a$ ranges from $50$ to $450 \text{ cm/s$^2$}$. The average RMSE of the three vertices’ positions for all algorithms increases with higher $\sigma_a$, as shown in Figure \ref{pc_v_snr_a}. The accuracy of the estimated position and orientation improves with three receivers. Because of the proposed orientation-correction method using the three receivers, the average RMSEs of the three vertices get reduced. The Riemannian-based Kalman filter algorithms (RSD-based and RTR-based) outperform benchmark algorithms in vertex position accuracy.
The average cumulative distribution function (CDF) of the RMSEs for the three vertices is plotted in Figure \ref{cum_a}. Algorithms that do not correct orientation using INS, i.e., those without acoustic-based orientation correction, achieve lower accuracy due to having the acceleration measured in the body frame. RSD and RTR methods perform better than GN-based methods. EKF and UKF show similar performance because of the slow movement relative to the high update rate, making the first-order approximation of EKF accurate.

\begin{figure}[!h]
    \centering
    \includegraphics[scale=0.17]{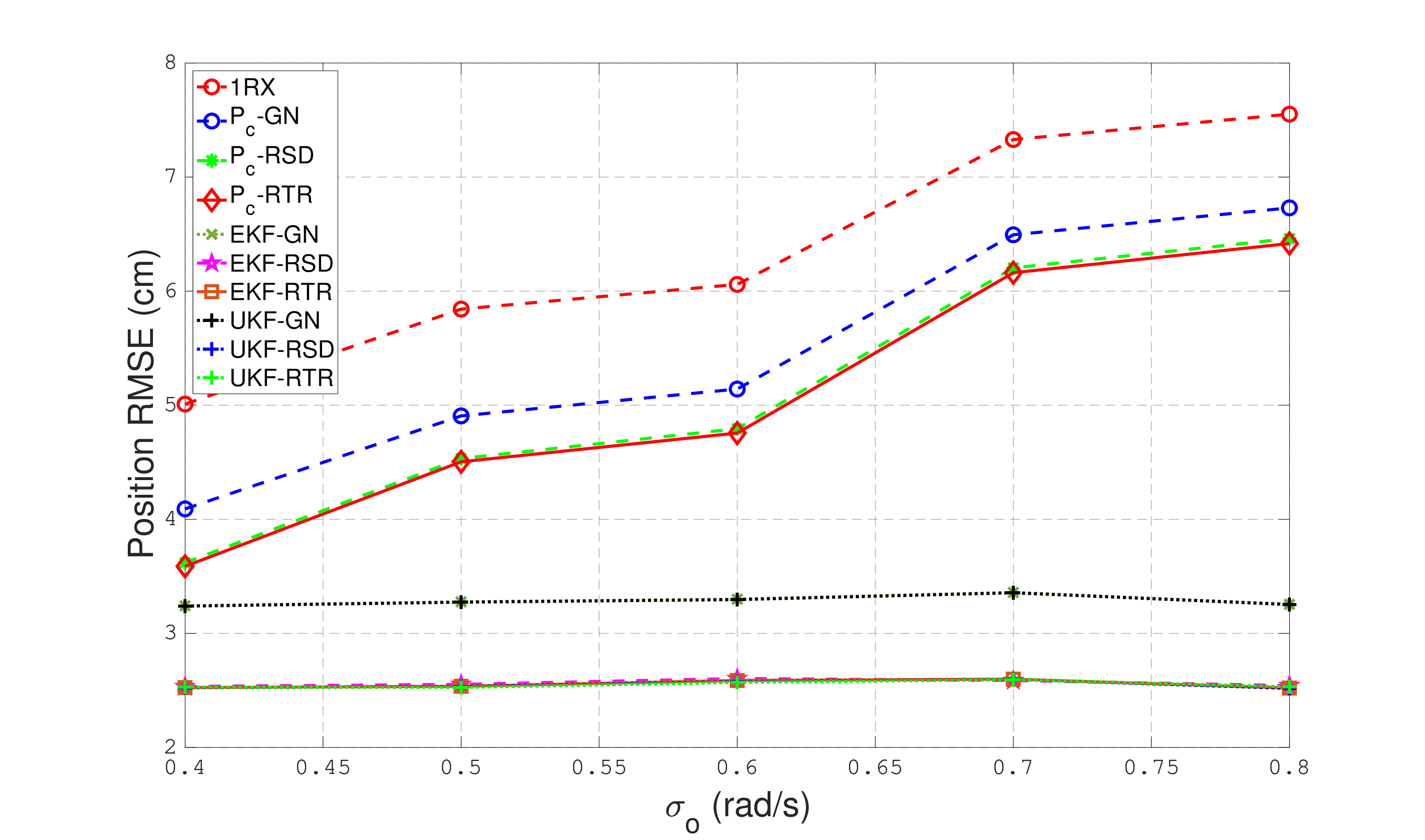}
    \caption{RMSE vs $\sigma_{\omega}$}
    \label{pc_v_snr_q}
\end{figure}

Next, we assess the effect of varying the standard deviation of Gaussian noise added to the measured angular velocity $\sigma_{\omega}$ from $0.4$ rad/s to $0.8$ rad/s while keeping the standard deviations of Gaussian noise added to measured acceleration and distances fixed at $\sigma_a$ = $50 \text{ cm/s$^2$}$ and $\sigma_d$ = $2.5$ cm. As the angular velocity noise increases, the accuracy of vertices positions degrades for algorithms that do not correct INS-based orientation ("1RX", "P$_c$-GN", "P$_c$-RSD", "P$_c$-RTR"). Conversely, algorithms that use acoustic-based orientation correction maintain high accuracy despite increased angular velocity noise, as shown in Figure \ref{pc_v_snr_q}. 
  
\begin{figure}[!h]
    \centering
    \includegraphics[scale=0.17]{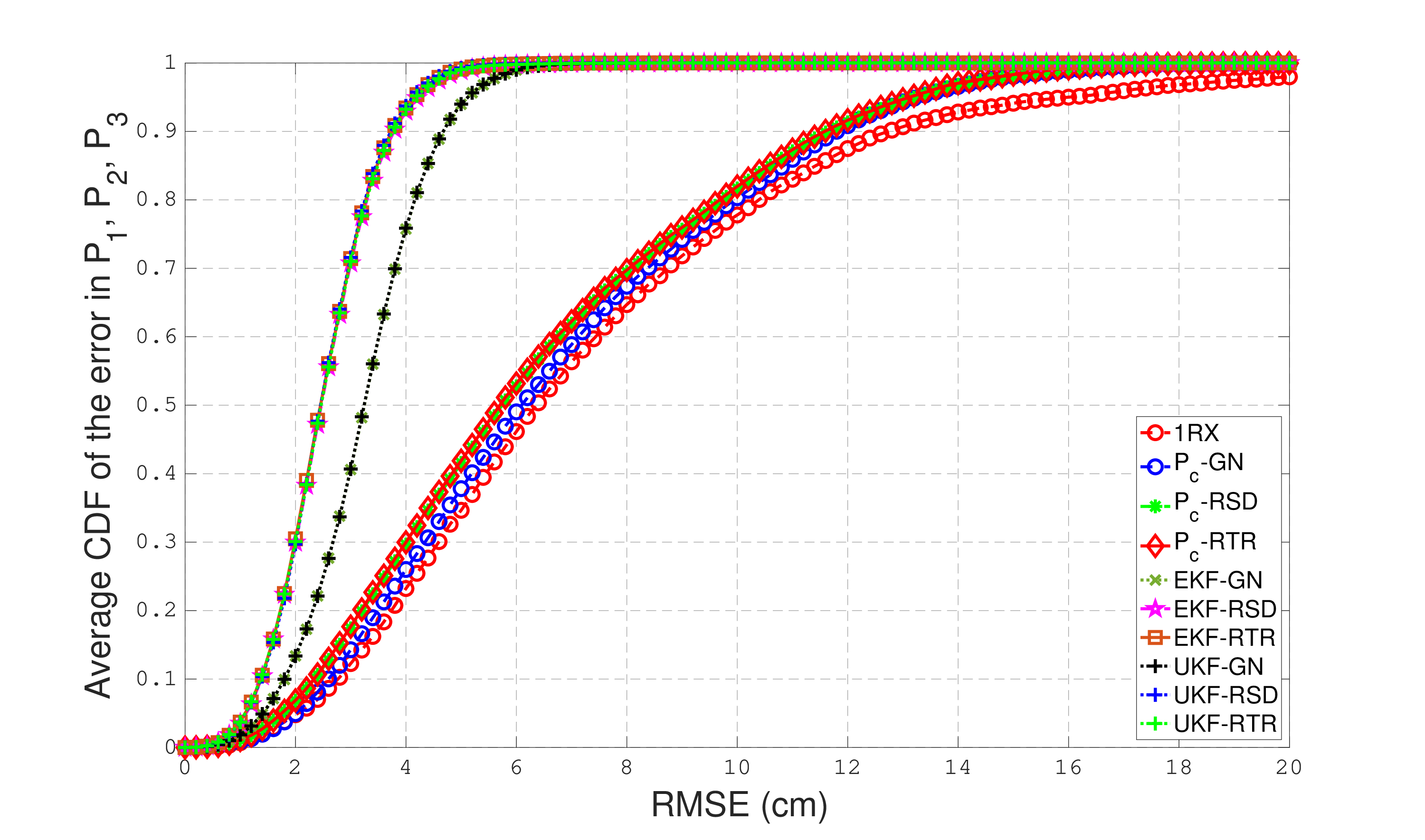}
    \caption{Average CDF at $\sigma_a = 50 \text{ cm/s}^2$, $\sigma_\omega = 0.8 \text{ rad/s}$, $\sigma_d = 2.5 \text{ cm}$}
    \label{cum_o}
\end{figure}

Figure \ref{cum_o} illustrates the average CDF of RMSEs at $\sigma_a = 50 \text{ cm/s}^2$, $\sigma_\omega = 0.8 \text{ rad/s}$, and $\sigma_d = 2.5 \text{ cm}$, showing that orientation-correcting algorithms maintain high accuracy.
	
\begin{figure}[!h]
    \centering
    \includegraphics[scale=0.17]{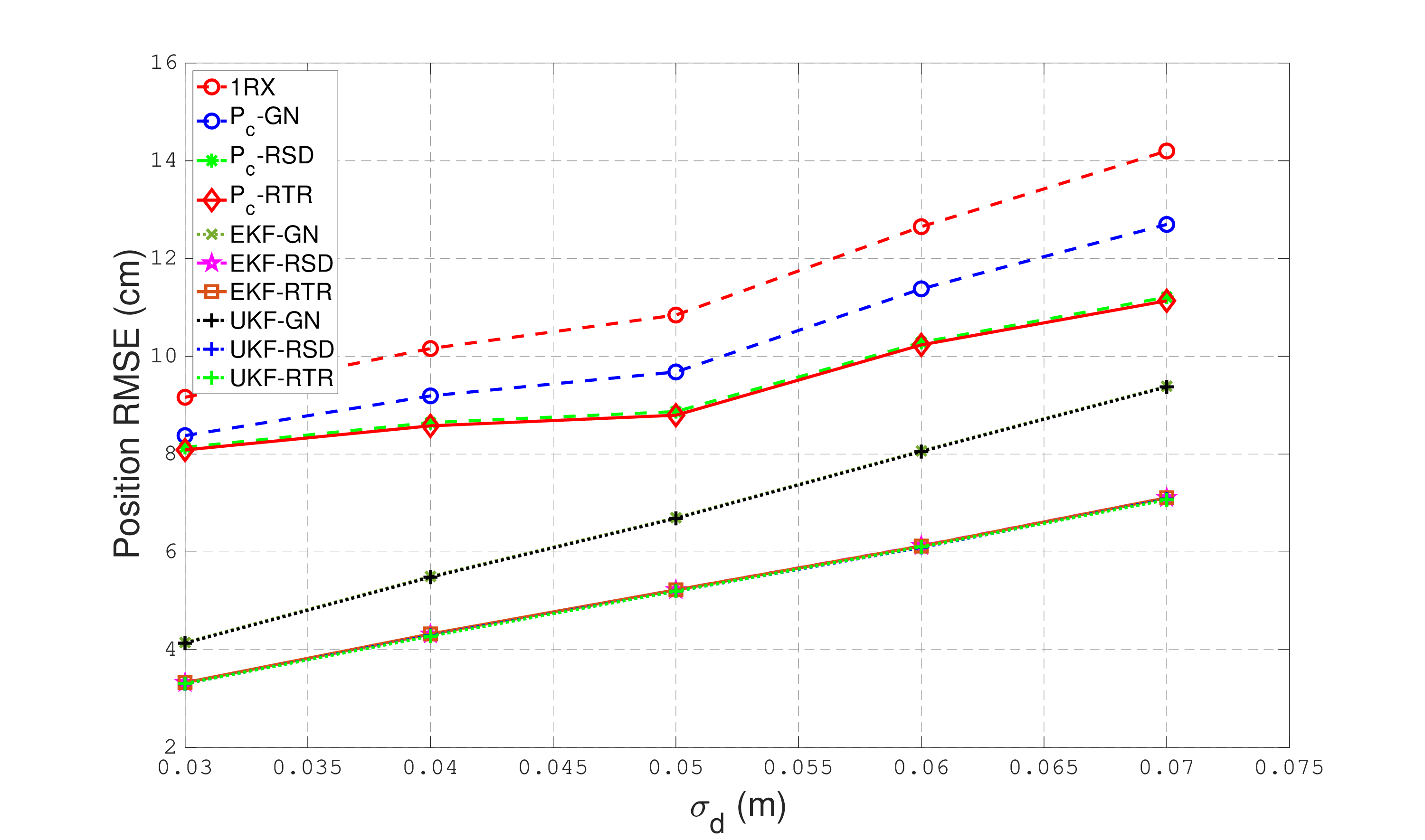}
    \caption{RMSE vs $\sigma_d$}
    \label{pc_v_snr_d}
\end{figure}

In a subsequent simulation, we evaluate performance with varying standard deviation $\sigma_d$ of Gaussian noise added to measured distances from $3$ cm to $7$ cm while fixing the standard deviations of noise in measured acceleration and angular velocity at $\sigma_a = 250 \text{ cm/s$^2$}$ and $\sigma_{\omega} = 1$ rad/s. At very low noise levels, the trilateration-based method performs similarly to Riemannian-optimization-based methods. However, as noise increases, benchmark methods degrade significantly compared to the proposed methods, as shown in Figure \ref{pc_v_snr_d}. Moreover, we notice that the degradation in the accuracy of the algorithms that correct the INS-based orientation is higher than the degradation in the algorithms that do not correct the orientation. This is expected since the accuracy of the estimated orientation using the acoustic-based method degrades with increasing the noise added to the measured distances. 

\begin{figure}[!h]
    \centering
    \includegraphics[scale=0.17]{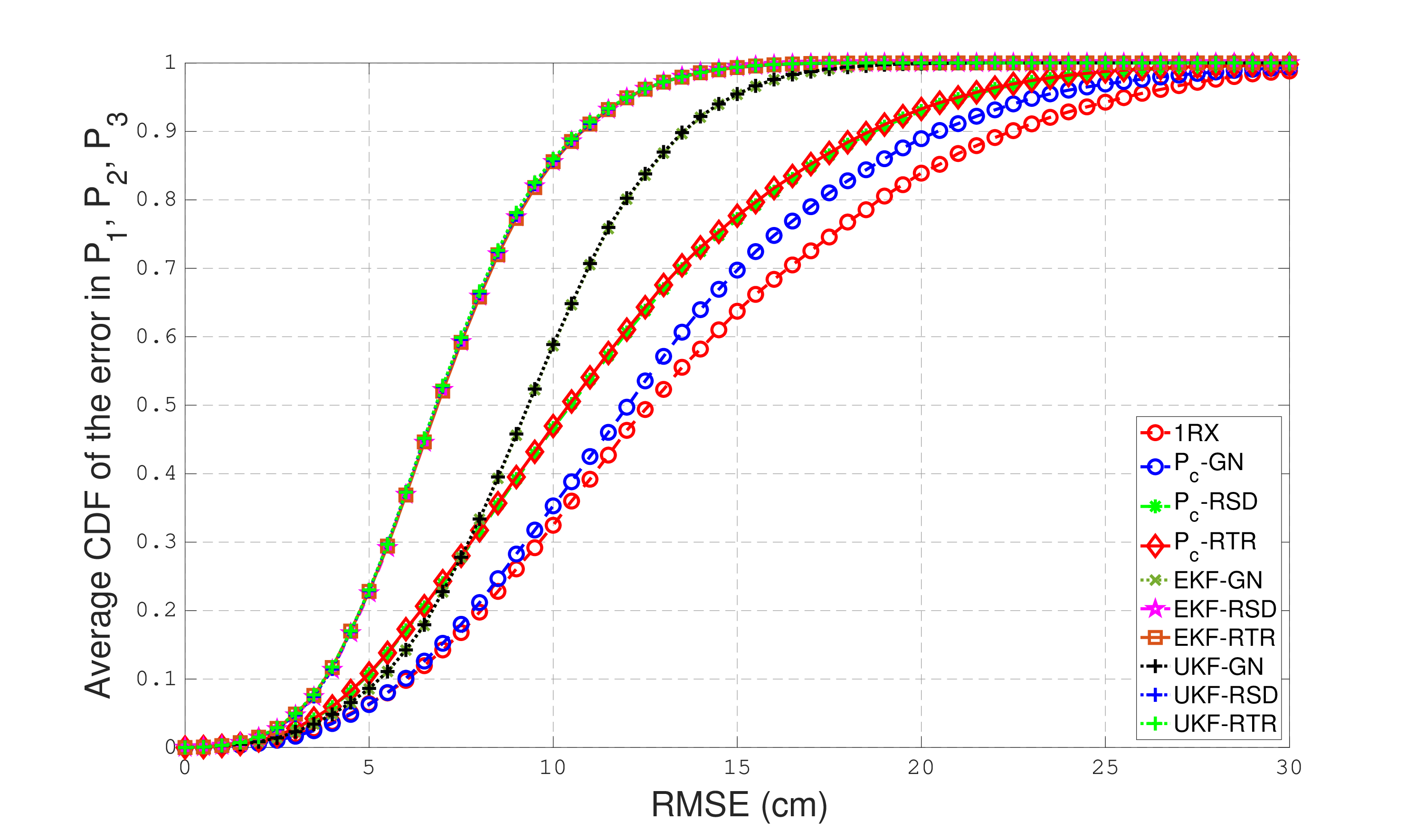}
    \caption{Average CDF at $\sigma_a = 250 \text{ cm/s}^2$, $\sigma_\omega = 1 \text{ rad/s}$, $\sigma_d = 7 \text{ cm}$}
    \label{cum_d}
\end{figure}

\begin{figure}[]
    \centering
    \includegraphics[scale=0.17]{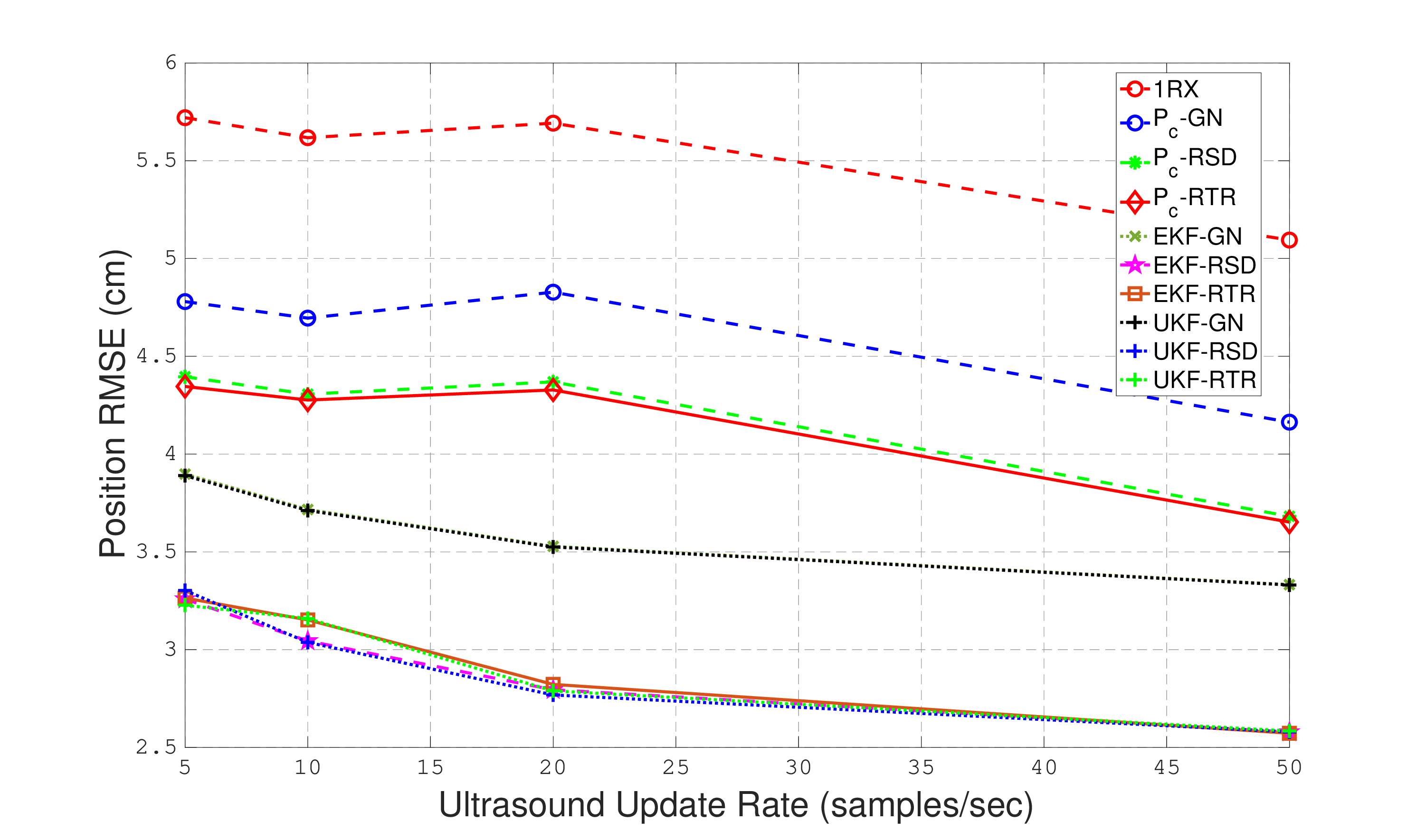}
    \caption{RMSE vs acoustic system update rate}
    \label{pc_v_snr_u}
\end{figure}

Figure \ref{cum_d} shows the average CDF of RMSE for the three vertices at $\sigma_a = 250 \text{ cm/s}^2$, $\sigma_\omega = 1 \text{ rad/s}$, and $\sigma_d = 7 \text{ cm}$. RTR- and RSD-based methods with orientation correction outperform all benchmark methods, with over $85\%$ of position estimates having less than $10$ cm RMSE, compared to less than $60\%$ for GN-based methods with orientation correction and less than $35\%$ without orientation correction.	
Finally, we assess the impact of varying the acoustic system’s update rate from $5$ to $50$ estimates per second, with fixed Gaussian noise standard deviations ($\sigma_a = 50 \text{ cm/s$^2$}$, $\sigma_{\omega} = 0.5$ rad/s and $\sigma_d = 2.5$ cm.). All methods show improved positioning accuracy with higher update rates, though at different rates, as shown in Figure \ref{pc_v_snr_u}. 
  
\begin{figure}[!h]
    \centering
    \includegraphics[scale=0.175]{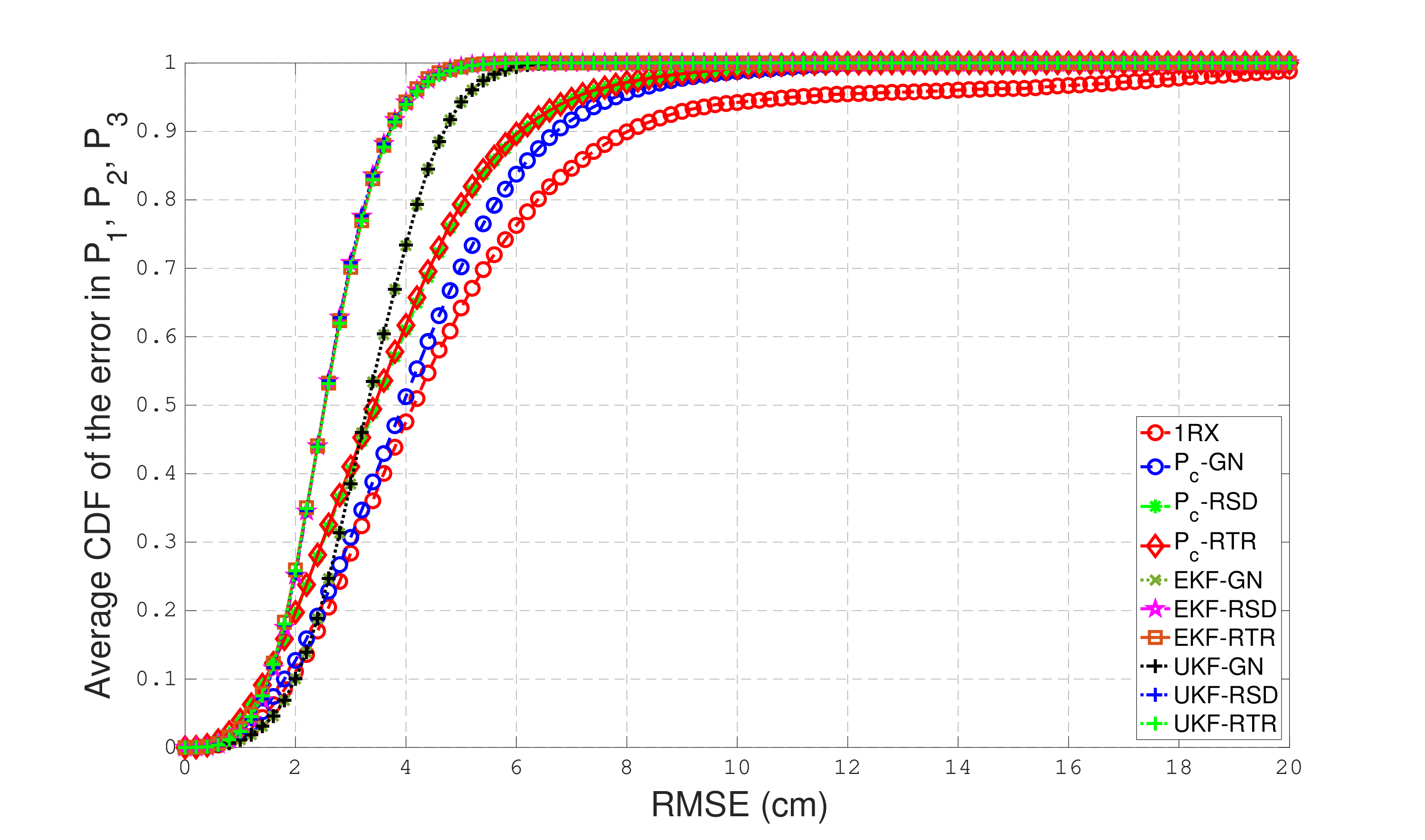}
    \caption{Average CDF at $\sigma_a = 50 \text{ cm/s}^2$, $\sigma_\omega = 0.5 \text{ rad/s}$, $\sigma_d = 2.5 \text{ cm}$ and acoustic update rate of $50$ Hz}
    \label{cum_u}
\end{figure}

\begin{figure}[h!]
\begin{minipage}[t]{0.3\textwidth}
\centering
    \subfloat[Error in yaw angle in the simulation]
    {\includegraphics[scale=0.15]{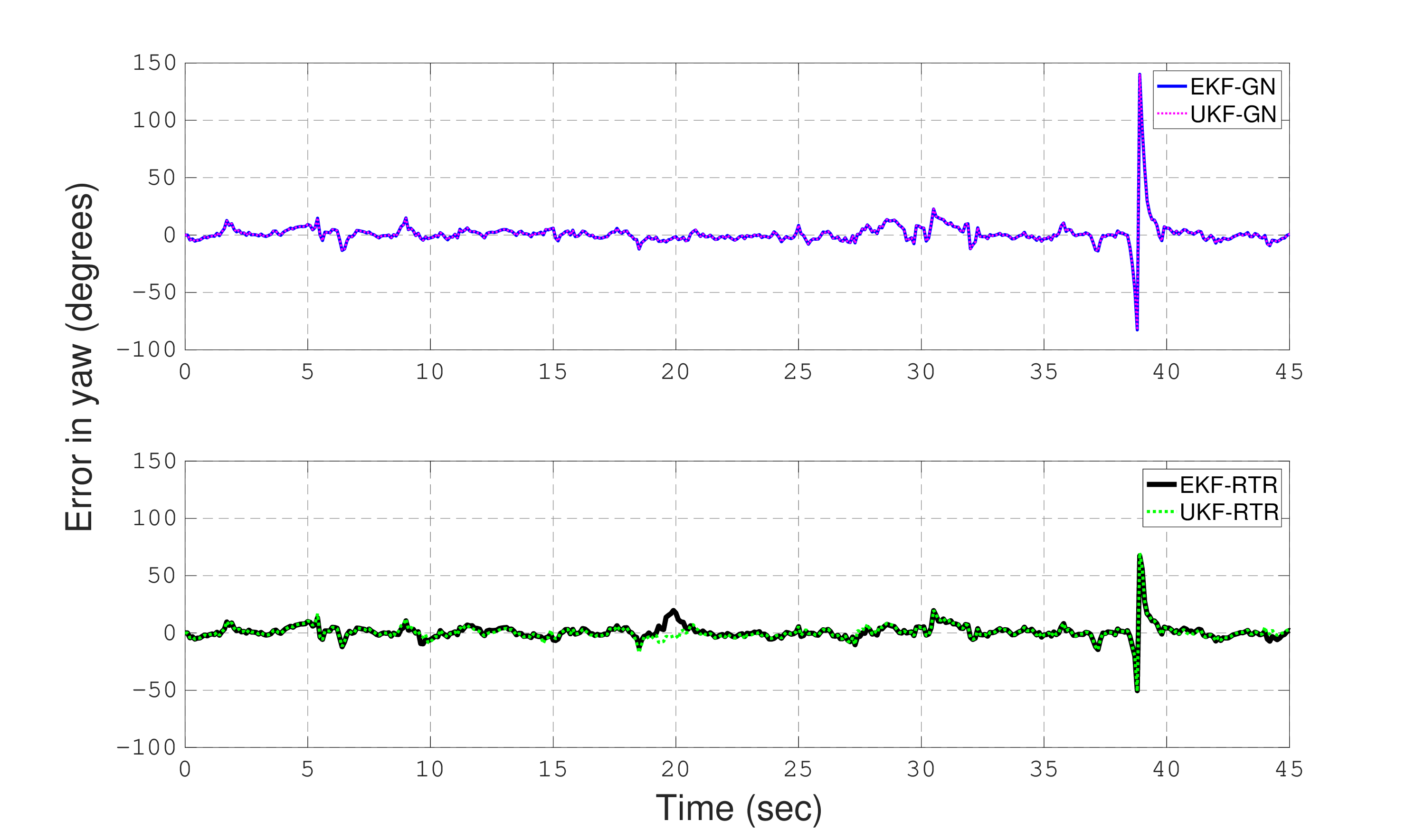}}
\end{minipage}
\begin{minipage}[t]{0.3\textwidth}
\centering
    \subfloat[Error in pitch angle in the simulation]
    {\includegraphics[scale=0.15]{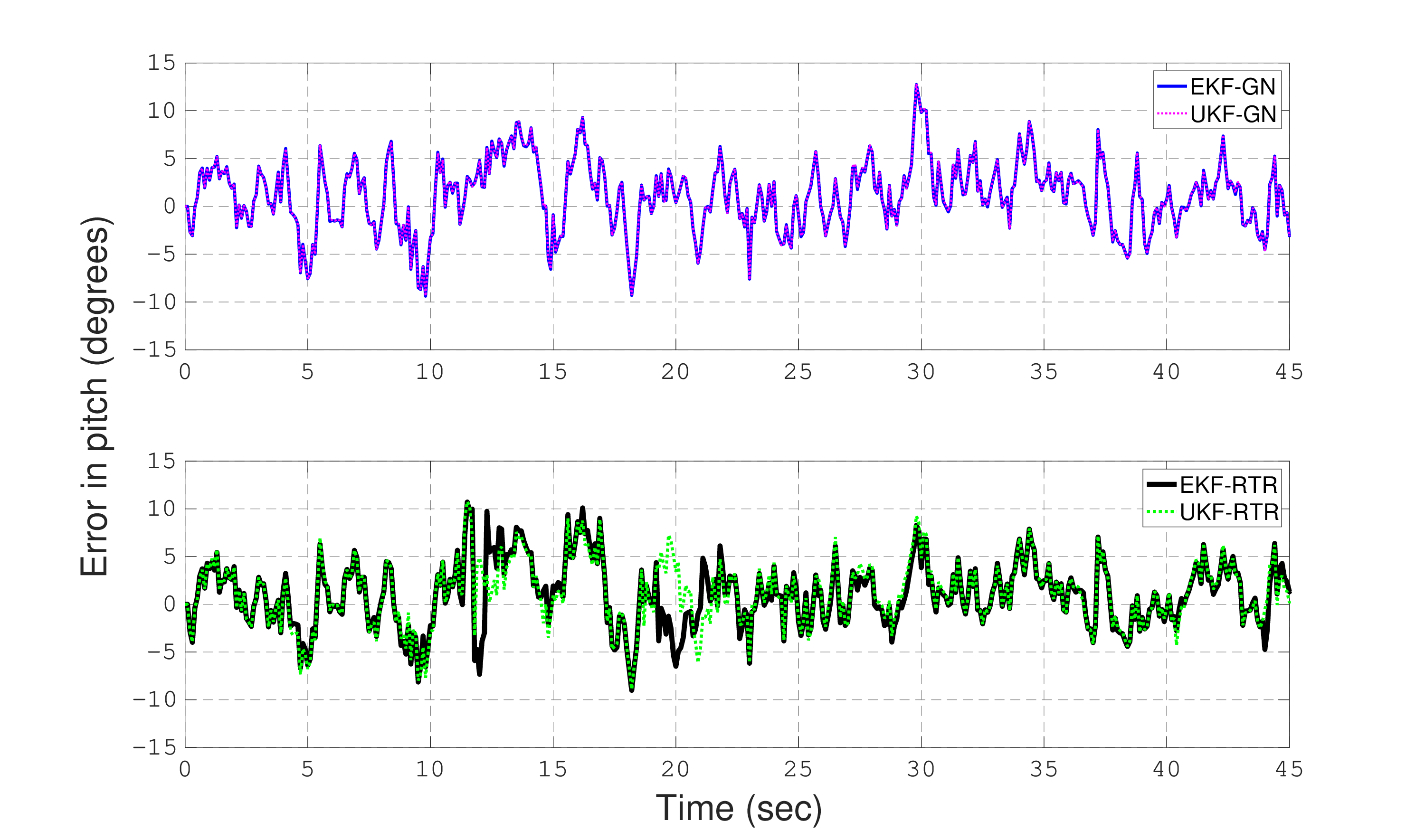}}
\end{minipage}
\begin{minipage}[t]{0.3\textwidth}
\centering
    \subfloat[Error in roll angle in the simulation]
    {\includegraphics[scale=0.15]{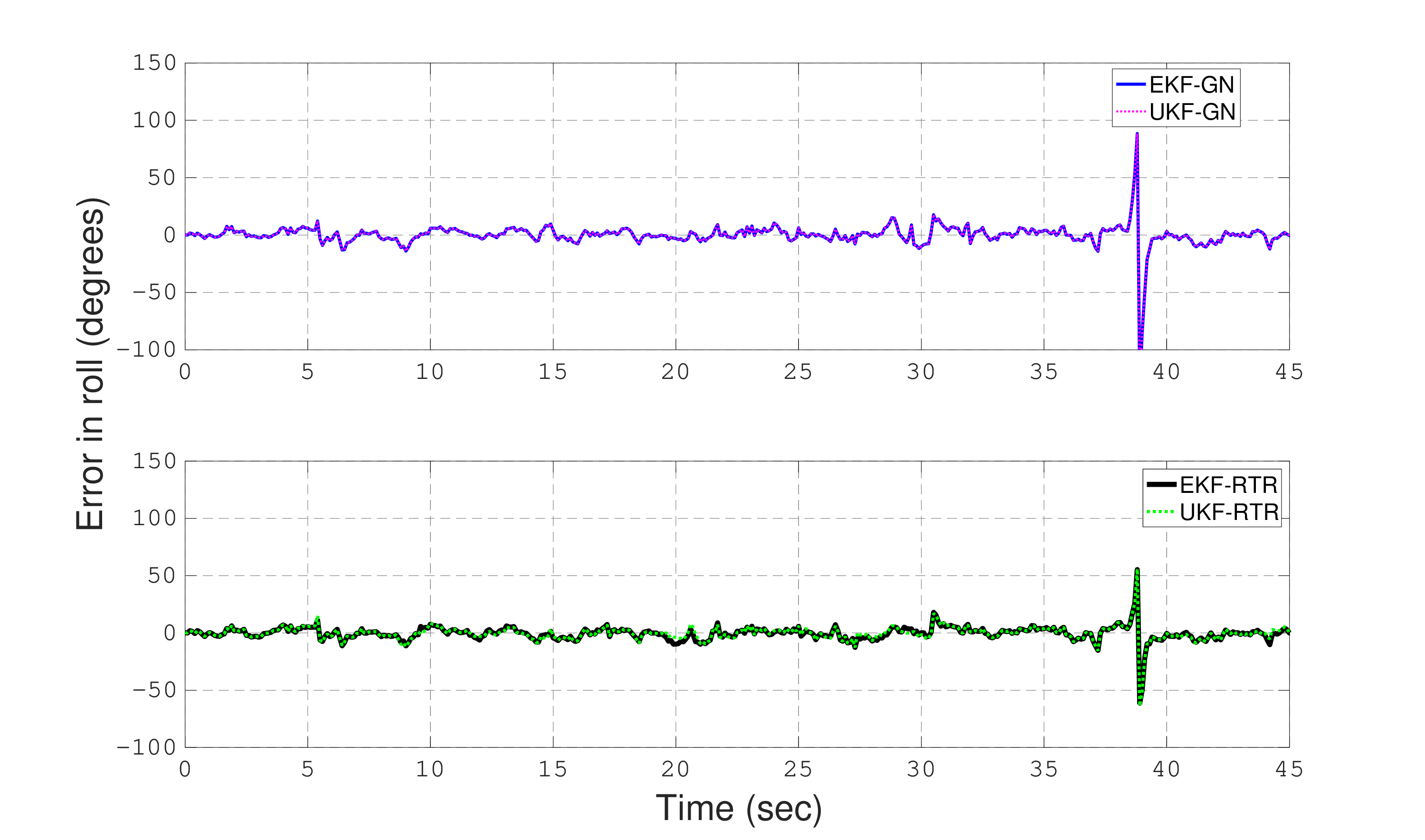}}
\end{minipage}
    \caption{Error in the estimated Euler angles}
    \label{sim_euler_error}
\end{figure}

Orientation-correction methods consistently outperform others under all scenarios, as depicted in Figure \ref{cum_u}. 
Figure \ref{sim_euler_error} shows the error of estimated Euler angles for GN-based and RTR-based algorithms at $\sigma_a = 250 \text{ cm/s}^2$, $\sigma_\omega = 1 \text{ rad/s}$, and $\sigma_d = 3 \text{ cm}$. The RMSE of estimated Euler angles is detailed in Table \ref{euler_table}. During the target’s movement, rotation occurs around all three axes (yaw, pitch, and roll), as seen in Figure \ref{sim_orient}. Algorithms without orientation correction suffer from error accumulation, thus their RMSE in Euler angle estimation is not presented. RTR-based and RSD-based methods show similar performance. Consequently, only RTR-based orientation estimation results are presented. The RTR-based method provides lower RMSE for all three angles, consistent with the RMSE results for vertices positions.	

\begin{table}
\centering
\begin{tabular}{|c|c|c|c|} 
  \hline
  Algorithm & Yaw & Pitch & Roll \\ 
   & RMSE & RMSE  & RMSE  \\ 
  \hline
  EKF-GN & $10.85$ & $4.48$ & $10.66$ \\ 
  \hline
  UKF-GN & $10.87$ & $4.48$ & $10.68$ \\ 
  \hline
  EKF-RTR & $7.80$ & $4.26$ & $7.60$ \\ 
  \hline
  UKF-RTR & $7.79$ & $4.14$ & $7.63$ \\ 
  \hline
\end{tabular}
\caption{RMSE in degrees of the estimated Euler angles}
\label{euler_table}
\end{table}
\subsection{NLOS Scenarios}

\begin{figure}[h!]
    \subfloat[3D motion path]{\includegraphics[scale=0.175,trim={60 0 60 60},clip]{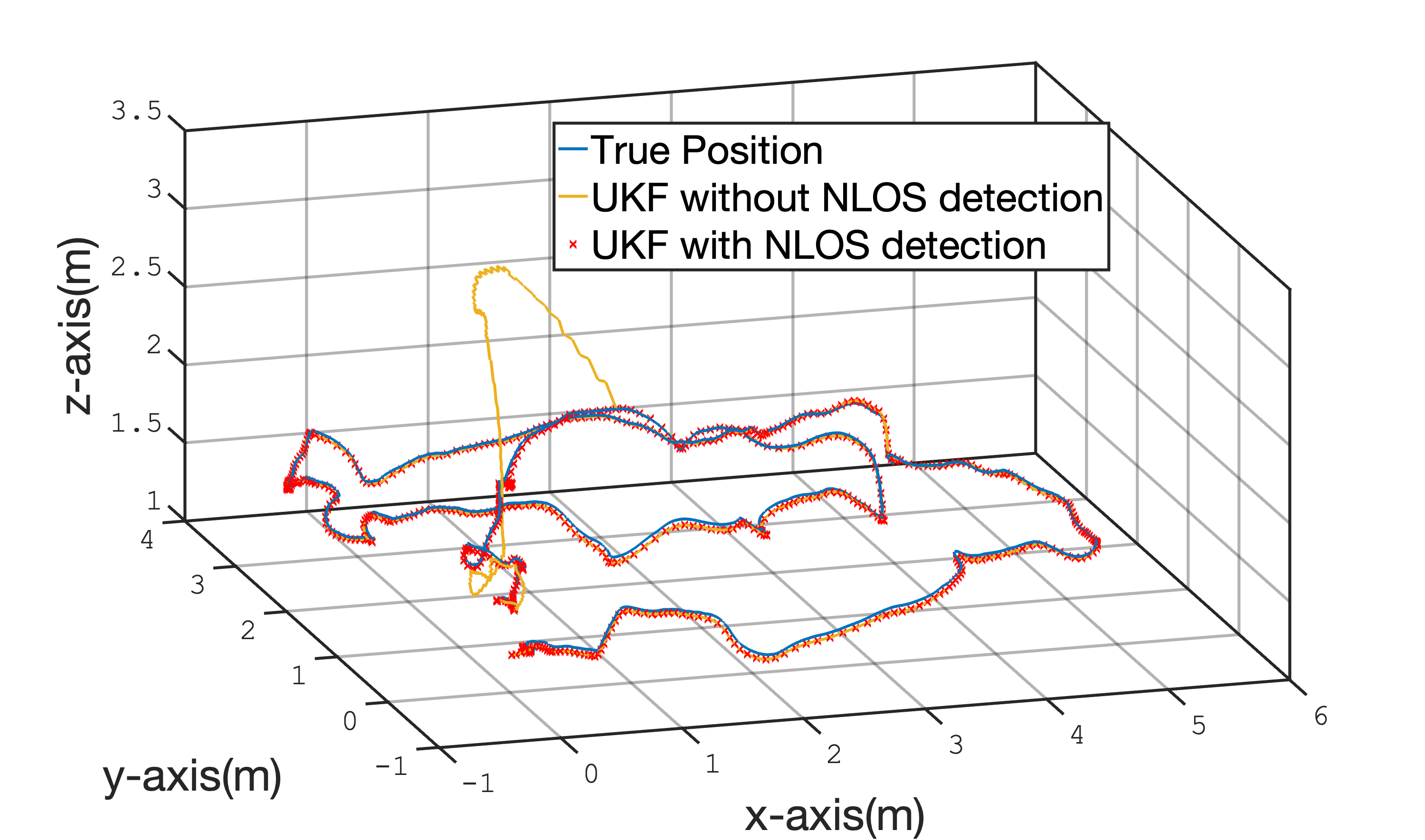}}
    \centering
    \hspace{-2cm}
    \subfloat[Top view]{\includegraphics[scale=0.175,trim={60 10 100 40},clip]{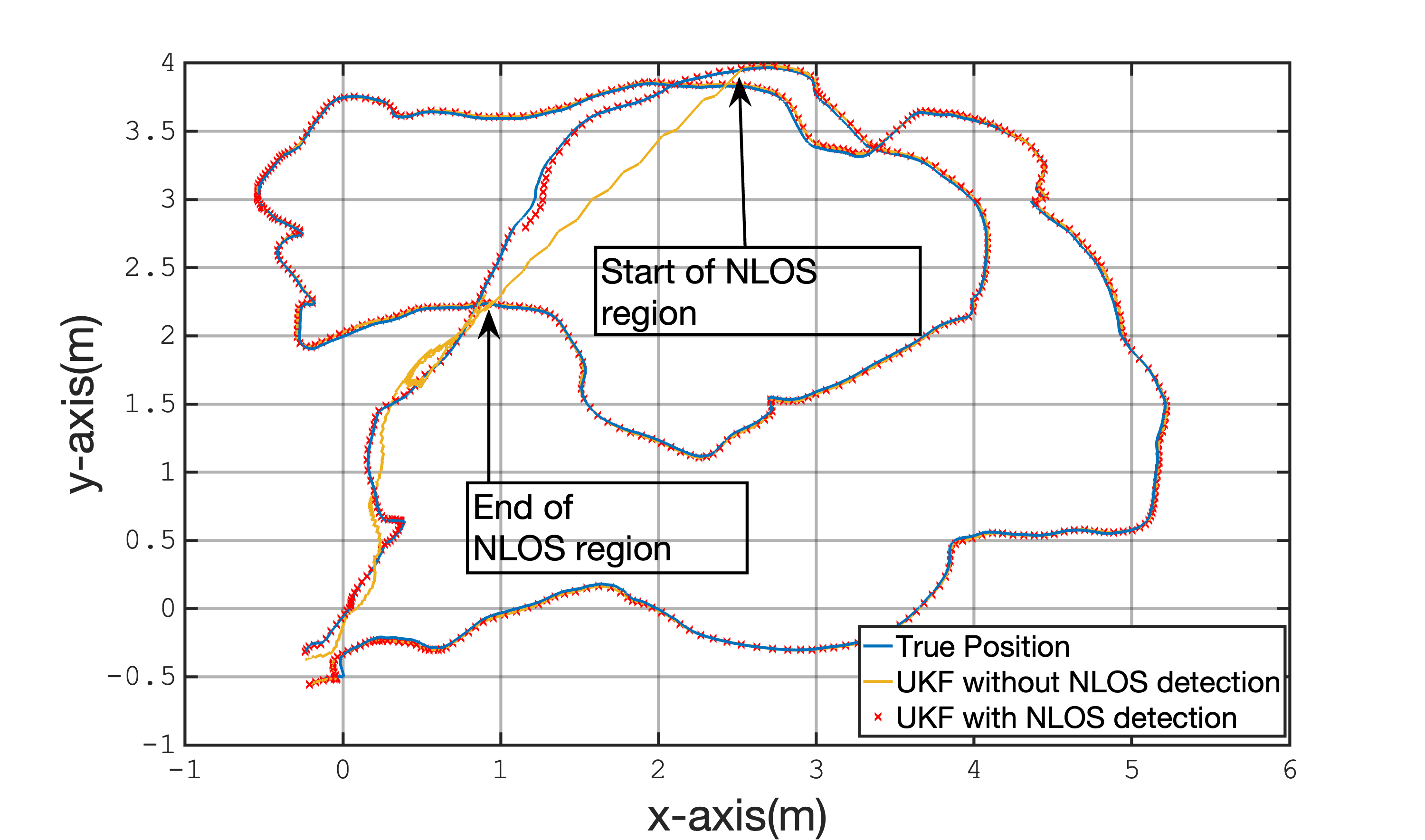}}
    \caption{Motion trajectory in NLOS scenario}
    \label{NLOS_traj}
\end{figure}

\begin{figure}[!h]
    \centering
    \includegraphics[scale=0.17]{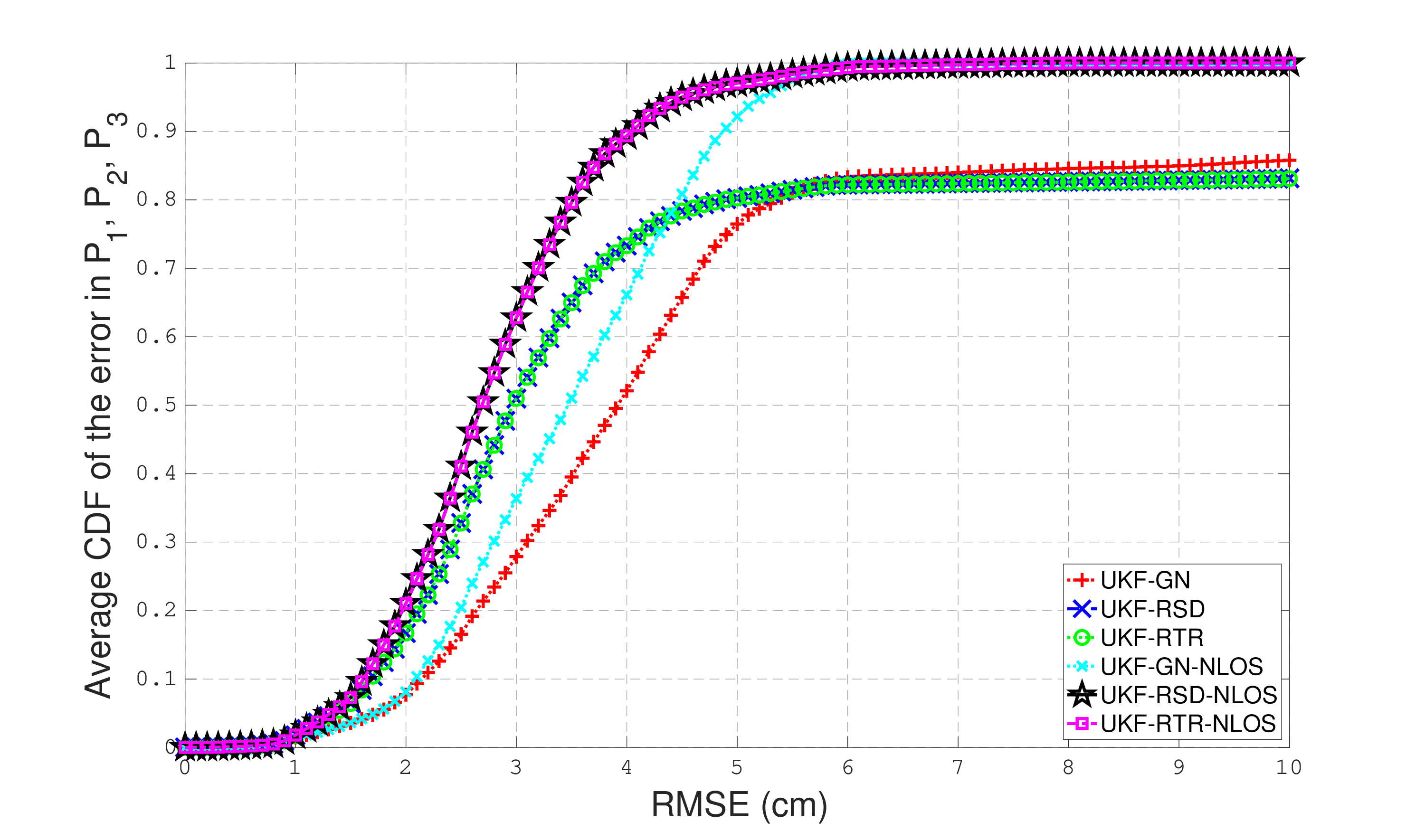}
    \caption{Average CDF at $\sigma_a = 5$ cm/s$^2$, $\sigma_{\omega} = 0.25$ rad/s, and $\sigma_d = 2.5$ cm}
    \label{cum_nlos}
\end{figure}

To simulate NLOS scenarios, distances between the three receivers and the first beacon are generated based on a uniform distribution over a $3$-second period, simulating a beacon blockage during this time. We will compare two situations. In the first situation, we input the distances between the three receivers and four beacons, along with synthetic IMU measurements, into the EKF and UKF algorithms and assess the accuracy of the estimated positions. In the second situation, we first determine whether the estimated distance is from an LOS or NLOS condition, using various NLOS detection algorithms \cite{alsharif2021robust}, \cite{zhu2023robust}. Subsequently, we apply the EKF and UKF algorithms only to the distances identified as LOS measurements. In the period of NLOS distances, the positioning rely completly on the synthetic IMU data. Figure \ref{NLOS_traj} illustrates the estimated position and motion trajectory, while Figure \ref{cum_nlos} presents the average CDF of RMSE for the three vertices, with parameters set at $\sigma_a = 5$ cm/s$^2$, $\sigma_{\omega} = 0.25$ rad/s, and $\sigma_d = 2.5$ cm. The proposed algorithms maintain their accuracy in NLOS scenarios, provided that NLOS detection is effective. Conversely, performance degrades in algorithms that do not incorporate NLOS detection.

\subsection{Semi-Experimental Results}
\begin{figure}[h!]
    \subfloat[3D motion path]{\includegraphics[scale=0.15]{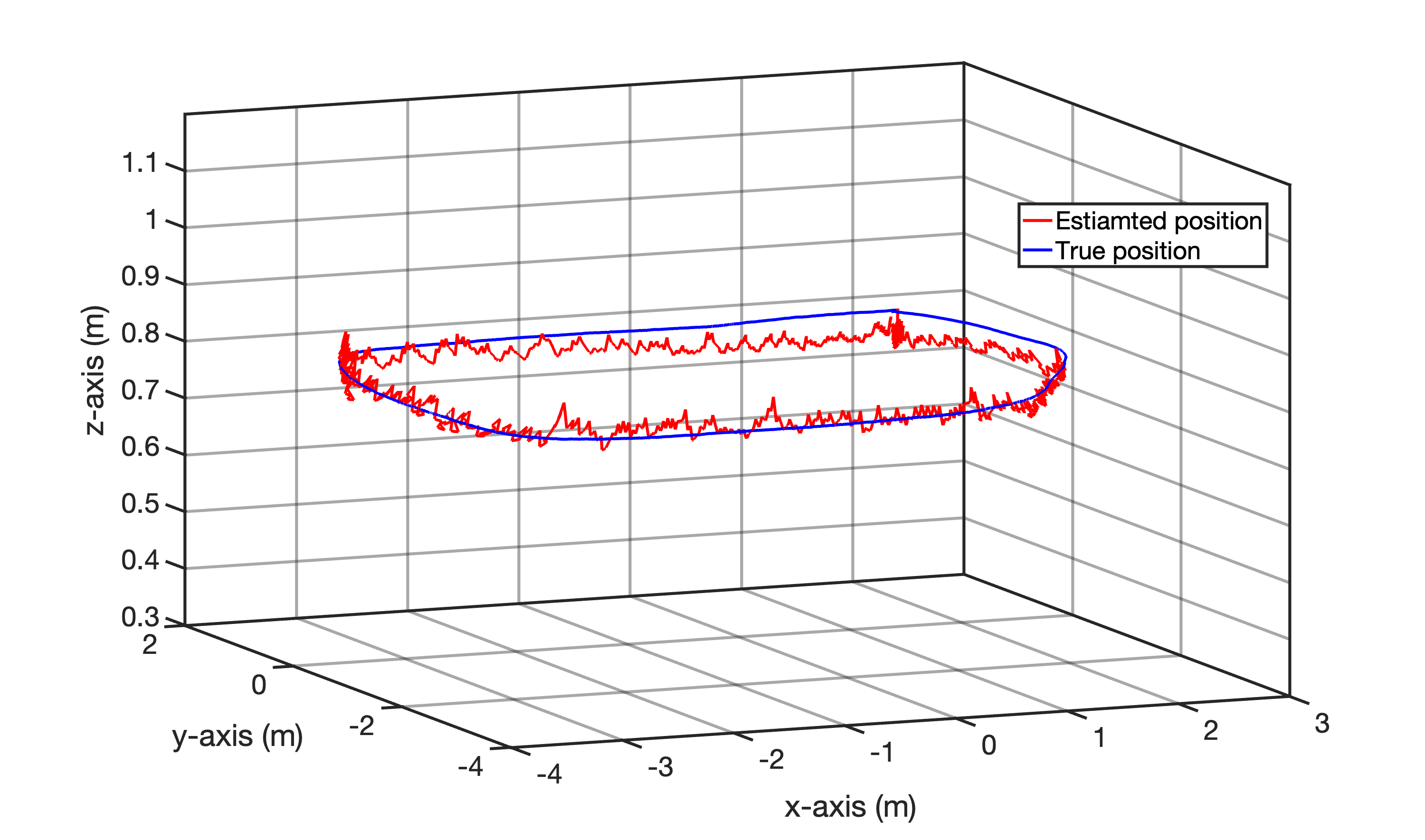}}
    \centering
    \hspace{-2cm}
    \subfloat[Top view]{\includegraphics[scale=0.15]{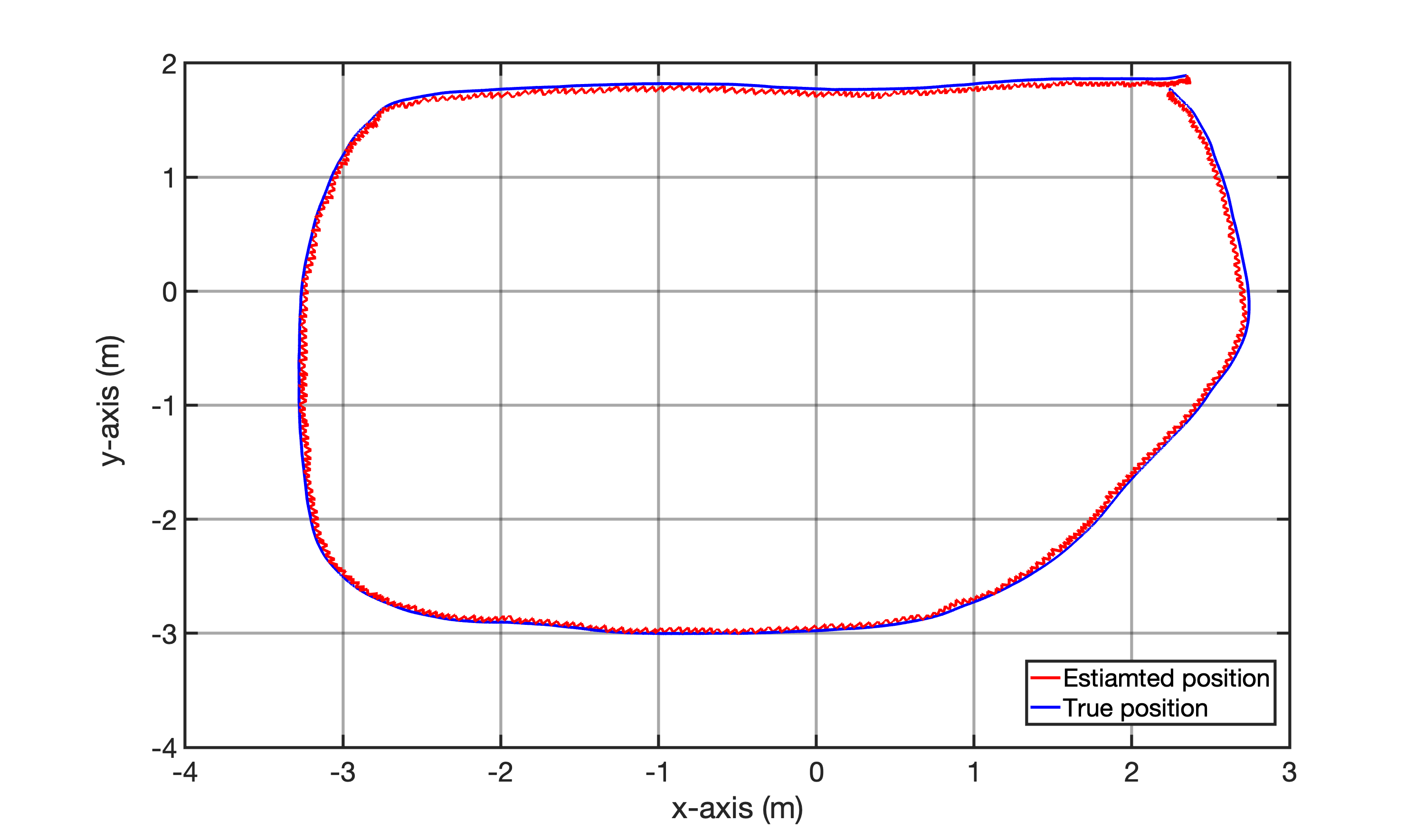}}
    \caption{iPhone's Experiment: Motion trajectory}
    \label{motion_traj_iphone}
\end{figure}

\begin{figure}[h!]
\begin{minipage}[t]{0.33\textwidth}    
\centering
\subfloat[Yaw angle during iPhone experiment]
    {\includegraphics[scale=0.15]{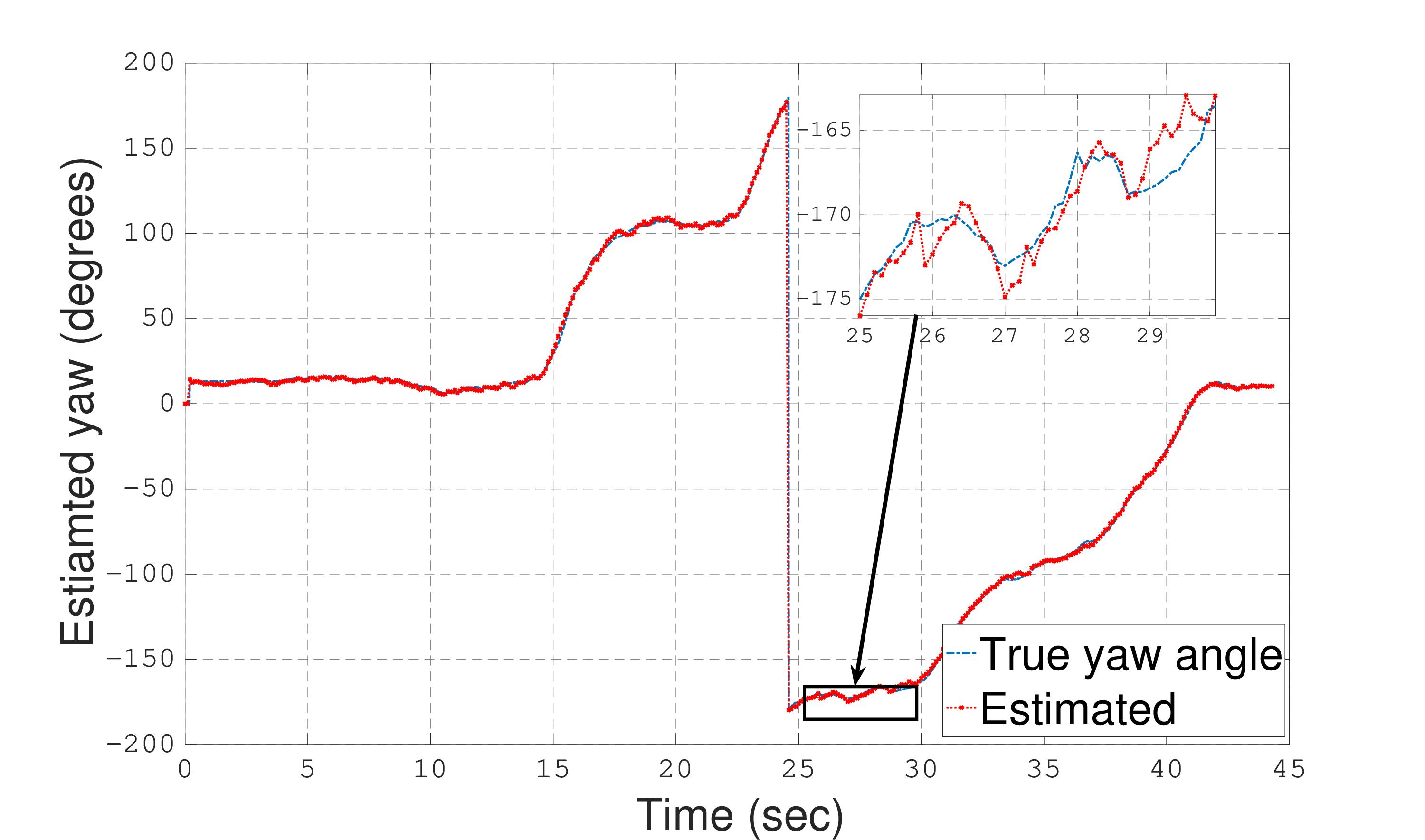}}
\end{minipage}
\begin{minipage}[t]{0.33\textwidth} 
\centering
\subfloat[Pitch angle during iPhone experiment]
    {\includegraphics[scale=0.15]{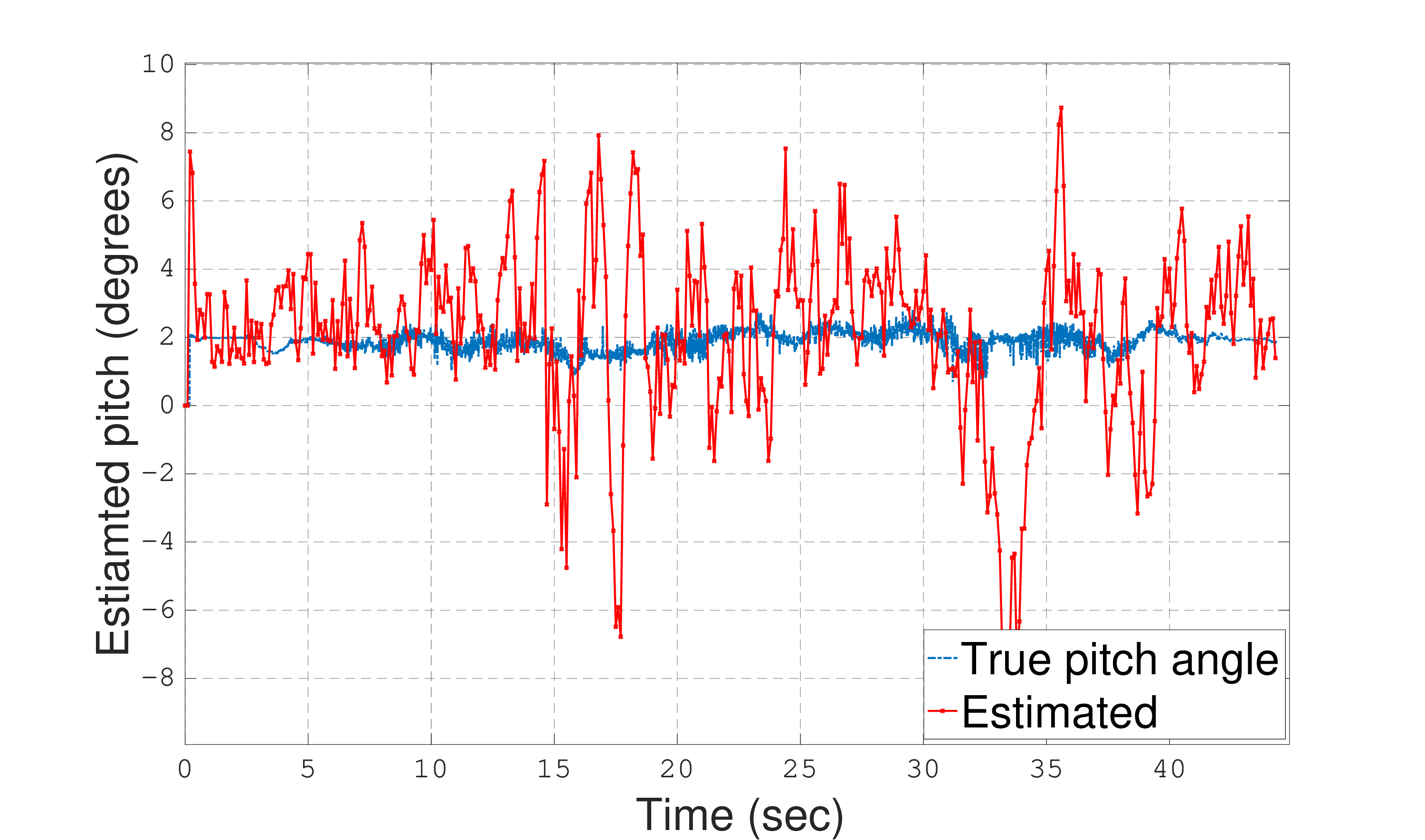}}
\end{minipage}
\begin{minipage}[t]{0.33\textwidth}  
\centering
\subfloat[Roll angle during iPhone experiment]
    {\includegraphics[scale=0.15]{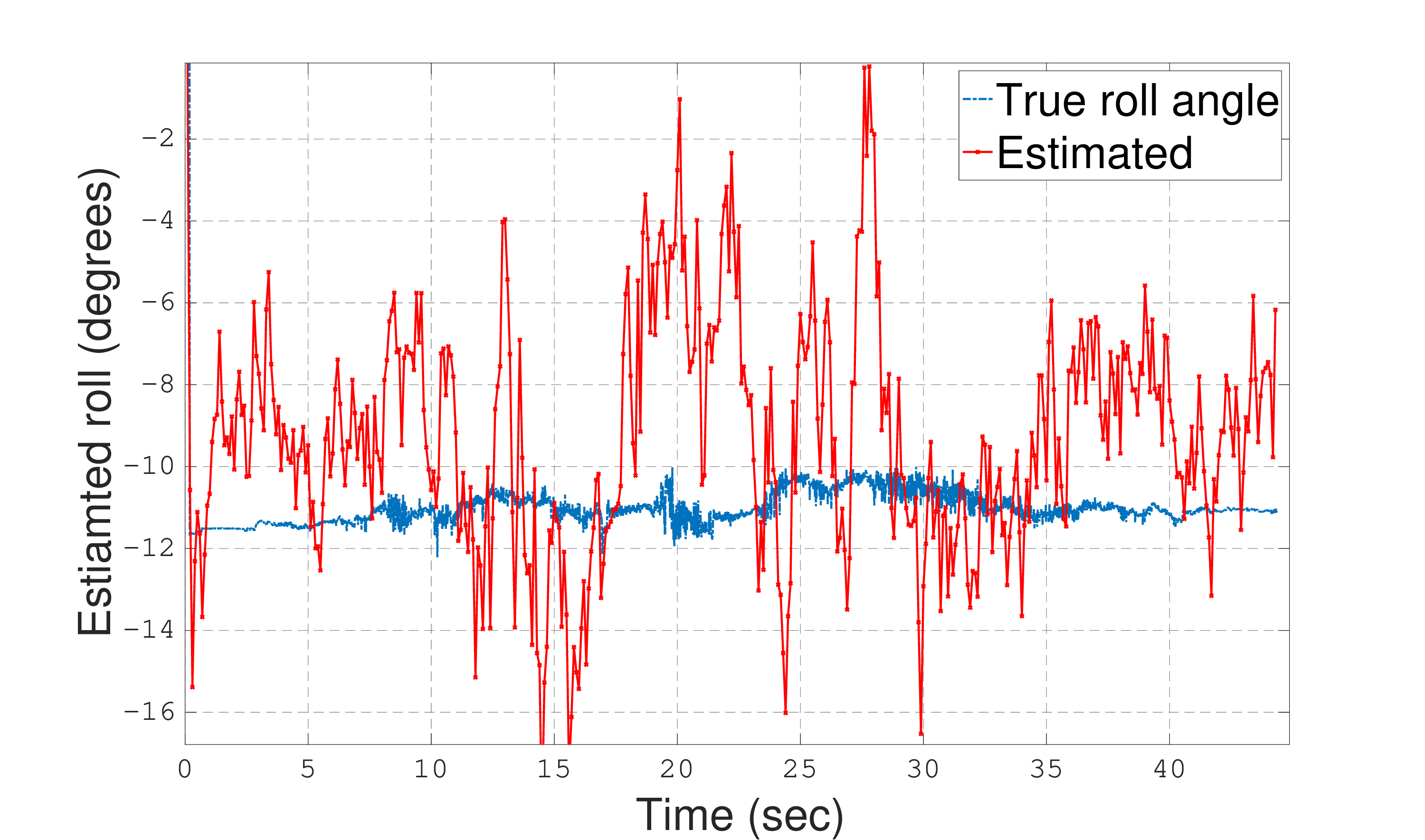}}
\end{minipage}
    \caption{iPhone's Experiment: Orientation of the phone}
    \label{iphone_orient}
\end{figure}

In this semi-experimental setup, we utilized real IMU measurements from an iPhone combined with synthetic distance measurements obtained using the OptiTrack system. Markers, as shown in Figure 11(a), were attached to the iPhone to provide ground-truth position and orientation data through the OptiTrack system. We then applied both proposed and benchmark methods to estimate the position and orientation of the iPhone. The RMSE of the three vertices of an equilateral triangle was calculated. The motion trajectory for this experiment is depicted in Figure \ref{motion_traj_iphone}, and the true and estimated Euler angles of the phone are shown in Figure \ref{iphone_orient}.

The noise characteristics of the iPhone’s gyroscope and accelerometer were as follows: the gyroscope noise had standard deviations of $0.11$ rad/s (x-axis), $0.05$ rad/s (y-axis), and $0.07$ rad/s (z-axis). The accelerometer noise had standard deviations of $53$ cm/s$^2$ (x-axis), $62$ cm/s$^2$ (y-axis), and $49$ cm/s$^2$ (z-axis). These parameters were estimated from the iPhone’s IMU measurements and the synthetic IMU data provided by the OptiTrack system.

\begin{figure}[!h]
    \centering
    \includegraphics[scale=0.17]{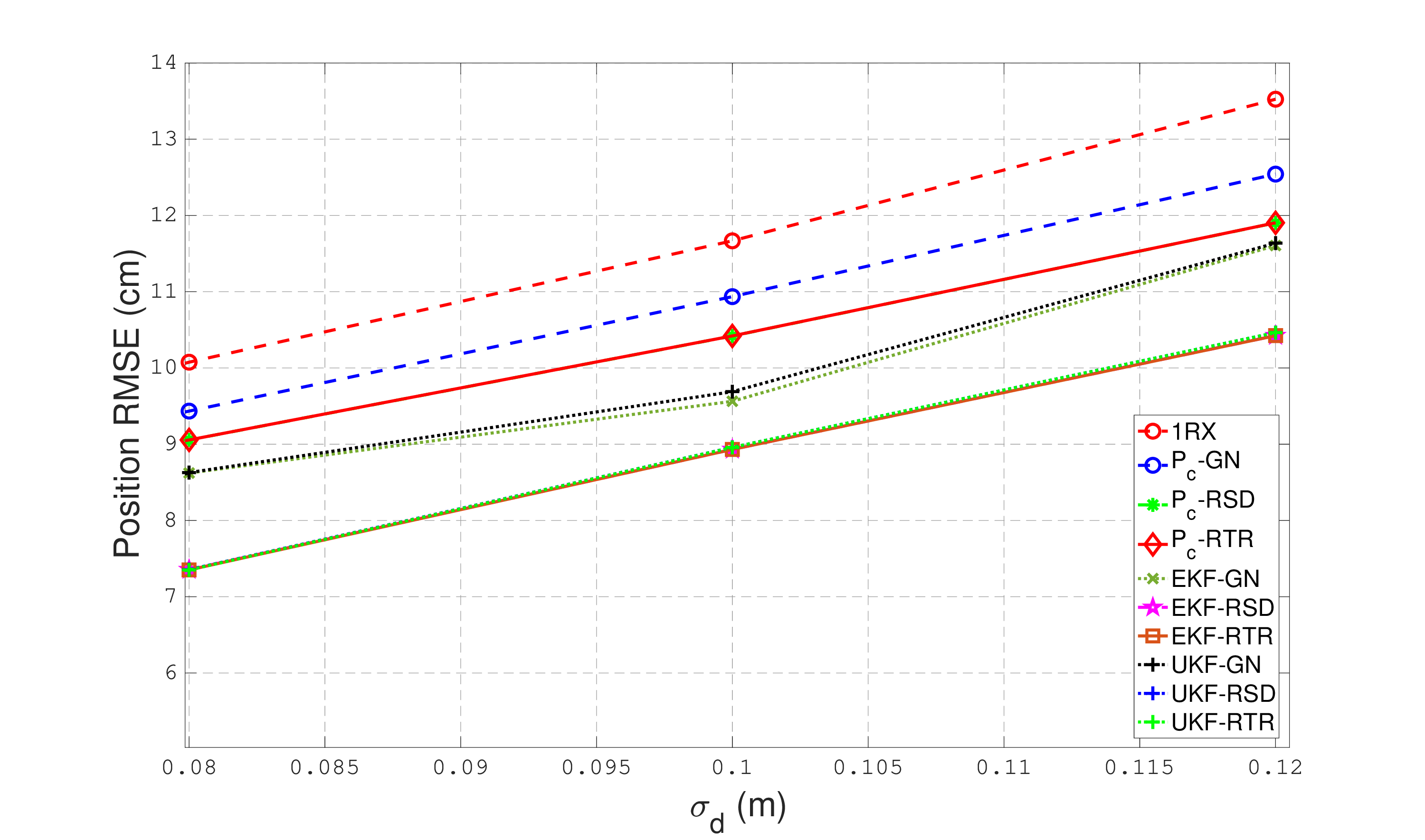}
    \caption{iPhone's Experiment: RMSE vs $\sigma_d$}
    \label{pc_v_snr_d_iphone}
\end{figure}

The experimental results are consistent with the previous numerical findings. The average RMSE of the three vertices' positions for all algorithms increases with the standard deviation of the distance measurements $\sigma_d$, as shown in Figure \ref{pc_v_snr_d_iphone}. Additionally, using three receivers rather than a single receiver improves the accuracy of the estimated position and orientation. The benefit of the proposed orientation-correction method using three receivers is evident, with Riemannian-based Kalman filtering methods outperforming benchmark algorithms in terms of vertices position accuracy.

\begin{figure}[t!]
    \centering
    \includegraphics[scale=0.17]{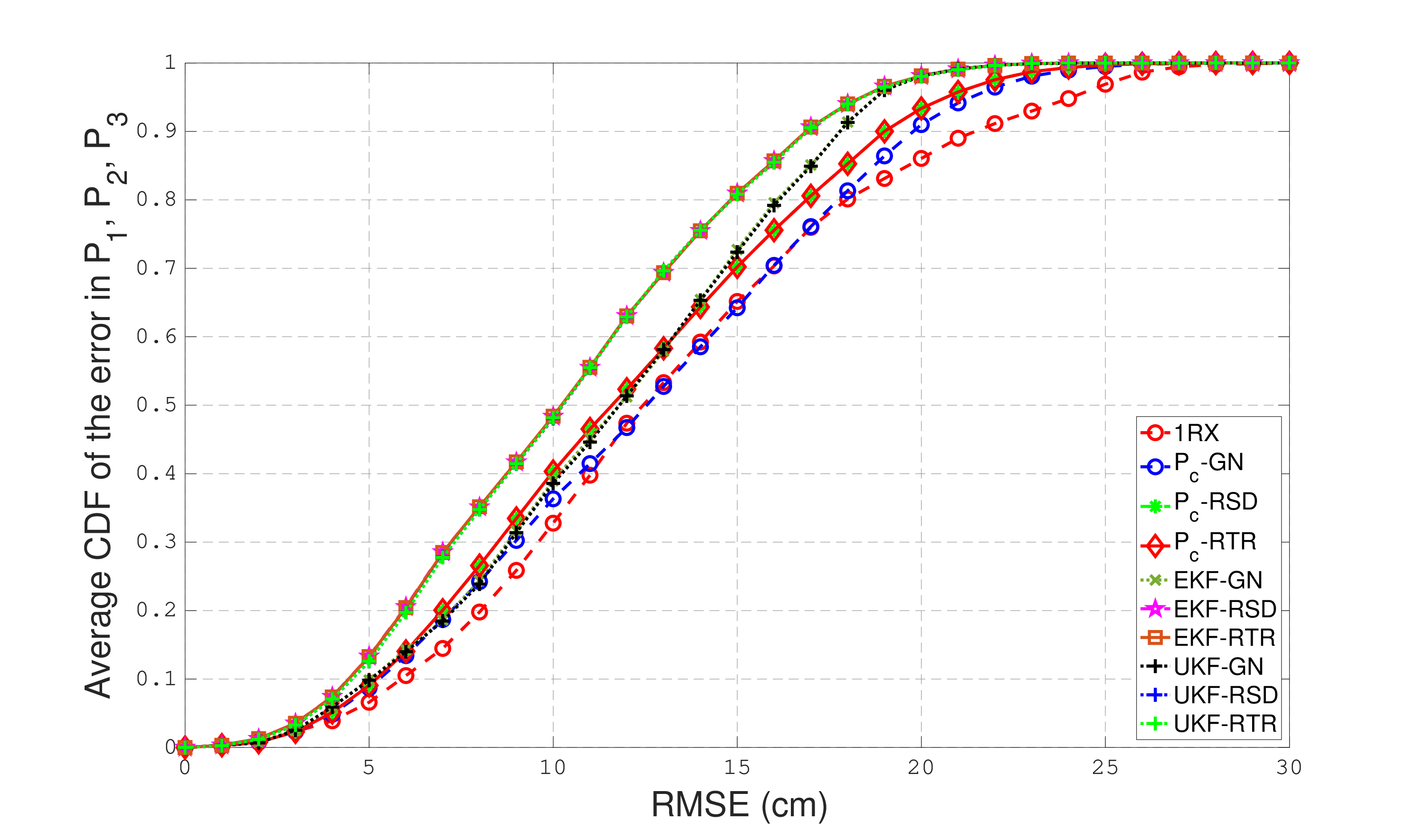}
    \caption{iPhone's Experiment: Average CDF at $\sigma_a = 55 \text{ cm/s}^2$, $\sigma_\omega = 0.08 \text{ rad/s}$, $\sigma_d = 12 \text{ cm}$}
    \label{cumm_d_iphone}
\end{figure}

The average cumulative distribution function (CDF) of the RMSEs for the three vertices is presented in Figure \ref{cumm_d_iphone}. Algorithms that do not correct the orientation obtained from INS exhibit lower accuracy, as expected. Furthermore, RTR and RSD methods outperform GN-based methods. The performance of EKF and UKF is similar due to the slow movement relative to the high update rate, which makes the first-order approximation of EKF accurate.

\begin{figure}
\begin{minipage}[t]{0.3\textwidth}  
\centering
\subfloat[Yaw angle during iPhone experiment]
    {\includegraphics[scale=0.15]{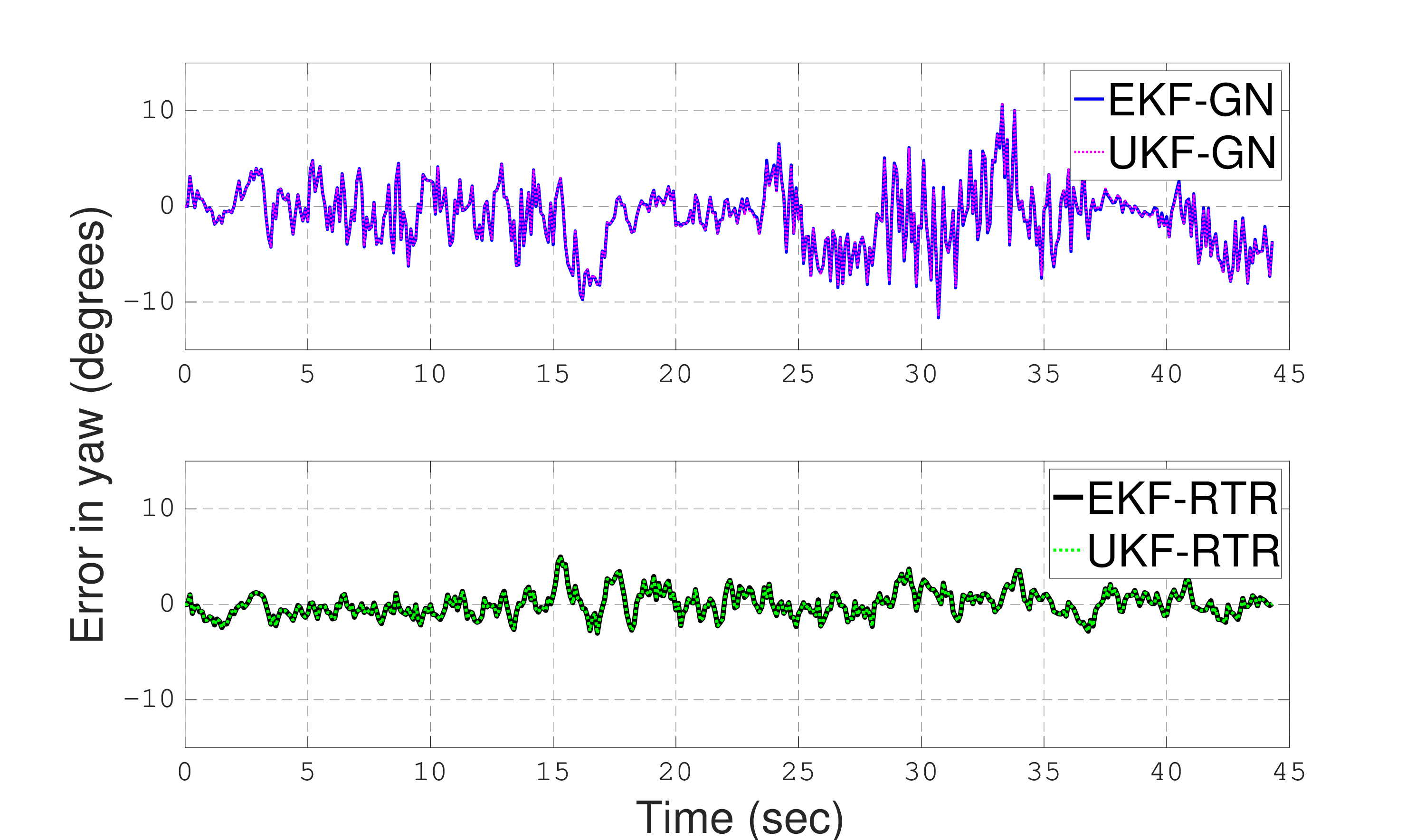}}
\end{minipage}
\begin{minipage}[t]{0.3\textwidth}    
\centering
\subfloat[Pitch angle during iPhone experiment]
    {\includegraphics[scale=0.15]{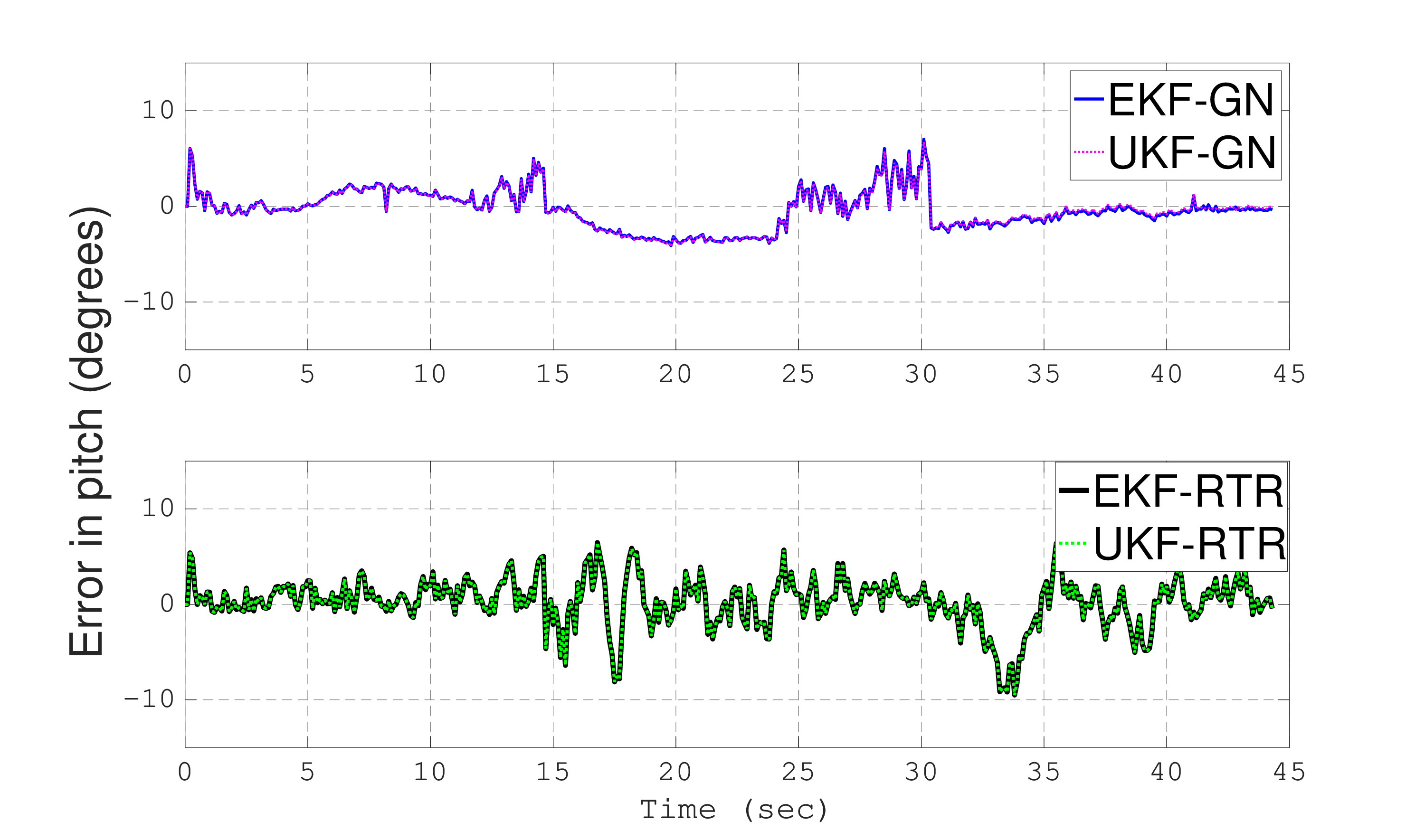}}
\end{minipage}
\begin{minipage}[t]{0.3\textwidth}    
\centering
\subfloat[Roll angle during iPhone experiment]
    {\includegraphics[scale=0.15]{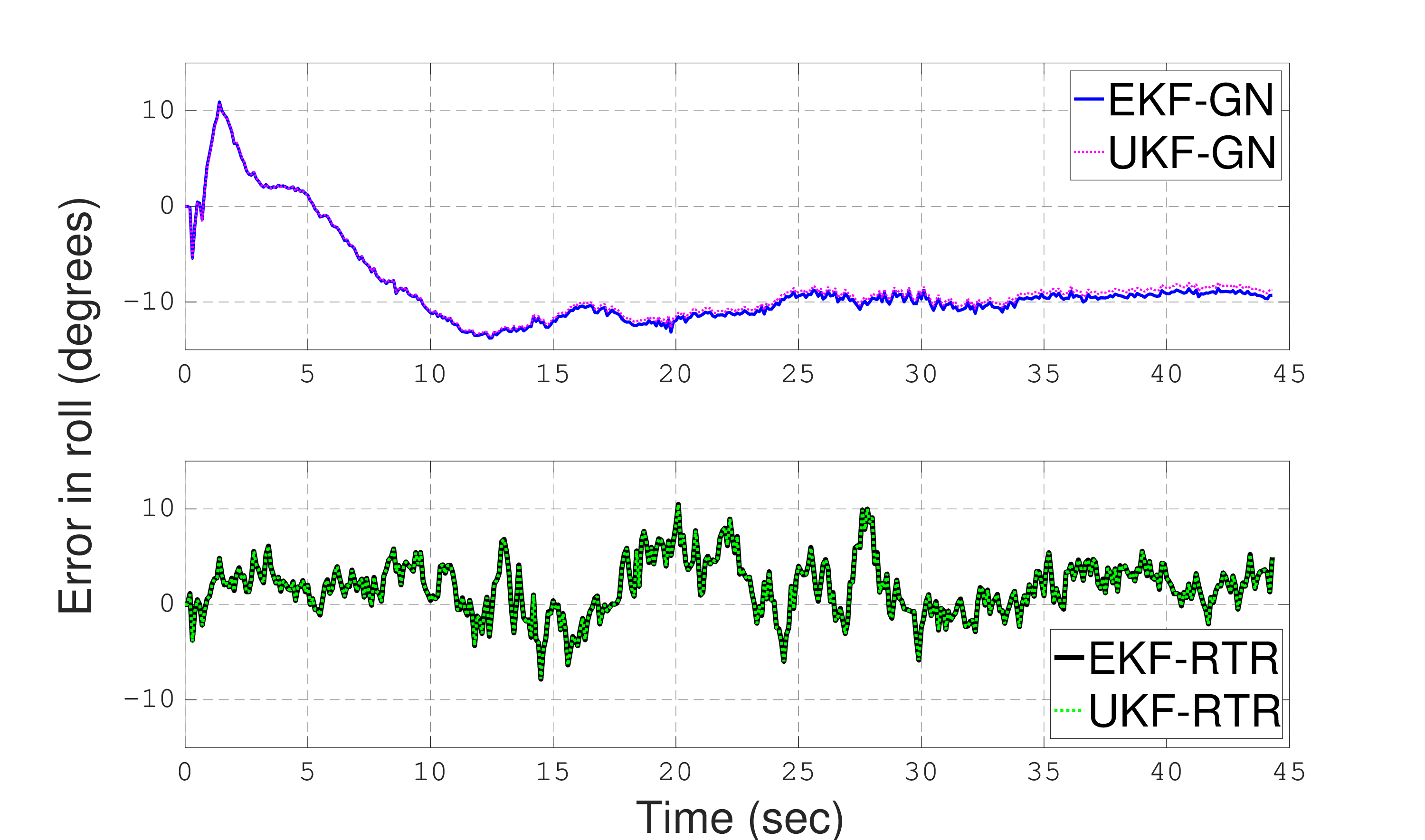}}
\end{minipage}
    \caption{iPhone's experiment: Error in the estimated Euler angles}
    \label{iphone_euler_error}
\end{figure}

Figure \ref{iphone_euler_error} shows the error in the estimated Euler angles for both GN-based and RTR-based algorithms, and Table \ref{euler_table2} provides the RMSE of the estimated Euler angles. During the experiment, the rotation primarily occurred around the yaw axis, as illustrated in Figure \ref{iphone_orient}. For this reason, we present only the results for GN-based and RTR-based orientation estimation. The RTR-based method demonstrates lower RMSE in estimating yaw and roll angles and comparable RMSE for the pitch angle. This outcome is consistent with the evaluation of the RMSE of the three vertices, reflecting the accuracy of orientation estimation.

\begin{table}[t!]
\centering
\begin{tabular}{|c|c|c|c|} 
  \hline
  Algorithm & Yaw & Pitch & Roll \\ 
   & RMSE & RMSE  & RMSE  \\ 
  \hline
  EKF-GN & $3.49$ & $2.07$ & $9.70$ \\ 
  \hline
  UKF-GN & $3.47$ & $2.03$ & $9.34$ \\ 
  \hline
  EKF-RTR & $1.35$ & $2.52$ & $3.34$ \\ 
  \hline
  UKF-RTR & $1.35$ & $2.52$ & $3.34$ \\ 
  \hline
\end{tabular}
\caption{RMSE in degrees of the estimated Euler angles}
\label{euler_table2}
\end{table}

\subsection{Experimental Results: MTi-1 with Acoustic Transmitters/ Receivers}

\begin{figure}[h!]
    \subfloat[3D motion path]{\includegraphics[scale=0.175,trim={40 0 0 60},clip]{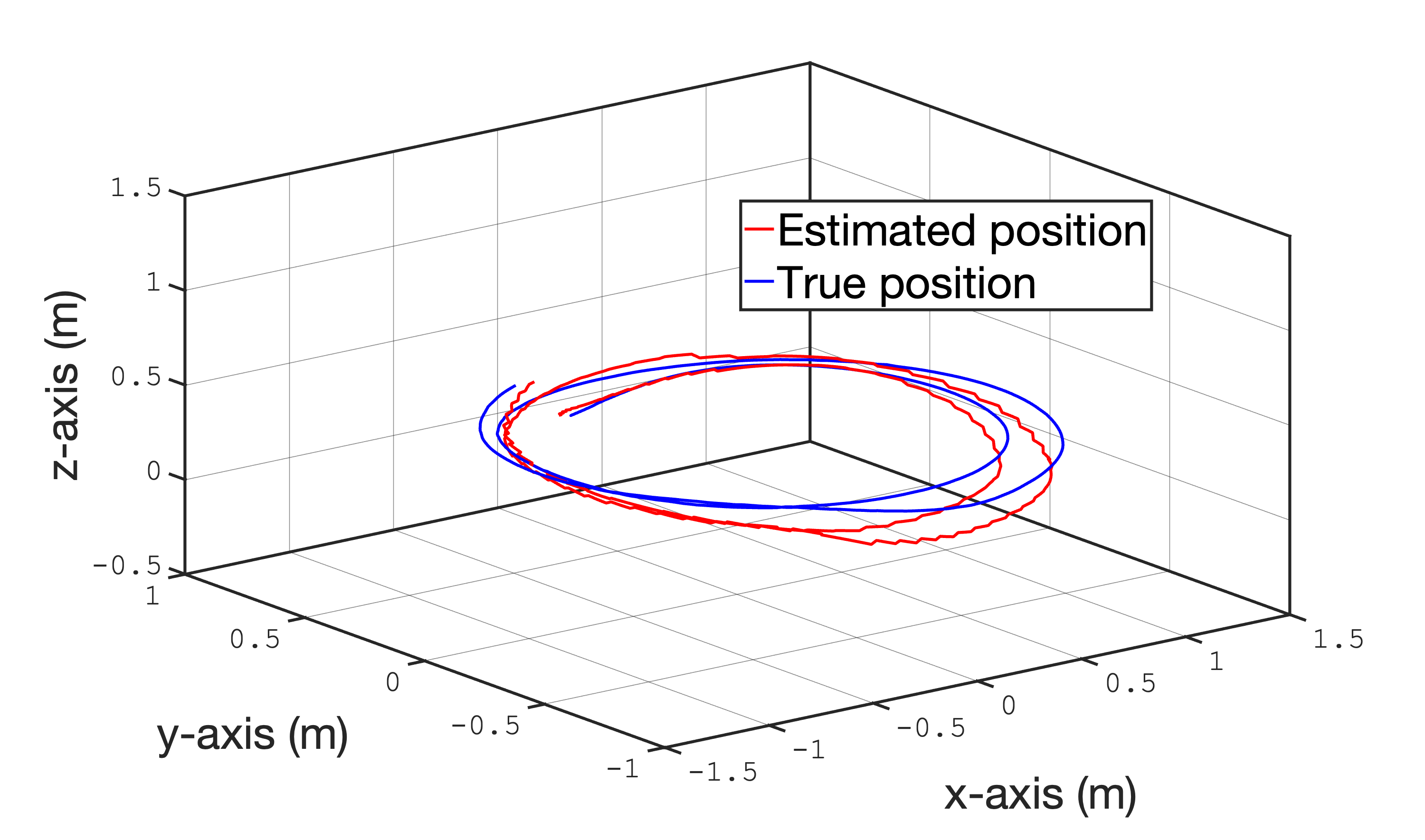}}
    \centering
    \hspace{-2cm}
    \subfloat[Top view]{\includegraphics[scale=0.175,trim={60 10 100 40},clip]{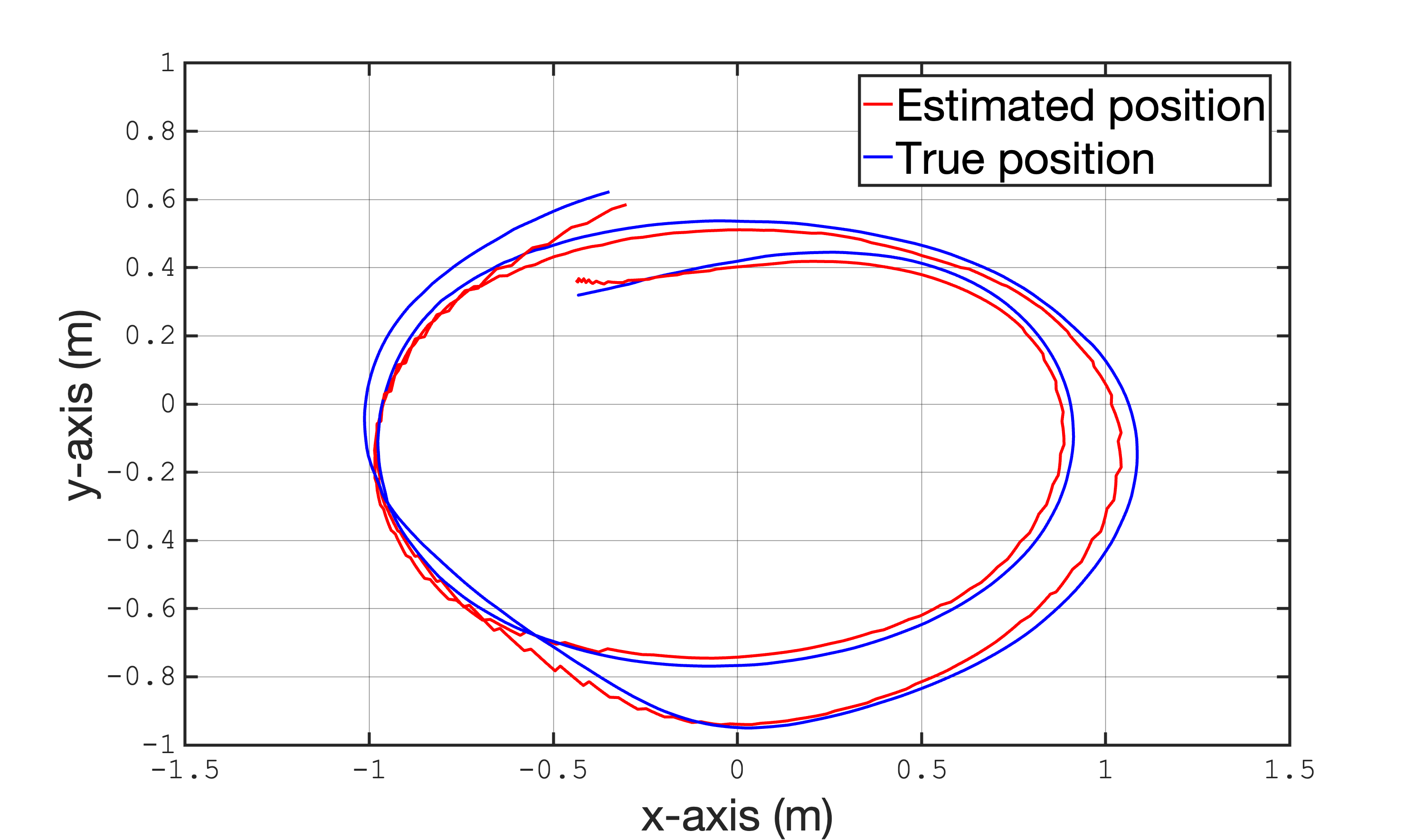}}
    \caption{Experimental evaluation: Motion trajectory}
    \label{motion_traj_mti}
\end{figure}

In this experiment, four acoustic transmitters (beacons) were positioned at known locations, while three acoustic receivers, arranged in an equilateral triangle on a printed circuit board, were fixed to an MD as shown in Figure \ref{exp_set}. The MD was equipped with an MTi-1 IMU, which includes a triaxial accelerometer and a triaxial gyroscope \cite{MTi}. The device was moved around the room, as depicted in Figure \ref{motion_traj_mti}. The IMU sampled at $100$ Hz, and the acoustic system measured distances between the transmitters and receivers at $5$ Hz. Synchronization between the acoustic transmitters and receivers was achieved using an RF signal. The experiment lasted $30$ seconds, and the recorded data were processed offline. 

Three variations of Kalman filters were evaluated as follows. A linear Kalman filter (LKF) where the state vector includes the centroid position and velocity of the MD. The observed variable consists of the positions of the three vertices of the equilateral triangle, which are estimated using the GN and RTR methods. The centroid position is obtained by averaging the three vertex positions and then used in the LKF. The other two variations of the Kalman filter were implemented using EKF and UKF where the state variable includes the centroid position, velocity, and the vector-form of the orientation matrix. In this case, the observed variable is the vertices' positions, which are utilized to estimate both the orientation matrix and the centroid position.

\begin{figure}[!h]
    \centering
    \includegraphics[scale=0.175]{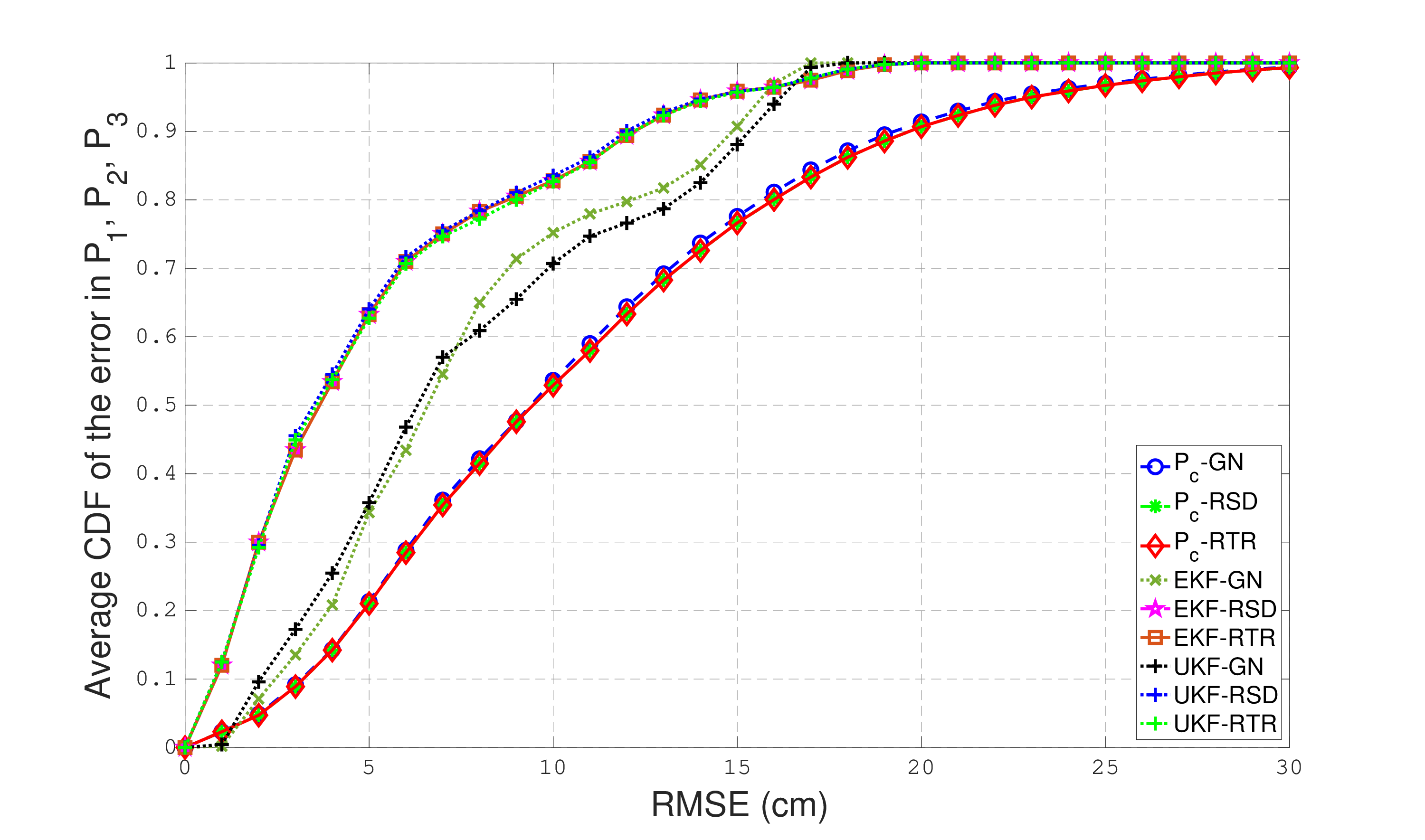}
    \caption{Experimental evaluation: Average CDF at $\sigma_a = 55 \text{ cm/s}^2$, $\sigma_\omega = 0.08 \text{ rad/s}$, $\sigma_d = 12 \text{ cm}$}
    \label{cumm_d_mti}
\end{figure}

The RMSE of the estimated positions, averaged over the three vertices and across time, is presented in Table \ref{rmse_table_exp}. The Riemannian-based UKF and EKF demonstrate improved accuracy compared to GN-based methods. This improvement is also reflected in the CDF plot shown in Figure \ref{cumm_d_mti}, where more than $80\%$ of positions estimated using Riemannian-based UKF and EKF have an error of less than $9$ cm, whereas less than $80\%$ of positions estimated using GN-based UKF and EKF have an error of less than $13$ cm.
 
\begin{table}[H]
\centering
\begin{tabular}{|c|c|} 
  \hline
  Algorithm & RMSE (cm) \\ 
   \hline
  LKF-GN & $10.44$ \\ 
  \hline
   LKF-RTR & $10.60$ \\ 
  \hline
  EKF-GN & $7.52$ \\ 
  \hline
  UKF-GN & $7.67$\\ 
  \hline
  EKF-RTR & $5.09$ \\ 
  \hline
  UKF-RTR & $5.00$ \\ 
  \hline
\end{tabular}
\caption{RMSE in cm averaged over the three receivers}
\label{rmse_table_exp}
\end{table}

\begin{figure}[h!]
\begin{minipage}[t]{0.3\textwidth}
\centering
    \subfloat[Yaw angle during  experiment]
    {\includegraphics[scale=0.15]{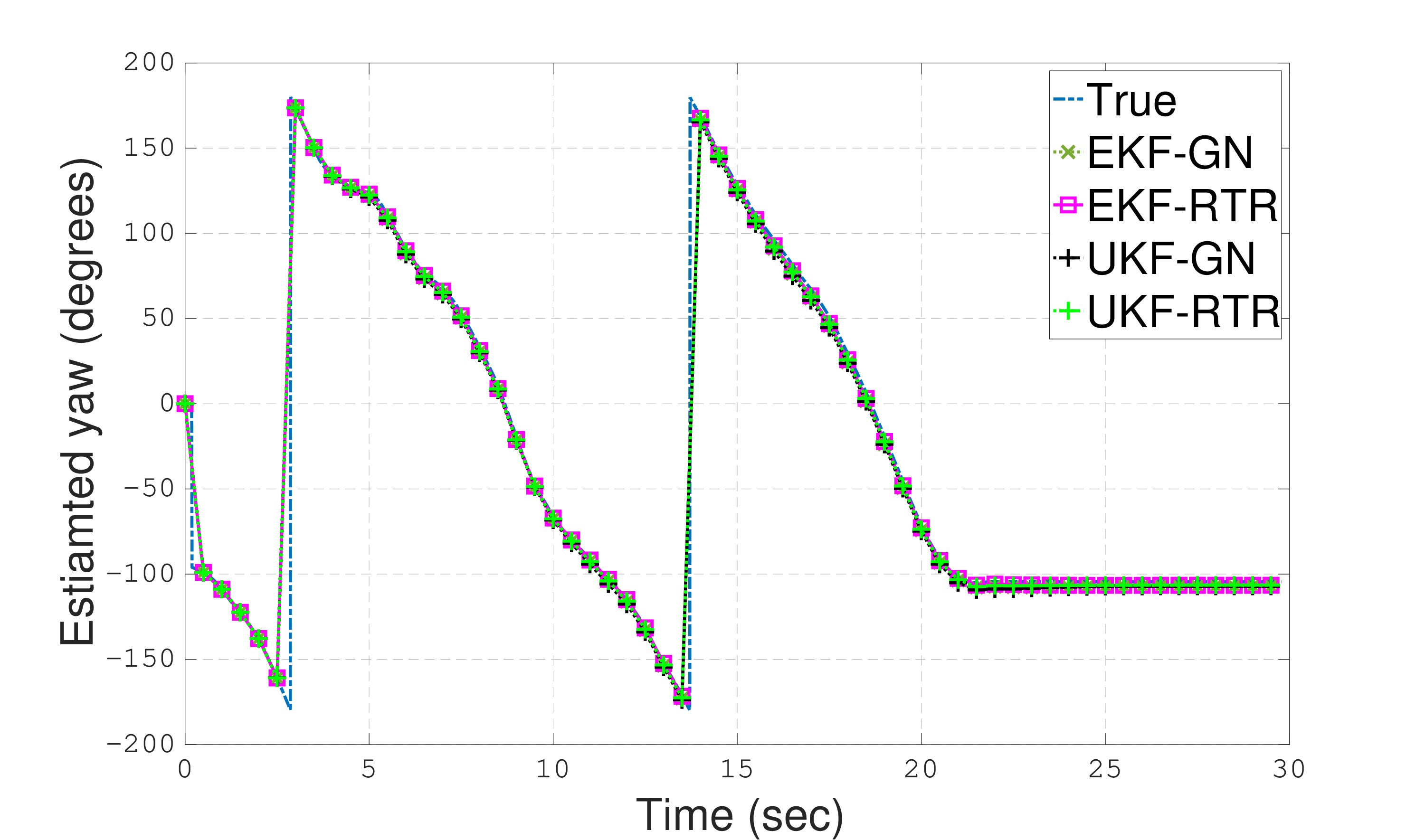}}
\end{minipage}
\begin{minipage}[t]{0.3\textwidth}
\centering
    \subfloat[Pitch angle during experiment]
    {\includegraphics[scale=0.15]{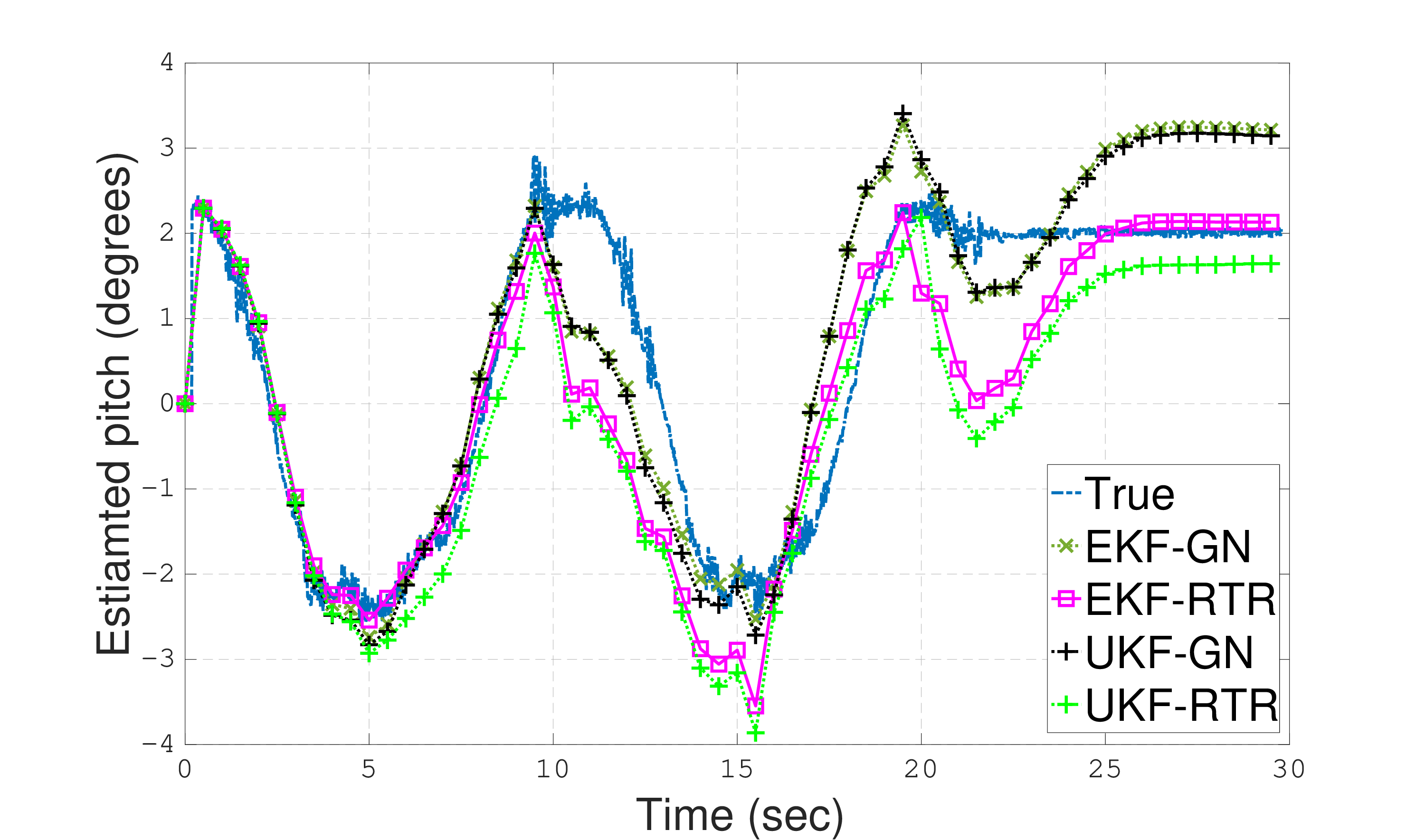}}
\end{minipage}
\begin{minipage}[t]{0.3\textwidth}
\centering
    \subfloat[Roll angle during experiment]
    {\includegraphics[scale=0.15]{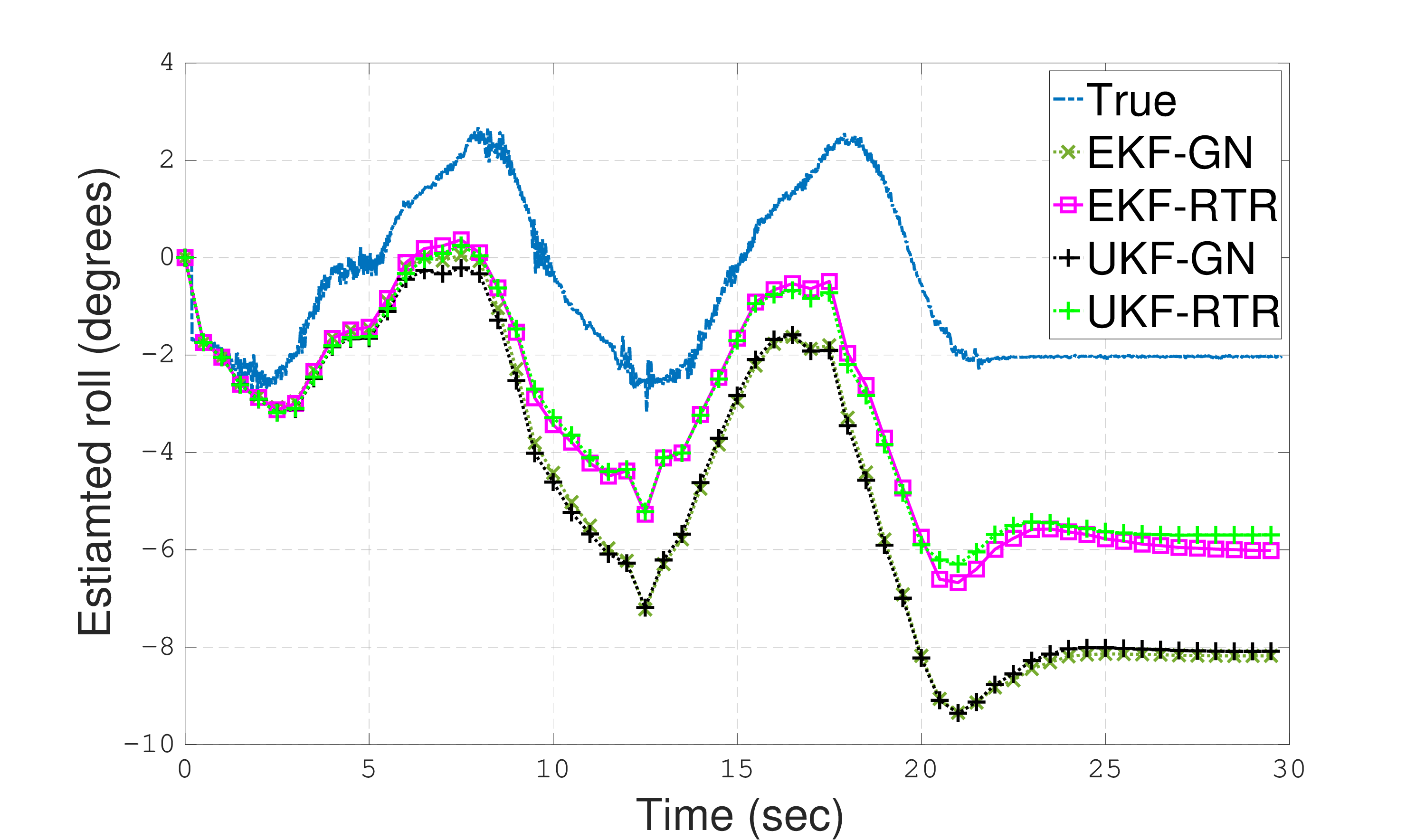}}
\end{minipage}
    \caption{Experimental evaluation: orientation of the MD}
    \label{mti_orient}
\end{figure}

The accuracy of the estimated Euler angles for the UKF and EKF methods was also evaluated. The estimated angles, along with the ground truth, are shown in Figure \ref{mti_orient}.  

\begin{figure}[h!]
\begin{minipage}[t]{0.3\textwidth}
\centering
    \subfloat[Error in yaw angle]
    {\includegraphics[scale=0.175]{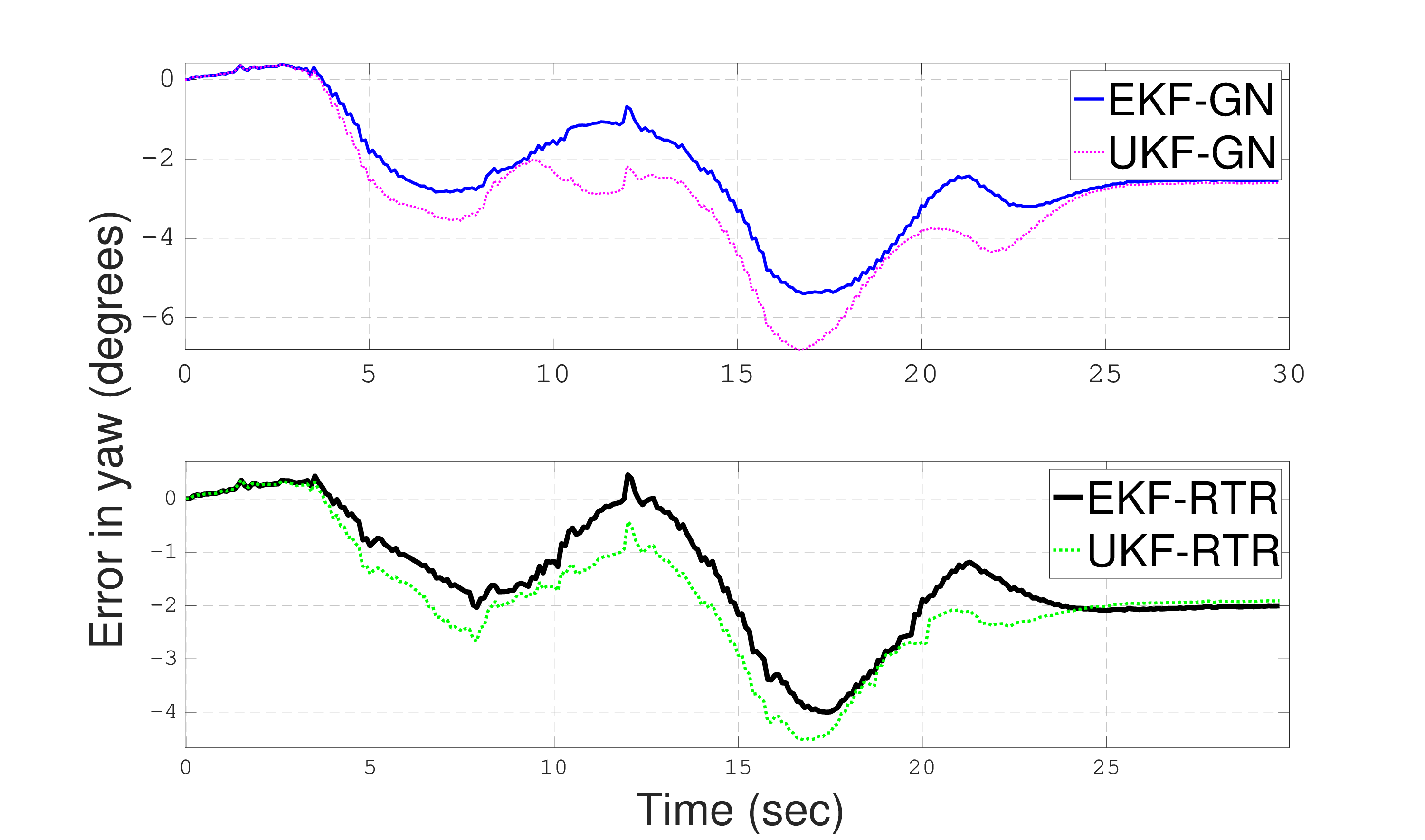}}
\end{minipage}
\begin{minipage}[t]{0.3\textwidth}
\centering
    \subfloat[Error in pitch angle ]
    {\includegraphics[scale=0.175]{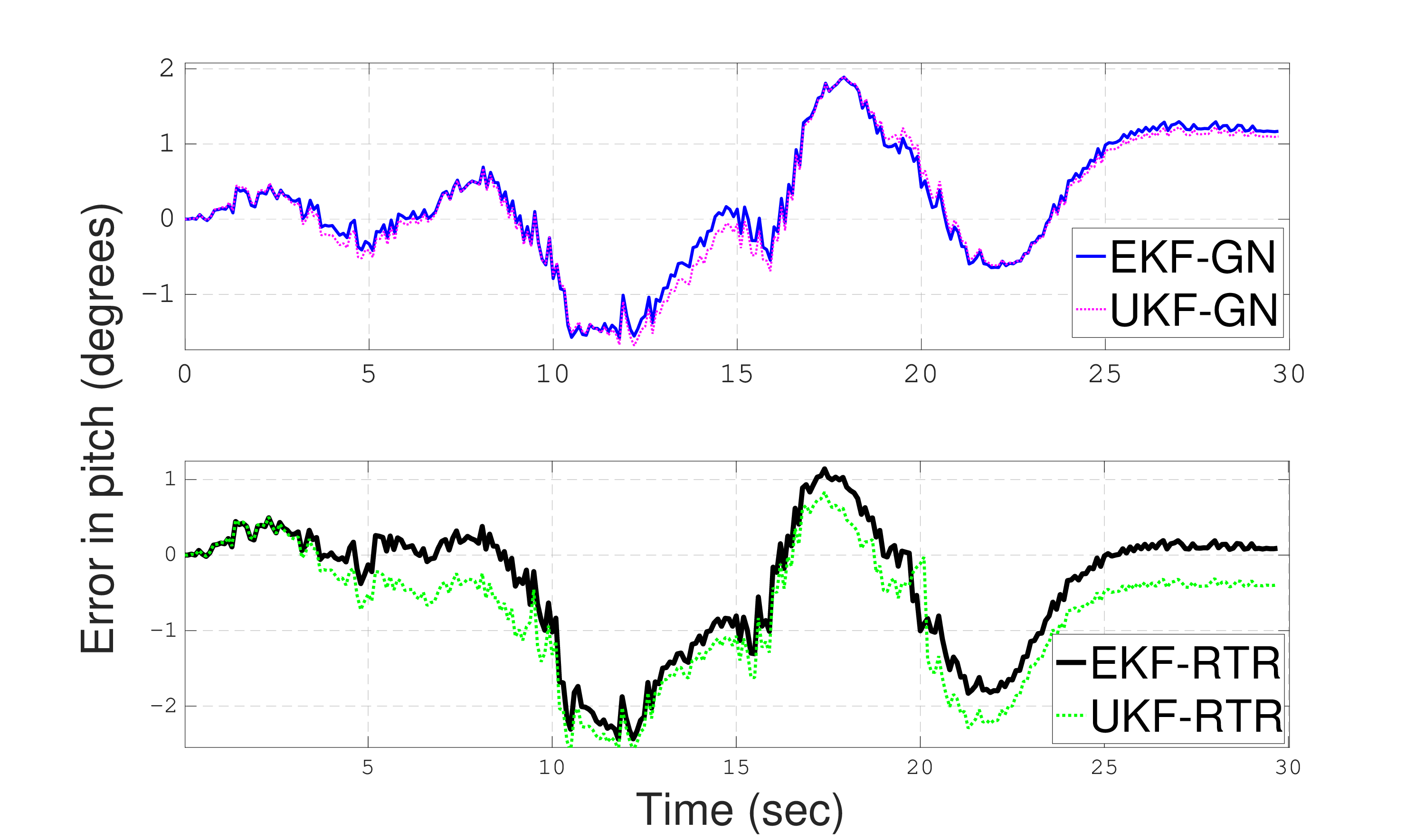}}
\end{minipage}
\begin{minipage}[t]{0.3\textwidth}
\centering
    \subfloat[Error in roll angle]
    {\includegraphics[scale=0.175]{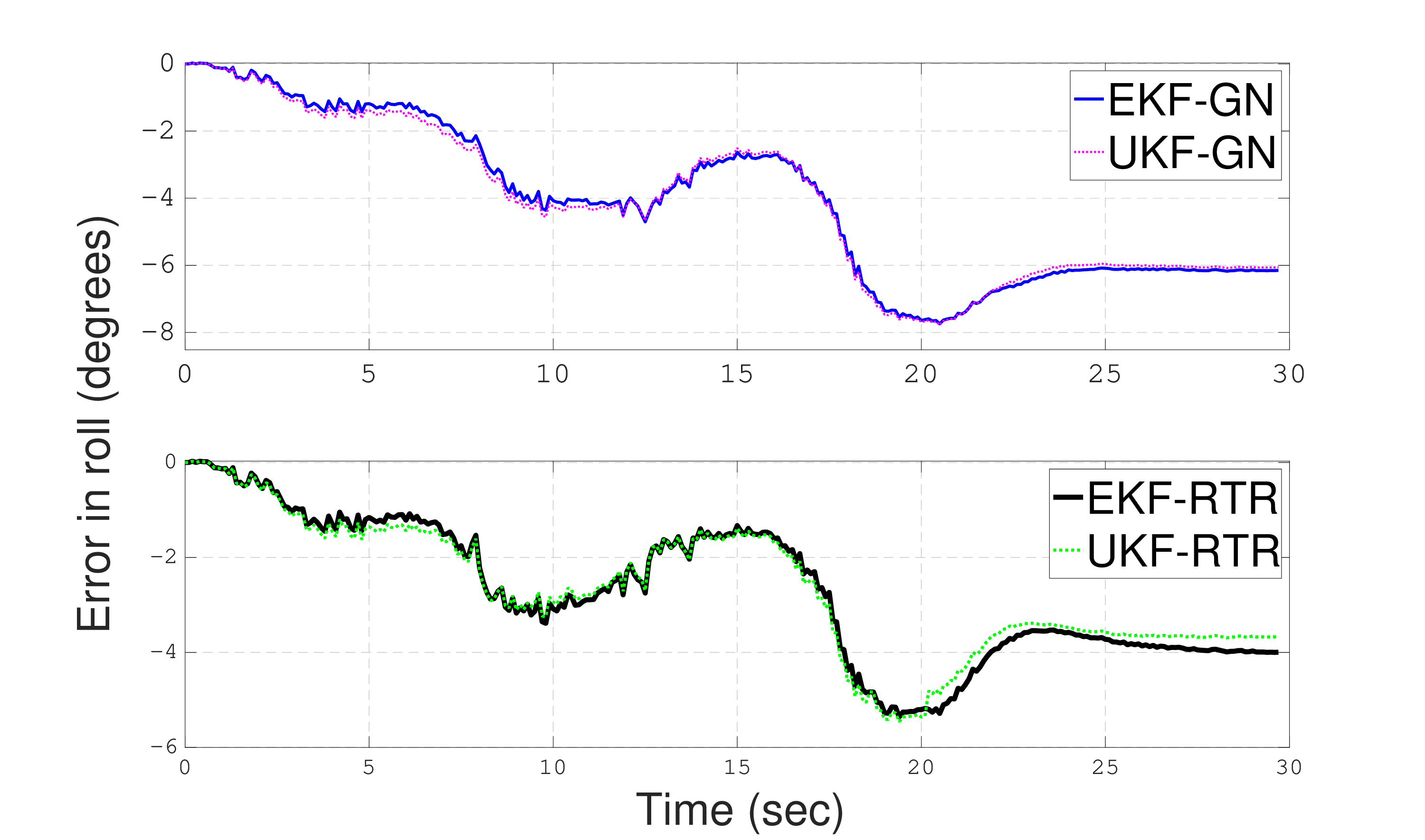}}
\end{minipage}
    \caption{Experimental evaluation: error in the estimated Euler angles}
    \label{mti_euler_error}
\end{figure}

The RMSE of the estimated angles is detailed in Table \ref{euler_table3}. Riemannian-based UKF and EKF show slight improvements over GN-based methods, as illustrated in the error plots in Figure \ref{mti_euler_error}.

\begin{table}[h!]
\centering
\begin{tabular}{|c|c|c|c|} 
  \hline
  Algorithm & Yaw & Pitch & Roll \\ 
   & RMSE & RMSE  & RMSE  \\ 
  \hline
  EKF-GN & $4.07$ & $0.67$ & $2.40$ \\ 
  \hline
  UKF-GN & $4.10$ & $0.69$ & $3.00$ \\ 
  \hline
  EKF-RTR & $2.68$ & $0.65$ & $1.55$ \\ 
  \hline
  UKF-RTR & $2.64$ & $0.86$ & $1.91$ \\ 
  \hline
\end{tabular}
\caption{RMSE in degrees of the estimated Euler angles}
\label{euler_table3}
\end{table}

\section{Conclusion}
This paper introduced an enhanced implementation of the Extended Kalman Filter (EKF) and Unscented Kalman Filter (UKF) for high-precision estimation of the 3D position and orientation of a moving target. The proposed filters integrate measurements from two navigation systems: an Inertial Navigation System (INS) based on IMU data and an acoustic navigation system, leveraging our previous work on Riemannian optimization \cite{9528945}. The output of these filters is projected onto an isosceles triangle manifold using a novel retraction algorithm.\par

The results demonstrate that using an array of acoustic receivers significantly improves accuracy compared to a single receiver. Furthermore, correcting the INS orientation with the receiver array results in substantial accuracy gains over benchmark methods. The Riemannian-based methods, which utilize the geometry of the receiver array as constraints, outperform the Gauss-Newton-based solver in localization performance.\par 

In summary, the proposed filter exhibits superior performance in all signal-to-noise ratio (SNR) scenarios, achieving better positioning and orientation accuracy than the benchmark methods.\par 

Future research will focus on implementing the UKF on the isosceles triangle manifold. This involves deriving the necessary tools to apply the unscented transform within the context of Riemannian manifolds \cite{hauberg2013unscented}.
	\bibliographystyle{IEEEtran}
	\bibliography{References_v2}

\begin{thebibliography}{10}
\providecommand{\url}[1]{#1}
\csname url@samestyle\endcsname
\providecommand{\newblock}{\relax}
\providecommand{\bibinfo}[2]{#2}
\providecommand{\BIBentrySTDinterwordspacing}{\spaceskip=0pt\relax}
\providecommand{\BIBentryALTinterwordstretchfactor}{4}
\providecommand{\BIBentryALTinterwordspacing}{\spaceskip=\fontdimen2\font plus
\BIBentryALTinterwordstretchfactor\fontdimen3\font minus \fontdimen4\font\relax}
\providecommand{\BIBforeignlanguage}[2]{{%
\expandafter\ifx\csname l@#1\endcsname\relax
\typeout{** WARNING: IEEEtran.bst: No hyphenation pattern has been}%
\typeout{** loaded for the language `#1'. Using the pattern for}%
\typeout{** the default language instead.}%
\else
\language=\csname l@#1\endcsname
\fi
#2}}
\providecommand{\BIBdecl}{\relax}
\BIBdecl

\bibitem{sun2023indoor}
Y.~Sun, W.~Wang, L.~Mottola, J.~Zhang, R.~Wang, and Y.~He, ``Indoor drone localization and tracking based on acoustic inertial measurement,'' \emph{IEEE Transactions on Mobile Computing}, 2023.

\bibitem{feng2020kalman}
D.~Feng, C.~Wang, C.~He, Y.~Zhuang, and X.-G. Xia, ``Kalman-filter-based integration of imu and uwb for high-accuracy indoor positioning and navigation,'' \emph{IEEE Internet of Things Journal}, vol.~7, no.~4, pp. 3133--3146, 2020.

\bibitem{khelifi2019survey}
F.~Khelifi, A.~Bradai, A.~Benslimane, P.~Rawat, and M.~Atri, ``A survey of localization systems in internet of things,'' \emph{Mobile Networks and Applications}, vol.~24, no.~3, pp. 761--785, 2019.

\bibitem{titterton2004strapdown}
D.~Titterton, J.~L. Weston, and J.~Weston, \emph{Strapdown inertial navigation technology}.\hskip 1em plus 0.5em minus 0.4em\relax IET, 2004, vol.~17.

\bibitem{9528945}
M.~H. AlSharif, A.~Douik, M.~Ahmed, T.~Y. Al-Naffouri, and B.~Hassibi, ``Manifold optimization for high-accuracy spatial location estimation using ultrasound waves,'' \emph{IEEE Transactions on Signal Processing}, vol.~69, pp. 5078--5093, 2021.

\bibitem{hightower2001location}
J.~Hightower and G.~Borriello, ``Location systems for ubiquitous computing,'' \emph{computer}, vol.~34, no.~8, pp. 57--66, 2001.

\bibitem{whitehouse2007practical}
K.~Whitehouse, C.~Karlof, and D.~Culler, ``A practical evaluation of radio signal strength for ranging-based localization,'' \emph{ACM SIGMOBILE Mobile Computing and Communications Review}, vol.~11, no.~1, pp. 41--52, 2007.

\bibitem{yuzbacsiouglu2005improved}
{\c{C}}.~Y{\"u}zba{\c{s}}io{\u{g}}lu and B.~Barshan, ``Improved range estimation using simple infrared sensors without prior knowledge of surface characteristics,'' \emph{Measurement Science and Technology}, vol.~16, no.~7, p. 1395, 2005.

\bibitem{amann2001laser}
M.-C. Amann, T.~Bosch, M.~Lescure, R.~Myllyla, and M.~Rioux, ``Laser ranging: a critical review of usual techniques for distance measurement,'' \emph{Optical engineering}, vol.~40, no.~1, pp. 10--19, 2001.

\bibitem{zafari2019survey}
F.~Zafari, A.~Gkelias, and K.~K. Leung, ``A survey of indoor localization systems and technologies,'' \emph{IEEE Communications Surveys \& Tutorials}, vol.~21, no.~3, pp. 2568--2599, 2019.

\bibitem{cappelli2023enhanced}
I.~Cappelli, F.~Carli, A.~Fort, M.~Intravaia, F.~Micheletti, G.~Peruzzi, and V.~Vignoli, ``Enhanced visible light localization based on machine learning and optimized fingerprinting in wireless sensor networks,'' \emph{IEEE Transactions on Instrumentation and Measurement}, vol.~72, pp. 1--10, 2023.

\bibitem{gabbrielli2022rails}
A.~Gabbrielli, J.~Bordoy, W.~Xiong, G.~K. Fischer, T.~Schaechtle, J.~Wendeberg, F.~H{\"o}flinger, C.~Schindelhauer, and S.~J. Rupitsch, ``Rails: 3-d real-time angle of arrival ultrasonic indoor localization system,'' \emph{IEEE Transactions on Instrumentation and Measurement}, vol.~72, pp. 1--15, 2022.

\bibitem{santoro2023uwb}
L.~Santoro, M.~Nardello, D.~Brunelli, and D.~Fontanelli, ``Uwb-based indoor positioning system with infinite scalability,'' \emph{IEEE Transactions on Instrumentation and Measurement}, vol.~72, pp. 1--11, 2023.

\bibitem{cypriani2009open}
M.~Cypriani, F.~Lassabe, P.~Canalda, and F.~Spies, ``Open wireless positioning system: A wi-fi-based indoor positioning system,'' in \emph{Vehicular Technology Conference Fall (VTC 2009-Fall), 2009 IEEE 70th}.\hskip 1em plus 0.5em minus 0.4em\relax IEEE, 2009, pp. 1--5.

\bibitem{kumar2014accurate}
S.~Kumar, S.~Gil, D.~Katabi, and D.~Rus, ``Accurate indoor localization with zero start-up cost,'' in \emph{Proceedings of the 20th annual international conference on Mobile computing and networking}.\hskip 1em plus 0.5em minus 0.4em\relax ACM, 2014, pp. 483--494.

\bibitem{alsharif2021range}
M.~H. AlSharif, M.~Saad, M.~Siala, M.~Ahmed, and T.~Y. Al-Naffouri, ``Range estimation of a moving target using ultrasound differential {Zadoff-Chu} codes,'' \emph{IEEE Transactions on Instrumentation and Measurement}, vol.~70, pp. 1--15, 2021.

\bibitem{wang2022smartphone}
H.~Wang, C.~Xue, Z.~Wang, L.~Zhang, X.~Luo, and X.~Wang, ``Smartphone-based pedestrian nlos positioning based on acoustics and imu parameter estimation,'' \emph{IEEE Sensors Journal}, 2022.

\bibitem{shi2023pedestrian}
L.-F. Shi, B.-L. Feng, Y.-F. Dai, G.-X. Liu, and Y.~Shi, ``Pedestrian indoor localization method based on integrated particle filter,'' \emph{IEEE Transactions on Instrumentation and Measurement}, vol.~72, pp. 1--10, 2023.

\bibitem{lee2021indoor}
G.~T. Lee, S.~B. Seo, and W.~S. Jeon, ``Indoor localization by kalman filter based combining of uwb-positioning and pdr,'' in \emph{2021 IEEE 18th Annual Consumer Communications \& Networking Conference (CCNC)}.\hskip 1em plus 0.5em minus 0.4em\relax IEEE, 2021, pp. 1--6.

\bibitem{priyantha2005cricket}
N.~B. Priyantha, ``The cricket indoor location system,'' Ph.D. dissertation, Massachusetts Institute of Technology, 2005.

\bibitem{fan2017data}
Q.~Fan, B.~Sun, Y.~Sun, Y.~Wu, and X.~Zhuang, ``Data fusion for indoor mobile robot positioning based on tightly coupled ins/uwb,'' \emph{The Journal of Navigation}, vol.~70, no.~5, pp. 1079--1097, 2017.

\bibitem{zhong2018integration}
Y.~Zhong, T.~Liu, B.~Li, L.~Yang, and L.~Lou, ``Integration of uwb and imu for precise and continuous indoor positioning,'' in \emph{2018 Ubiquitous Positioning, Indoor Navigation and Location-Based Services (UPINLBS)}.\hskip 1em plus 0.5em minus 0.4em\relax IEEE, 2018, pp. 1--5.

\bibitem{li2019indoor}
B.~Li, Z.~Hao, and X.~Dang, ``An indoor location algorithm based on kalman filter fusion of ultra-wide band and inertial measurement unit,'' \emph{AIP Advances}, vol.~9, no.~8, p. 085210, 2019.

\bibitem{gualda2021locate}
D.~Gualda, M.~C. P{\'e}rez-Rubio, J.~Ure{\~n}a, S.~P{\'e}rez-Bachiller, J.~M. Villadangos, {\'A}.~Hern{\'a}ndez, J.~J. Garc{\'\i}a, and A.~Jim{\'e}nez, ``Locate-us: Indoor positioning for mobile devices using encoded ultrasonic signals, inertial sensors and graph-matching,'' \emph{Sensors}, vol.~21, no.~6, p. 1950, 2021.

\bibitem{xu2020tightly}
Y.~Xu, Y.~S. Shmaliy, C.~K. Ahn, T.~Shen, and Y.~Zhuang, ``Tightly coupled integration of ins and uwb using fixed-lag extended ufir smoothing for quadrotor localization,'' \emph{IEEE Internet of Things Journal}, vol.~8, no.~3, pp. 1716--1727, 2020.

\bibitem{xie2017holding}
L.~Xie, J.~Tian, G.~Ding, and Q.~Zhao, ``Holding-manner-free heading change estimation for smartphone-based indoor positioning,'' in \emph{2017 IEEE 86th Vehicular Technology Conference (VTC-Fall)}.\hskip 1em plus 0.5em minus 0.4em\relax IEEE, 2017, pp. 1--5.

\bibitem{nguyen2016user}
P.~Nguyen, T.~Akiyama, H.~Ohashi, G.~Nakahara, K.~Yamasaki, and S.~Hikaru, ``User-friendly heading estimation for arbitrary smartphone orientations,'' in \emph{2016 International Conference on Indoor Positioning and Indoor Navigation (IPIN)}.\hskip 1em plus 0.5em minus 0.4em\relax IEEE, 2016, pp. 1--7.

\bibitem{bravo2017comparison}
J.~Bravo, E.~P. Herrera, and D.~A. Sierra, ``Comparison of step length and heading estimation methods for indoor environments,'' in \emph{2017 IEEE XXIV International Conference on Electronics, Electrical Engineering and Computing (INTERCON)}.\hskip 1em plus 0.5em minus 0.4em\relax IEEE, 2017, pp. 1--4.

\bibitem{manos2018gravity}
A.~Manos, I.~Klein, and T.~Hazan, ``Gravity direction estimation and heading determination for pedestrian navigation,'' in \emph{2018 International Conference on Indoor Positioning and Indoor Navigation (IPIN)}.\hskip 1em plus 0.5em minus 0.4em\relax IEEE, 2018, pp. 206--212.

\bibitem{hong2010heading}
S.~K. Hong and Y.-s. Ryuh, ``Heading measurements for indoor mobile robots with minimized drift using a mems gyroscopes,'' in \emph{Robot localization and map Building}.\hskip 1em plus 0.5em minus 0.4em\relax IntechOpen, 2010.

\bibitem{zhou2016pedestrian}
R.~Zhou, ``Pedestrian dead reckoning on smartphones with varying walking speed,'' in \emph{2016 IEEE International Conference on Communications (ICC)}.\hskip 1em plus 0.5em minus 0.4em\relax IEEE, 2016, pp. 1--6.

\bibitem{afzal2011magnetic}
M.~H. Afzal, V.~Renaudin, and G.~Lachapelle, ``Magnetic field based heading estimation for pedestrian navigation environments,'' in \emph{2011 International Conference on Indoor Positioning and Indoor Navigation}.\hskip 1em plus 0.5em minus 0.4em\relax IEEE, 2011, pp. 1--10.

\bibitem{angermann2012characterization}
M.~Angermann, M.~Frassl, M.~Doniec, B.~J. Julian, and P.~Robertson, ``Characterization of the indoor magnetic field for applications in localization and mapping,'' in \emph{2012 International Conference on Indoor Positioning and Indoor Navigation (IPIN)}.\hskip 1em plus 0.5em minus 0.4em\relax IEEE, 2012, pp. 1--9.

\bibitem{cho2006mems}
S.~Y. Cho and C.~G. Park, ``Mems based pedestrian navigation system,'' \emph{The Journal of Navigation}, vol.~59, no.~1, pp. 135--153, 2006.

\bibitem{kim2004step}
J.~W. Kim, H.~J. Jang, D.-H. Hwang, and C.~Park, ``A step, stride and heading determination for the pedestrian navigation system,'' \emph{Journal of Global Positioning Systems}, vol.~3, no. 1-2, pp. 273--279, 2004.

\bibitem{shin2012indoor}
B.~Shin, J.~H. Lee, H.~Lee, E.~Kim, J.~Kim, S.~Lee, Y.-s. Cho, S.~Park, and T.~Lee, ``Indoor 3d pedestrian tracking algorithm based on pdr using smarthphone,'' in \emph{2012 12th International Conference on Control, Automation and Systems}.\hskip 1em plus 0.5em minus 0.4em\relax IEEE, 2012, pp. 1442--1445.

\bibitem{ma2015hiheading}
W.~Ma, J.~Wu, C.~Long, and Y.~Zhu, ``Hiheading: smartphone-based indoor map construction system with high accuracy heading inference,'' in \emph{2015 11th International Conference on Mobile Ad-hoc and Sensor Networks (MSN)}.\hskip 1em plus 0.5em minus 0.4em\relax IEEE, 2015, pp. 172--177.

\bibitem{yang2016step}
X.~Yang, B.~Huang, and Q.~Miao, ``A step-wise algorithm for heading estimation via a smartphone,'' in \emph{2016 Chinese Control and Decision Conference (CCDC)}.\hskip 1em plus 0.5em minus 0.4em\relax IEEE, 2016, pp. 4598--4602.

\bibitem{poulose2019performance}
A.~Poulose, B.~Senouci, and D.~S. Han, ``Performance analysis of sensor fusion techniques for heading estimation using smartphone sensors,'' \emph{IEEE Sensors Journal}, vol.~19, no.~24, pp. 12\,369--12\,380, 2019.

\bibitem{alsharif2017zadoff}
M.~H. AlSharif, M.~Saad, M.~Siala, T.~Ballal, H.~Boujemaa, and T.~Y. Al-Naffouri, ``Zadoff-{Chu} coded ultrasonic signal for accurate range estimation,'' in \emph{Signal Processing Conference (EUSIPCO), 2017 25th European}.\hskip 1em plus 0.5em minus 0.4em\relax IEEE, 2017, pp. 1250--1254.

\bibitem{absil2009optimization}
P.-A. Absil, R.~Mahony, and R.~Sepulchre, \emph{Optimization algorithms on matrix manifolds}.\hskip 1em plus 0.5em minus 0.4em\relax Princeton University Press, 2009.

\bibitem{wolfe1969convergence}
P.~Wolfe, ``Convergence conditions for ascent methods,'' \emph{SIAM Review}, vol.~11, no.~2, pp. 226--235, 1969.

\bibitem{absil2007trust}
P.-A. Absil, C.~G. Baker, and K.~A. Gallivan, ``Trust-region methods on riemannian manifolds,'' \emph{Foundations of Computational Mathematics}, vol.~7, no.~3, pp. 303--330, 2007.

\bibitem{jazwinski2007stochastic}
A.~H. Jazwinski, \emph{Stochastic processes and filtering theory}.\hskip 1em plus 0.5em minus 0.4em\relax Courier Corporation, 2007.

\bibitem{wan2000unscented}
E.~A. Wan and R.~Van Der~Merwe, ``The unscented {Kalman} filter for nonlinear estimation,'' in \emph{Proceedings of the IEEE 2000 Adaptive Systems for Signal Processing, Communications, and Control Symposium (Cat. No. 00EX373)}.\hskip 1em plus 0.5em minus 0.4em\relax Ieee, 2000, pp. 153--158.

\bibitem{STM32}
\BIBentryALTinterwordspacing
STMicroelectronics. (2024) Stm32f469 discovery kit. [Online]. Available: \url{https://www.st.com/en/evaluation-tools/32f469idiscovery.html}
\BIBentrySTDinterwordspacing

\bibitem{NRF}
\BIBentryALTinterwordspacing
N.~Semiconductor. (2024) nrf24l. [Online]. Available: \url{https://www.nordicsemi.com/Products/nRF24-series}
\BIBentrySTDinterwordspacing

\bibitem{sony}
\BIBentryALTinterwordspacing
Sony. (2024) Xm-n1004 stereo amplifier. [Online]. Available: \url{https://www.sony.com/en-sa/electronics/car-amplifiers/xm-n1004}
\BIBentrySTDinterwordspacing

\bibitem{MTi}
\BIBentryALTinterwordspacing
Movella. (2024) Mti-1 datasheet. [Online]. Available: \url{https://www.xsens.com/hubfs/Downloads/Leaflets/MTi-1.pdf?__hstc=157421285.3f2f86467b8e14374d4860c1552d523b.1715494986238.1719859798440.1721973471183.3&__hssc=157421285.2.1721973471183&__hsfp=3563223768}
\BIBentrySTDinterwordspacing

\bibitem{OPTR}
\BIBentryALTinterwordspacing
https://optitrack.com/cameras/primex 41/. (2021) Optitrack system overview. [Online]. Available: \url{https://optitrack.com/cameras/primex-41/}
\BIBentrySTDinterwordspacing

\bibitem{alsharif2021robust}
M.~H. AlSharif, M.~Ahmed, A.~Felemban, A.~Zayat, A.~Muqaibel, M.~Masood, and T.~Y. Al-Naffouri, ``Robust 2d indoor positioning algorithm in the presence of non-line-of-sight signals,'' in \emph{2020 28th European Signal Processing Conference (EUSIPCO)}.\hskip 1em plus 0.5em minus 0.4em\relax IEEE, 2021, pp. 1802--1806.

\bibitem{zhu2023robust}
F.~Zhu, K.~Yu, Y.~Lin, C.~Wang, J.~Wang, and M.~Chao, ``Robust los/nlos identification for uwb signals using improved fuzzy decision tree under volatile indoor conditions,'' \emph{IEEE Transactions on Instrumentation and Measurement}, vol.~72, pp. 1--11, 2023.

\bibitem{hauberg2013unscented}
S.~Hauberg, F.~Lauze, and K.~S. Pedersen, ``Unscented {Kalman} filtering on riemannian manifolds,'' \emph{Journal of Mathematical Imaging and Vision}, vol.~46, no.~1, pp. 103--120, 2013.

\end{thebibliography}
\end{document}